\documentclass[acmtog]{acmart}
\usepackage{color}
\usepackage{subfigure}
\usepackage{units}
\usepackage{expl3}
\usepackage{amsmath}
\usepackage{multirow}

\usepackage{colortbl}

\definecolor{Yellow}{rgb}{1,1, 0.6}
\definecolor{Red}{rgb}{1, 0.6, 0.6}

\usepackage{booktabs} % For formal tables
\acmJournal{TOG}
\acmVolume{37}
\acmNumber{4}
\acmArticle{64}
\acmYear{2018}
\acmMonth{8}

\copyrightyear{2018} 
\acmYear{2018} 
\setcopyright{rightsretained} 
\acmConference[SIGGRAPH '18 Technical Paper]{Special Interest Group on Computer Graphics and Interactive Techniques Conference Technical Paper}{August 12--16, 2018}{Vancouver, BC, Canada}
\acmBooktitle{SIGGRAPH '18 Technical Paper: Special Interest Group on Computer Graphics and Interactive Techniques Conference Technical Paper, August 12--16, 2018, Vancouver, BC, Canada}
\acmDOI{10.1145/3197517.3201329}
\acmISBN{978-1-4503-5763-0/18/08}

\setcitestyle{square}

\begin{document}

\title{Synthetic Depth-of-Field with a Single-Camera Mobile Phone}

% \author{Anonymous Authors}
% \author{Anonymous Institute}
\author{Neal Wadhwa}
\author{Rahul Garg}
\author{David E. Jacobs}
\author{Bryan E. Feldman} 
\author{Nori Kanazawa}
\author{Robert Carroll}
\author{Yair Movshovitz-Attias}
\author{Jonathan T. Barron}
\author{Yael Pritch}
\author{Marc Levoy}
\affiliation{ Google Research\streetaddress{1600 Amphitheater Parkway}\city{Mountain View}\state{CA}\postcode{94043}}

%\authorsaddresses{}

\renewcommand{\shortauthors}{Wadhwa et al.}
% \renewcommand{\shortauthors}{Authors et al.}
% Teaser image is 
% Girl with fur hat: 0578_20170817_173642_151
\begin{teaserfigure}
\centering
\includegraphics[width=\textwidth]{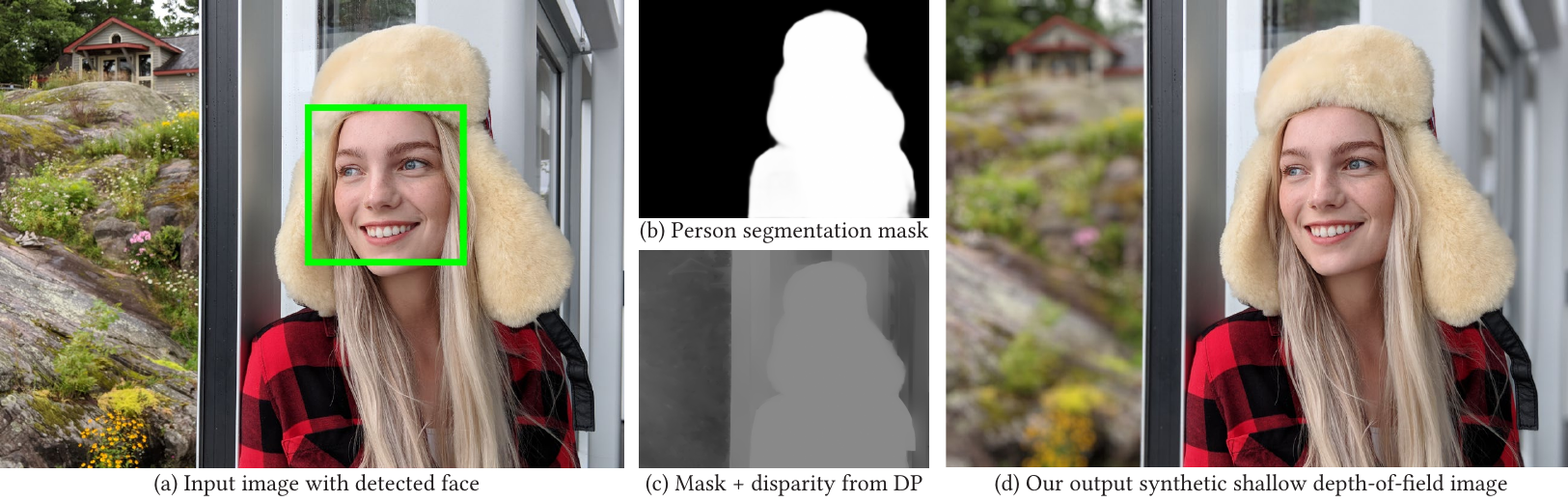}
   \caption{We present a system that uses a person segmentation mask (b) and a noisy depth map computed using the camera's dual-pixel (DP) auto-focus hardware (c) to produce a synthetic shallow depth-of-field image (d) with a depth-dependent blur on a mobile phone. 
   Our system is marketed as ``Portrait Mode'' on several Google-branded phones.}
   \label{fig:teaser}
\end{teaserfigure}

%%% The ``\maketitle'' command must appear after ``\begin{document}'' and,
%%% if you have one, after the definition of your ``teaser'' image, and
%%% before the first ``\section'' command.

\ExplSyntaxOn
\newcommand\latinabbrev[1]{
  \peek_meaning:NTF . {% Same as \@ifnextchar
    #1\@}%
  { \peek_catcode:NTF a {% Check whether next char has same catcode as \'a, i.e., is a letter
      #1.\@ }%
    {#1.\@}}}
\ExplSyntaxOff
\def\etal{\latinabbrev{et al}}

\newcommand{\dof}{DoF}

\begin{abstract}
% This should all be moved to the introduction. 
%Shallow depth-of-field is a photographic property frequently used to focus attention on a subject by blurring out a distracting background.
%It arises naturally from the optical characteristics of large aperture lenses and is a significant part of the distinctive look of photographs from a professional DSLR-style cameras.
%Reproducing the same effect on mobile phone cameras is difficult because their small form factor precludes using lenses large enough to produce the effect with optics alone.
%Computational approaches rely on generating a depth map using either expensive additional hardware, such as a second camera, or a burdensome user experience, such as being required to move to the phone during capture. 

Shallow depth-of-field is commonly used by photographers to isolate a subject from a distracting background.
However, standard cell phone cameras cannot produce such images optically, as their short focal lengths and small apertures capture nearly all-in-focus images.
We present a system to computationally synthesize shallow depth-of-field images with a single mobile camera and a single button press.
If the image is of a person, we use a person segmentation network to separate the person and their accessories from the background. 
If available, we also use dense \emph{dual-pixel} auto-focus hardware, effectively a $2$-sample light field with an approximately $1$ millimeter baseline, to compute a dense depth map.
These two signals are combined and used to render a defocused image.
Our system can process a $5.4$ megapixel image in $4$ seconds on a mobile phone, is fully automatic, and is robust enough to be used by non-experts. 
The modular nature of our system allows it to degrade naturally in the absence of a dual-pixel sensor or a human subject. 
%If triggered on a camera without PDAF pixels e.g., most front-facing ``selfie'' cameras, it will use learning-based segmentation alone, and if pointed at an object rather than a person, it will use the depth map alone. 

\end{abstract}


\maketitle

\section{Introduction}

Depth-of-field is an important aesthetic quality of photographs.
It refers to the range of depths in a scene that are imaged sharply in focus.
This range is determined primarily by the aperture of the capturing camera's lens: a wide aperture produces a shallow (small) depth-of-field, while a narrow aperture produces a wide (large) depth-of-field.
Professional photographers frequently use depth-of-field as a compositional tool.
In portraiture, for instance, a strong background blur and shallow depth-of-field allows the photographer to isolate a subject from a cluttered, distracting background.
The hardware used by DSLR-style cameras to accomplish this effect also makes these cameras expensive, inconvenient, and often difficult to use. Therefore, the compelling images they produce are largely limited to professionals.
Mobile phone cameras are ubiquitous, but their lenses have apertures too small to produce the same kinds of images optically.
%\david{On the fence about the DoF abbreviation.}

Recently, mobile phone manufacturers have started computationally producing shallow depth-of-field images.
The most common technique is to include two cameras instead of one and to apply stereo algorithms to captured image pairs to compute a depth map.
One of the images is then blurred according to this depthmap.
However, adding a second camera raises manufacturing costs, increases power consumption during use, and takes up space in the phone.
Some manufacturers have instead chosen to add a time-of-flight or structured-light direct depth sensor to their phones, but these also tend to be expensive and power intensive, in addition to not working well outdoors.
{\em Lens Blur} \cite{LensBlur} is a method of producing shallow depth-of-field images without additional hardware, but it requires the user to move the phone during capture to introduce parallax.
This can result in missed photos and negative user experiences if the photographer fails to move the camera at the correct speed and trajectory, or if the subject of the photo moves.

\begin{figure}[t]
\centering
\centering
\subfigure[A segmentation mask obtained from the front-facing camera.]{
\label{fig:teaser2-a}
\begin{tabular}{c@{\hskip1pt}c@{\hskip1pt}c}
\includegraphics[width=0.32\linewidth]{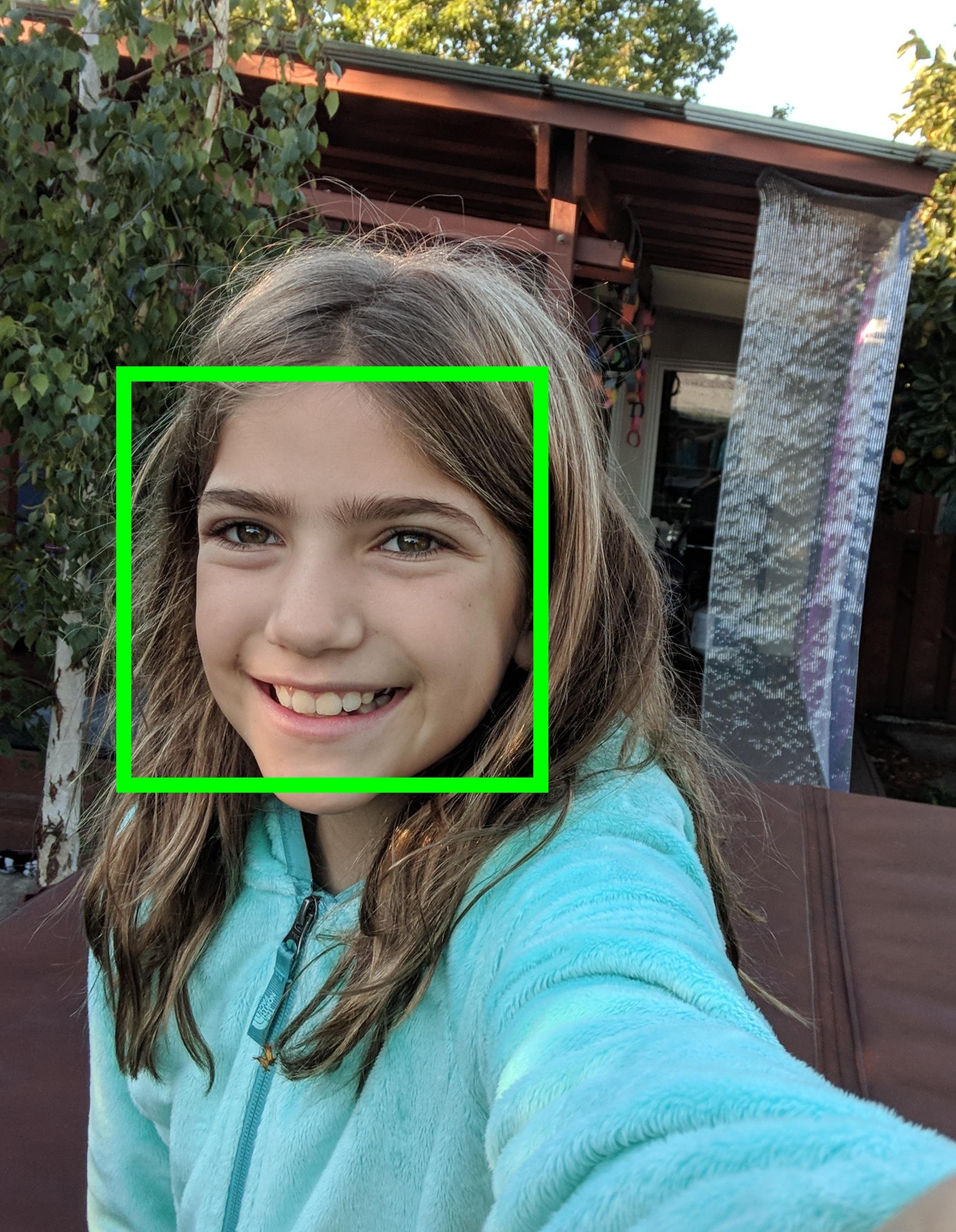} & 
\includegraphics[width=0.32\linewidth]{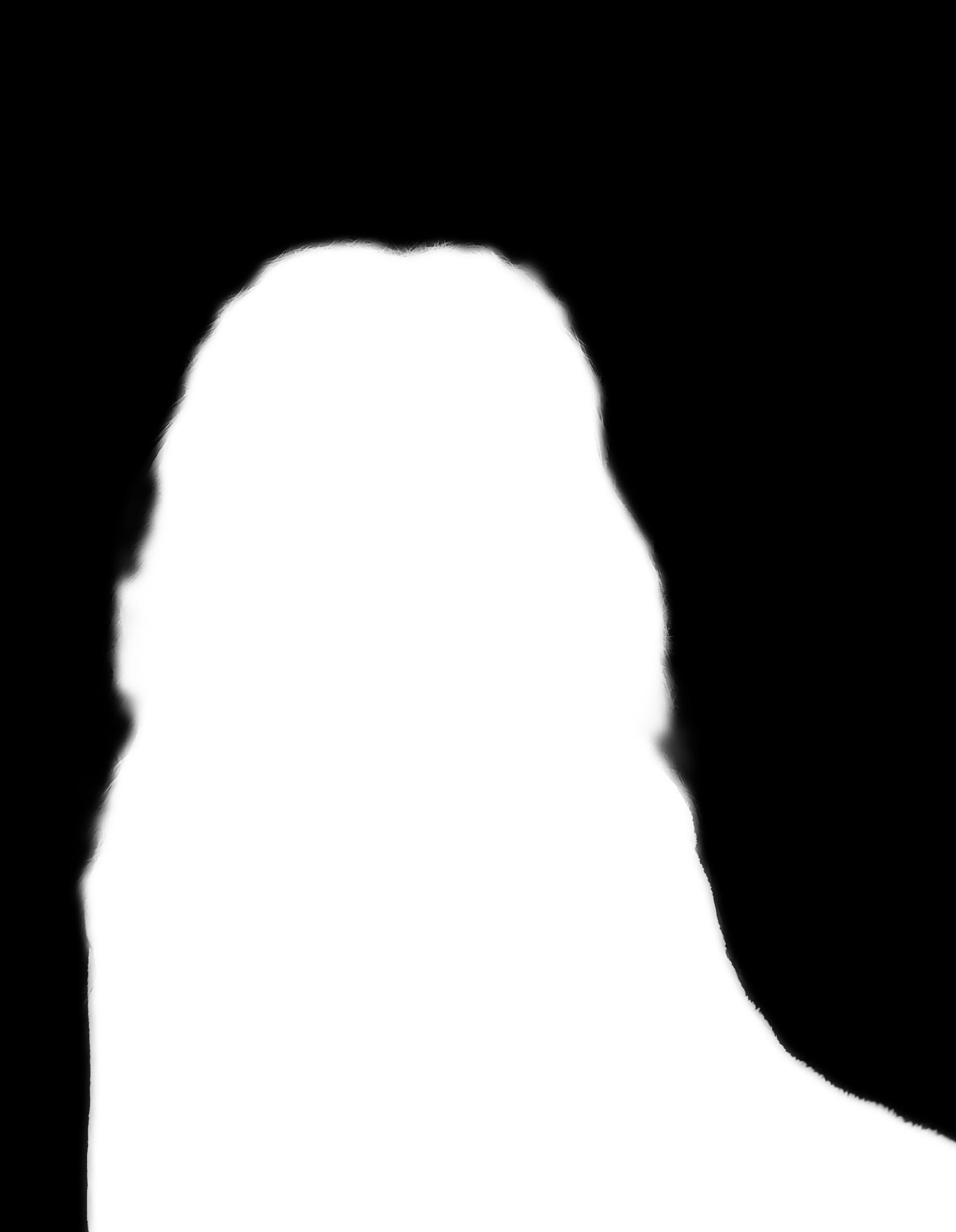} & 
\includegraphics[width=0.32\linewidth]{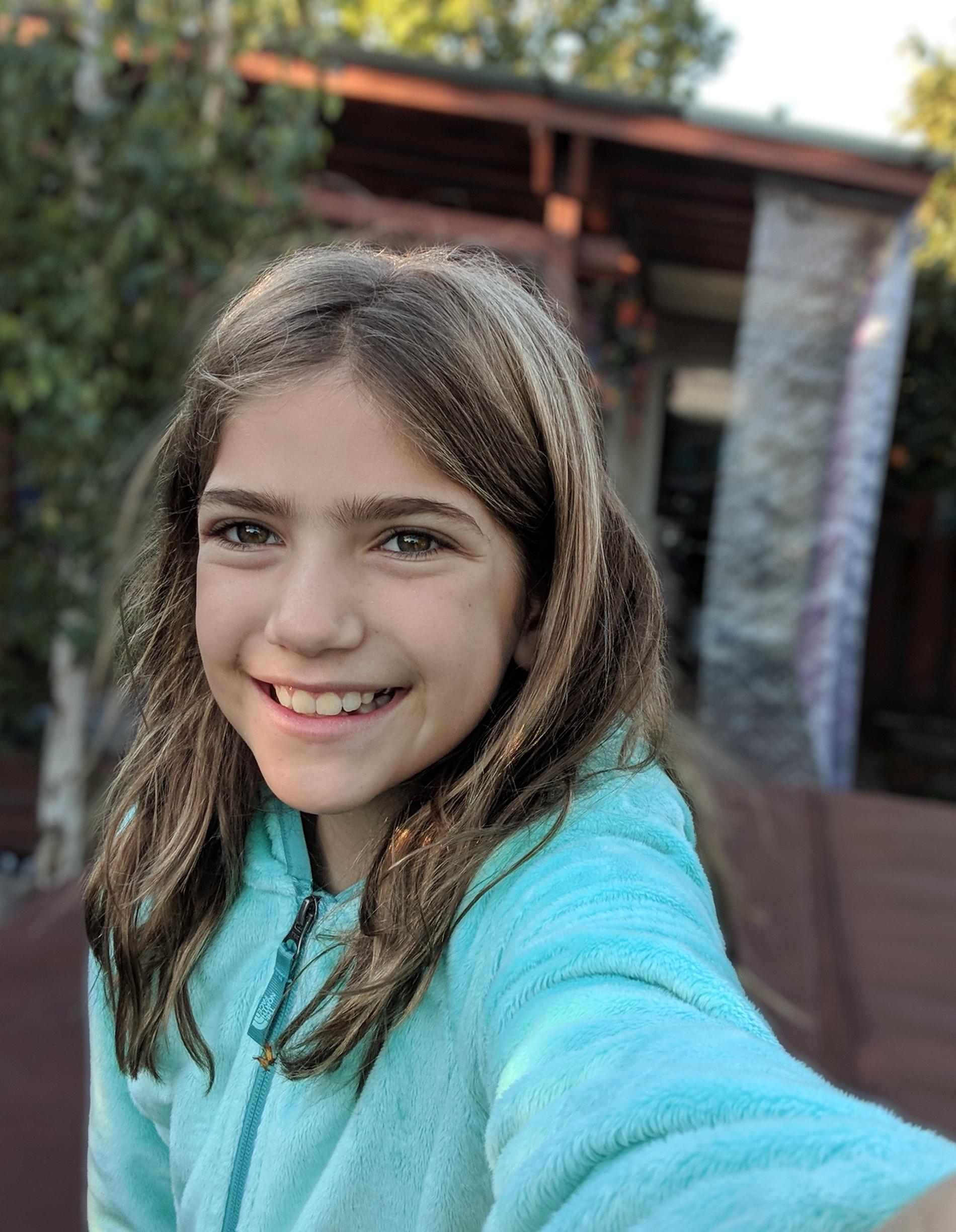}  \\
\small Input & \small Mask & \small Output \\
\end{tabular}
}\\
\vspace{-6pt}
\subfigure[The dual-pixel (DP) disparity from a scene without people.]{
\label{fig:teaser2-b}
\begin{tabular}{c@{\hskip1pt}c@{\hskip1pt}c}
\includegraphics[width=0.32\linewidth]{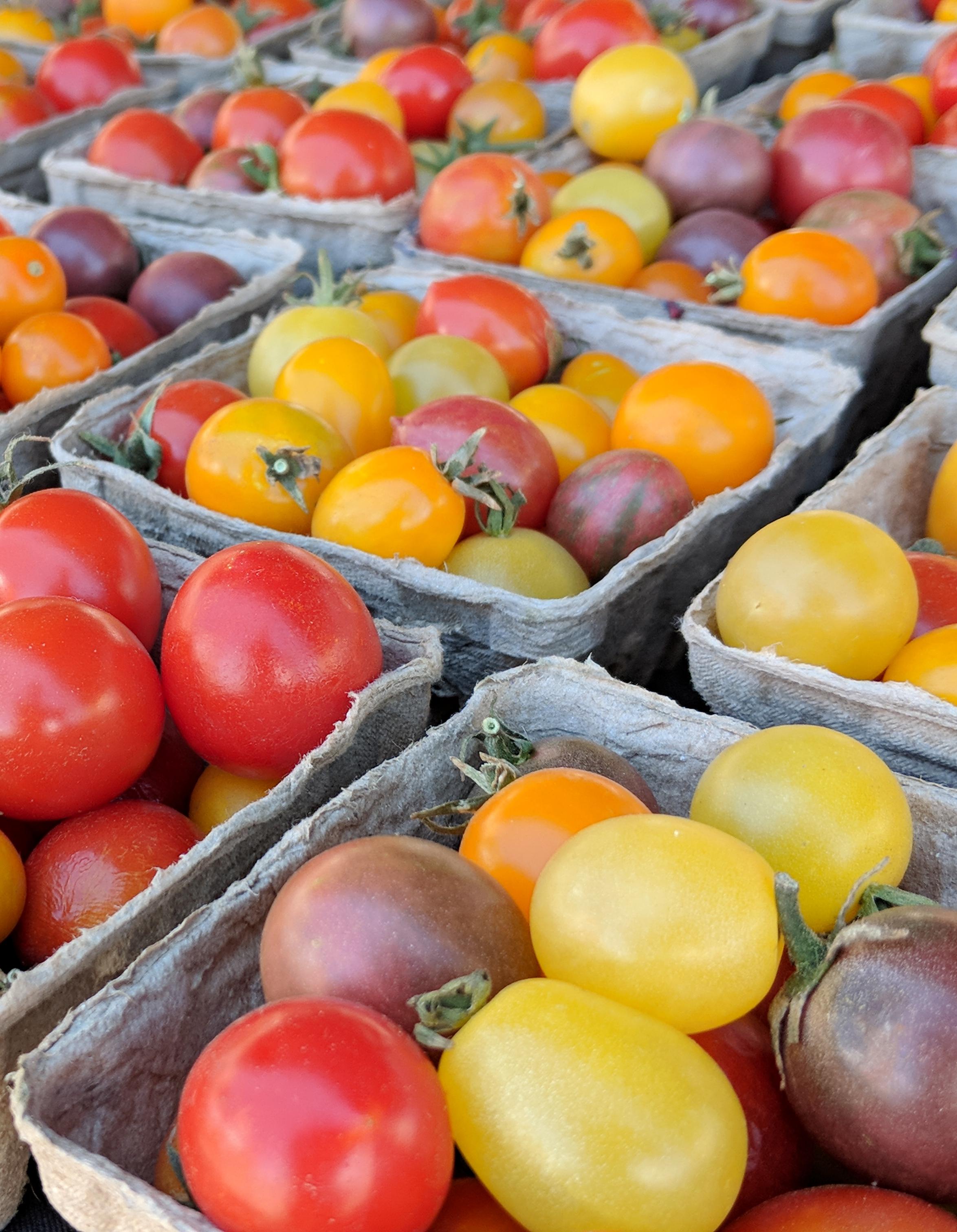} & 
\includegraphics[width=0.32\linewidth]{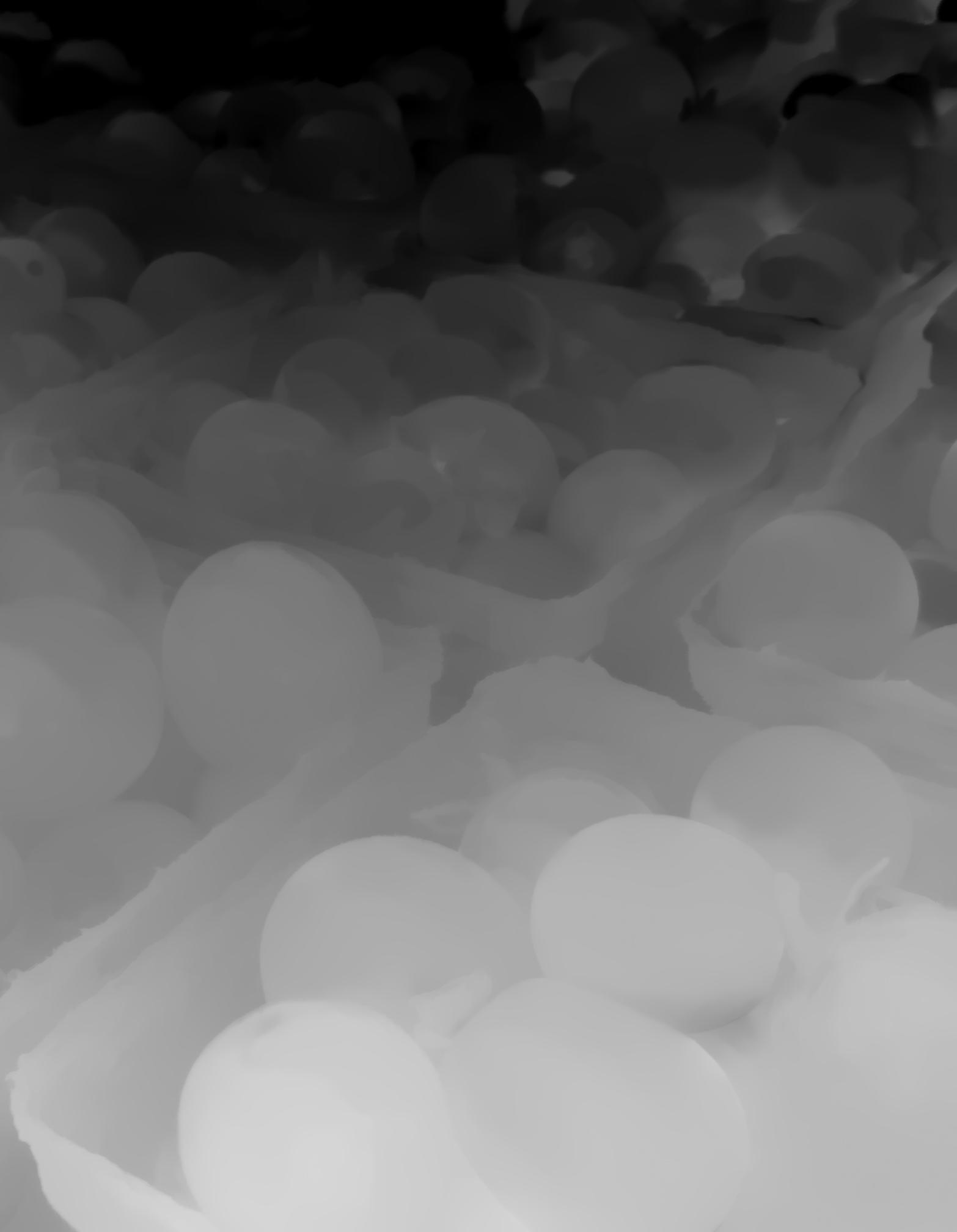} & 
\includegraphics[width=0.32\linewidth]{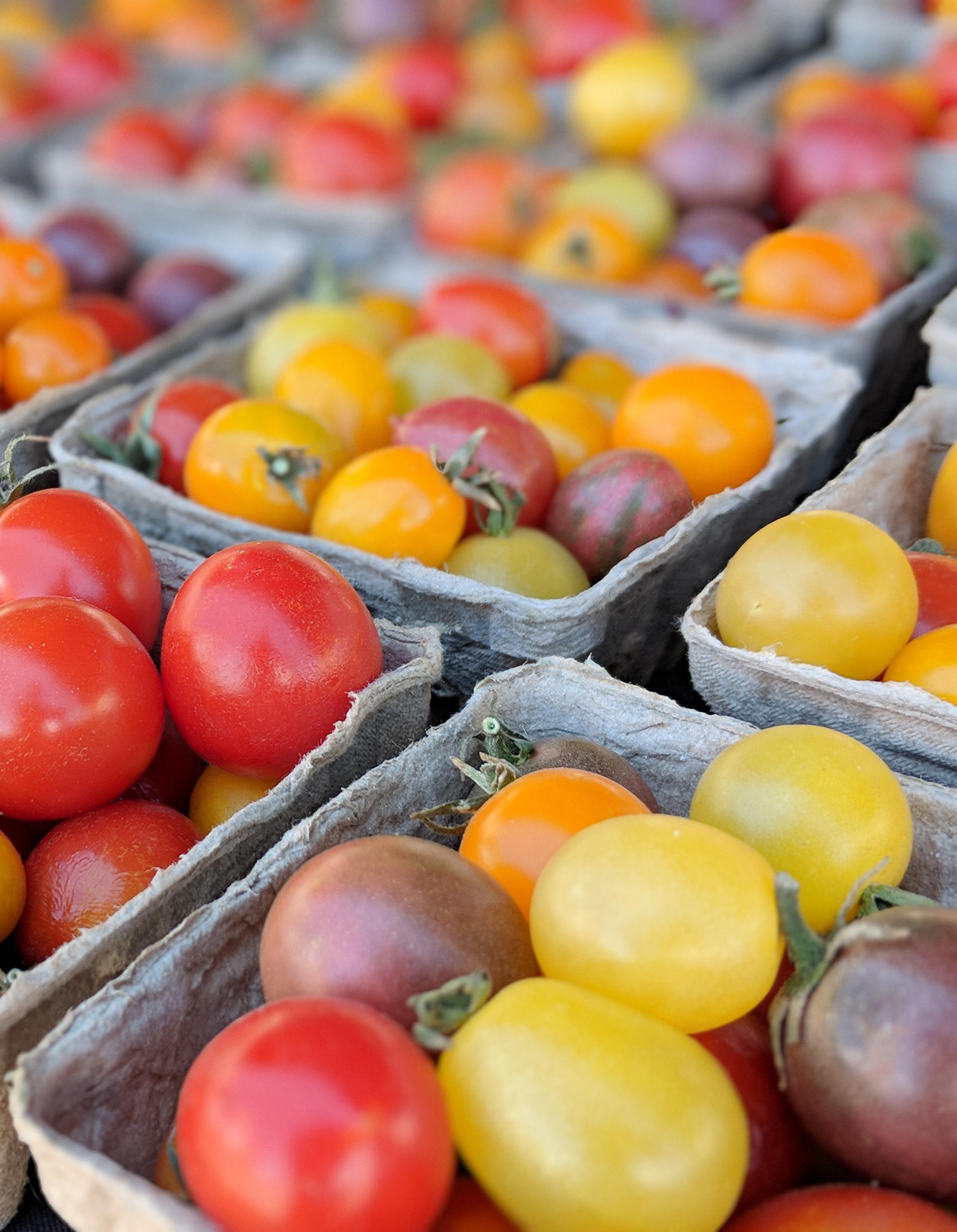}  \\
\small Input & \small Disparity & \small Output \\
\end{tabular}
}%
\vspace{-8pt}
\caption{Our system gracefully falls back to one of the two inputs depending on availability. On the front-facing camera (which lacks dual-pixels), the image is almost always of a person in front of a distant background, so using the segmentation alone is sufficient (a). For close-up shots of objects, disparity data from dual-pixels alone is often sufficient to produce a high-quality output (b). }
\vspace{-12pt}
\label{fig:teaser2}
\end{figure}

We introduce a system that allows untrained photographers to take shallow depth-of-field images on a wide range of mobile cameras with a single button press.
%If available, we leverage the camera's \emph{dual-pixel} (DP) auto-focus hardware.
%Such hardware is increasingly common on modern mobile phones, where it is traditionally used to provide fast auto-focus.
%But our system is also designed to degrade gracefully to work on a single camera without any additional hardware for depth sensing.
We aim to provide a user experience that combines the best features of a DSLR and a smartphone. This leads us to the following requirements for such a system:
\begin{enumerate}
    %\item Zero shutter lag, to allow the user to capture the intended moment.
    \item Fast processing and high resolution output.
    \item A standard smartphone capture experience---a single button-press to capture, with no extra controls and no requirement that the camera is moved during capture.
    \item Convincing-looking shallow depth-of-field results with plausible blur and the subject in sharp focus.
    \item Works on a wide range of scenes.
\end{enumerate}

Our system opportunistically combines two different technologies and is able to function with only one of them.
The first is a neural network trained to segment out people and their accessories. This network takes an image and a face position as input and outputs a mask that indicates the pixels which belong to the person or objects that the person is holding.
Second, if available, we use a sensor with \emph{dual-pixel} (DP) auto-focus hardware, which effectively gives us a 2-sample light field with a narrow ${\sim}1$ millimeter baseline. Such hardware is increasingly common on modern mobile phones, where it is traditionally used to provide fast auto-focus.
From this new kind of DP imagery, we extract dense depth maps.

%Making the face detection boxes part of the input allows us to easily prevent people in the background from being made part of the mask.

Modern mobile phones have both front and rear facing cameras.
The front-facing camera is typically used to capture \emph{selfies}, i.e., a close up of the photographer's head and shoulders against a distant background. This camera is usually fixed-focused and therefore lacks dual-pixels.
However, for the constrained category of selfie images, we found it sufficient to only segment out people using the trained segmentation model and to apply a uniform blur to the background (Fig.~\ref{fig:teaser2}(a)).

%Considering now the rear-facing camera, its photos may not contain people, and
%even if they do, scene content may be such that a non-depth dependent blur may
%look artificial, e.g., a person standing on a ground plane that is entirely visible to the camera.  

%For photos without people, we rely solely on a depthmap computed from DP pixels, which we use to apply a depth dependent blur (Fig.~\ref{fig:teaser2}(b)). For photos with people, we \emph{improve} that depthmap using the computed segmentation and then use it to blur the image (Fig.~ref{fig:teaser}). 

In contrast, we need depth information for photos taken by the rear-facing camera. Depth variations in scene content may make a uniform blur look unnatural, e.g., a person standing on the ground. For such photos of people, we
augment our segmentation with a depthmap computed from dual-pixels, and use this augmented input to drive our synthetic blur (Fig.~\ref{fig:teaser}).  If there are no people in the photo, we use the
DP depthmap alone (Fig.~\ref{fig:teaser2}(b)).  Since the stereo baseline of dual-pixels is very small (${\sim}1$ mm), this latter solution works only for macro-style photos of small objects or nearby scenes.

%The inputs to our system are a color image and, if available, raw dual-pixel data comprising two single channel images.
%We run a face detector on the color image and identify any \emph{salient} faces, i.e., faces corresponding to the subject(s) being photographed.
Our system works as follows.
We run a face detector on an input color image and identify the faces of the subjects being photographed.
%A neural network is used to infer a low resolution mask segmenting people corresponding to the identified faces in the color image, which is then upsampled to full resolution using a edge-aware filtering step.
A neural network uses the color image and the identified faces to infer a low-resolution mask that segments the people that the faces belong to. The mask is then upsampled to full resolution using edge-aware filtering.
This mask can be used to uniformly blur the background while keeping the subject sharp.

If DP data is available, we compute a depthmap by first aligning and averaging a burst of DP images to reduce noise using the method of Hasinoff \etal~\shortcite{hasinoff2016burst}.
We then use a stereo algorithm based on Anderson \etal~\shortcite{anderson2016jump} to infer a set of low resolution and noisy disparity estimates.
The small stereo baseline of the dual-pixels causes these estimates to be strongly affected by optical aberrations. We present a calibration procedure to correct for them.
The corrected disparity estimates are upsampled and smoothed using bilateral space techniques \cite{barron2015fast,kopf2007joint} to yield a high resolution disparity map.

Since disparity in a stereo system is proportional to defocus blur from a lens
having an aperture as wide as the stereo baseline, we can use these disparities
to apply a synthetic blur, thereby simulating shallow depth of field.  While
this effect is not the same as optical blur, it is similar enough in most
situations that people cannot tell the difference.  In fact, we deviate further
from physically correct defocusing by forcing a range of depths on either side
of the in-focus plane to stay sharp; this ``trick'' makes it easier for novices
to take compelling shallow-depth-of-field pictures. For pictures of people, where we have
a segmentation mask, we further deviate from physical correctness by keeping
pixels in the mask sharp.
%- again to make it easier for novices to take good
%shallow-depth-of-field pictures.

%We show that these disparities are proportional to the amount of defocus introduced by the camera lens and can be used directly to amplify the defocus, thereby yielding a synthetic shallow depth-of-field image with physically correct blur.
%However, we deviate from this physically correct model to one that yields plausible blur appearance while making it easier for the user to keep the entire subject in focus.
%A conventional DSLR requires choosing the aperture size and focal length carefully to achieve the same effect.
%For photos containing people, we leverage the segmentation mask to clearly identify which pixels belong to the subject and map them to a very narrow depth range to ensure that they are in focus.

Our rendering technique divides the scene into several layers at different disparities, splats pixels to translucent disks according to disparity and then composites the different layers weighted by the actual disparity.
This results in a pleasing, smooth depth-dependent rendering.
Since the rendered blur reduces camera noise which looks unnatural adjacent to in-focus regions that retain that noise, we add synthetic noise to our defocused regions to make the results appear more realistic.

The wide field-of-view of a typical mobile camera is ill-suited for portraiture. It causes a photographer to stand near subjects leading to unflattering perspective distortion of their faces. To improve the look of such images, we impose a forced $1.5\times$ digital zoom. In addition to forcing the photographer away from the subject, the zoom also leads to faster running times as we process fewer pixels ($5.4$ megapixels instead of the full sensor's $12$ megapixels). Our entire system (person segmentation, depth estimation, and defocus rendering) is fully automatic and runs in ${\sim}4$ seconds on a modern smartphone.

\section{Related Work}
Besides the approach of Hern{\'a}ndez \shortcite{LensBlur}, there is academic work on rendering synthetic shallow depth-of-field images from a single camera. While Hern{\'a}ndez \shortcite{LensBlur} requires deliberate up-down translation of the camera during capture, other works exploit parallax from accidental hand shake \cite{Ha2016,Yu2014}. Both these approaches suffer from frequent failures due to insufficient parallax due to the user not moving the camera correctly or the accidental motion not being sufficiently large.
Suwajanakorn \etal~\shortcite{Supasorn2015} and Tang \etal~\shortcite{Tang2017} use defocus cues to extract depth but require capturing multiple images that increases the capture time.
Further, these approaches have trouble with non-static scenes and are too compute intensive to run on a mobile device.

Monocular depth estimation methods may also be used to infer depth from a single image and use it to render a synthetic shallow depth-of-field image.
Such techniques pose the problem as either inverse rendering \cite{Horn75,Barron15} or supervised machine learning \cite{Hoiem2005,saxena2009,eigen2014,liu2016} and have seen significant progress, but this problem is highly underconstrained compared to multi-image depth estimation and hence difficult.
Additionally, learning-based approaches often fail to generalize well beyond the datasets they are trained on and do not produce the high resolution depth maps needed to synthesize shallow depth-of-field images. Collecting a diverse and high quality depth dataset is challenging. Past work has used direct depth sensors, but these only work well indoors and have low spatial resolution. Self-supervised approaches \cite{garg2016,xie2016,zhou2017,godard2017} do not require ground truth depth and can learn from stereo data but fail to yield high quality depth.

Shen \etal~\shortcite{shen2016automatic} achieve impressive results on generating synthetic shallow depth-of-field from a single image by limiting to photos of people against a distant background.
They train a convolutional neural network to segment out people and then blur the background assuming the person and the background are at two different but constant depths.
In \cite{shen2016matting}, they extend the approach by adding a differentiable matting layer to their network.
Both these approaches are computationally expensive taking $0.2$ and $0.6$ seconds respectively for a $800 \times 600$ output on a NVIDIA Titan X, a powerful desktop GPU, making them infeasible for a mobile platform.
Zhu \etal~\shortcite{Zhu2017} use smaller networks, but segmentation based approaches do not work for photos without people and looks unnatural for more complex scenes in which there are objects at the same depth as the person, e.g., Fig.~\ref{fig:teaser}(a). 

\section{Person Segmentation}
\label{sec:segmentation}
A substantial fraction of images captured on mobile phones are of people. Since such images are ubiquitous, we trained a neural network to segment people and their accessories in images. We use this segmentation both on its own and to augment the noisy disparity from DP data (Sec.~\ref{sec:merge_mask}). 

The computer vision community has put substantial effort into creating high-quality algorithms to semantically segment objects and people in images \cite{girshick2015fast,maskrcnn2017}. % ,pinheiro2016learning
While Shen \etal~\shortcite{shen2016automatic} also learn a neural network to segment out people in photos to render a shallow depth-of-field effect, our contributions include:
(a) training and data collection methodologies to train a fast and accurate segmentation model capable of running on a mobile device, and
(b) edge-aware filtering to upsample the mask predicted by the neural network (Sec.~\ref{sec:edge_aware_filtering_of_masks}).

%We are guided by two principles in learning a neural network model to segment people: The model should (a) be small and run fast on a mobile device, and (b) accurately %segment out people in images representative of those typically taken by using their mobile devices. We describe the decisions we make as a result.

%\neal{How about (c) the ability to control who is segmented by making face locations an input to the network?}

%Also see Fast Deep Matting for Portrait Animation on Mobile Phone.
\begin{figure}[t]
    \centering

    \includegraphics[width=\linewidth]{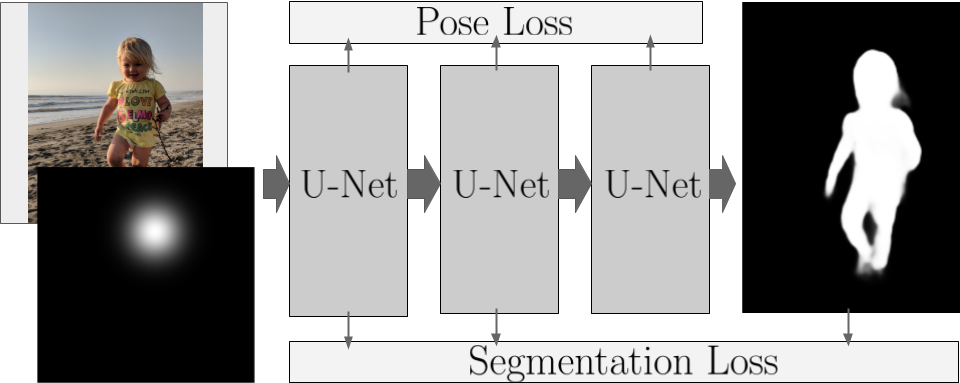}
    \vspace{-12pt}
    \caption{\textbf{Person Segmentation Network.} RGB image (top-left) and face location (bottom-left) are the inputs to a three stage model with pose and segmentation losses after each stage.}
    \vspace{-8pt}
    \label{fig:segmentation_network}
\end{figure}

\subsection{Data Collection}
\label{sec:data_collection}

To train our neural network, we downloaded $122$k images from Flickr (\url{www.flickr.com}) that contain between $1$ to $5$ faces and annotated a polygon mask outlining the people in the image.
The mask is refined using the filtering approach described in Sec.~\ref{sec:edge_aware_filtering_of_masks}.
We augment this with data from Papandreou \etal~\shortcite{Papandreou2017} consisting of $73$k images with $227$k person instances containing \emph{only} pose labels, i.e., locations of $17$ different keypoints on the body. While we do not infer pose, we predict pose at training time which is known to improve segmentation results \cite{tripathi2017}.
%Specifically, at a given point in our data collection/annotation we run our current best prediction model on a batch of unlabeled images and ask human annotators to identify prediction failures. Annotators then draw a polygon mask only for such failures. Polygon masks are then refined using the filtering approach described in Section \ref{sec:edge_aware_filtering_of_masks}.
%As our model gets better at segmenting out people, we focus on segmenting out harder scenarios like people wearing hats or sunglasses, people with their pets, etc. by restricting  to images containing the problematic objects using pre-trained image object classifiers.
Finally, as in Xu \etal~\shortcite{Xu2017}, we create a set of synthetic training images by compositing the people in portrait images onto different backgrounds generating an additional $465$k images. Specifically, we downloaded $30{,}974$ portraits images and $13{,}327$ backgrounds from Flickr.
For each of the portraits images, we compute an alpha matte using Chen \etal~\shortcite{Chen2012}, and composite the person onto $15$ randomly chosen background images.

We cannot stress strongly enough the
importance of good training data for this segmentation task: choosing a wide
enough variety of poses, discarding poor training images, cleaning up
inaccurate polygon masks, etc.  With each improvement we made over a 9-month
period in our training data, we observed the quality of our defocused portraits
to improve commensurately.

\subsection{Training}
\label{sec:training}

Given the training set, we use a network architecture consisting of 3 stacked U-Nets \cite{Ronneberger2015} with intermediate supervision after each stage similar to Newell \etal~\shortcite{Newell2016} (Fig.~\ref{fig:segmentation_network}).
The network takes as input a $4$ channel $256 \times 256$ image, where $3$ of the channels correspond to the RGB image resized and padded to $256 \times 256$ resolution preserving the aspect ratio.
The fourth channel encodes the location of the face as a posterior distribution of an isotropic Gaussian centered on the face detection box with a standard deviation of $21$ pixels and scaled to be $1$ at the mean location.
Each of the three stages outputs a segmentation mask --- a $256 \times 256 \times 1$ output of a layer with sigmoid activation, and a $64 \times 64 \times 17$ output containing heatmaps corresponding to the locations of the $17$ keypoints similar to Tompson \etal~\shortcite{Tompson2014}.

We use a two stage training process. In the first stage, we train with cross entropy losses for both segmentation and pose, which are weighted by a $1:5$ ratio. After the first stage of training has converged, we remove the pose loss and prune training images for which the model predictions had large $L1$ error for pixels in the interior of the mask,
%whose distance to the ground truth mask's boundary was $\ge 0.05\max(w,h)$
i.e., we only trained using examples with errors near the object boundaries. Large errors distant from the object boundary can be attributed to either annotation error or model error.
It is obviously beneficial to remove training examples with annotation error.
In the case of model error, we sacrifice performance on a small percentage of images to focus on improving near the boundaries for a large percentage of images.

%Specifically, we observed that despite a large number of accurate training examples our initial converged model still produced results that often did not conform well to object boundaries. To improve prediction at object boundaries we continued training using a pruned version of the data which 

Our implementation is in Tensorflow \cite{tensorflow2015-whitepaper}. We use $660$k images for training which are later pruned to $511$k images by removing images with large prediction errors in the interior of the mask.
Our evaluation set contains $1700$ images. We use a batch size of $16$ and our model was trained for a month on $40$ GPUs across $10$ machines using stochastic gradient descent with a learning rate of $0.1$, which was later lowered to $10^{-4}$. We augment the training data by applying a rotation chosen uniformly between $\left[-10, 10\right]$ degrees, an isotropic scaling chosen uniformly in the range $\left[0.4, 1.2\right]$ and a translation of up to $10\%$ of each of the image dimensions.
The values given in this section were arrived through empirical testing.

\subsection{Inference}
\label{sec:inference}

At inference time, we are provided with an RGB image and face rectangles output by a face detector.
Our model is trained to predict the segmentation mask corresponding to the face location in the input (Fig.~\ref{fig:segmentation_network}).  As a heuristic to avoid including bystanders in the segmentation mask, we seed the network only with faces that are at least one third the area of the largest face and larger than 1.3\% the area of the image.
%\nori{The face seed allows us to control which people are segmented out. We employ a strategy where the largest faces are considered foreground and faces smaller than a percentage of the largest face are considered background.}
When there are multiple faces, we perform inference for each of the faces and take the maximum of each face's real-valued segmentation mask $M_i(\mathbf{x})$ as our final mask $M(\mathbf{x}) = \max_i M_i(\mathbf{x})$.
$M(\mathbf{x})$ is upsampled and filtered to become a high resolution edge-aware mask (Sec.~\ref{sec:edge_aware_filtering_of_masks}). This mask can be used to generate a shallow depth-of-field result, or combined with disparity (Sec.~\ref{sec:merge_mask}).

\subsection{Edge-Aware Filtering of a Segmentation Mask}
\label{sec:edge_aware_filtering_of_masks}

\begin{figure}[t]%
\centering
\subfigure[RGB Image]{%
\label{fig:refinement-a}%
\includegraphics[width=0.32\linewidth]{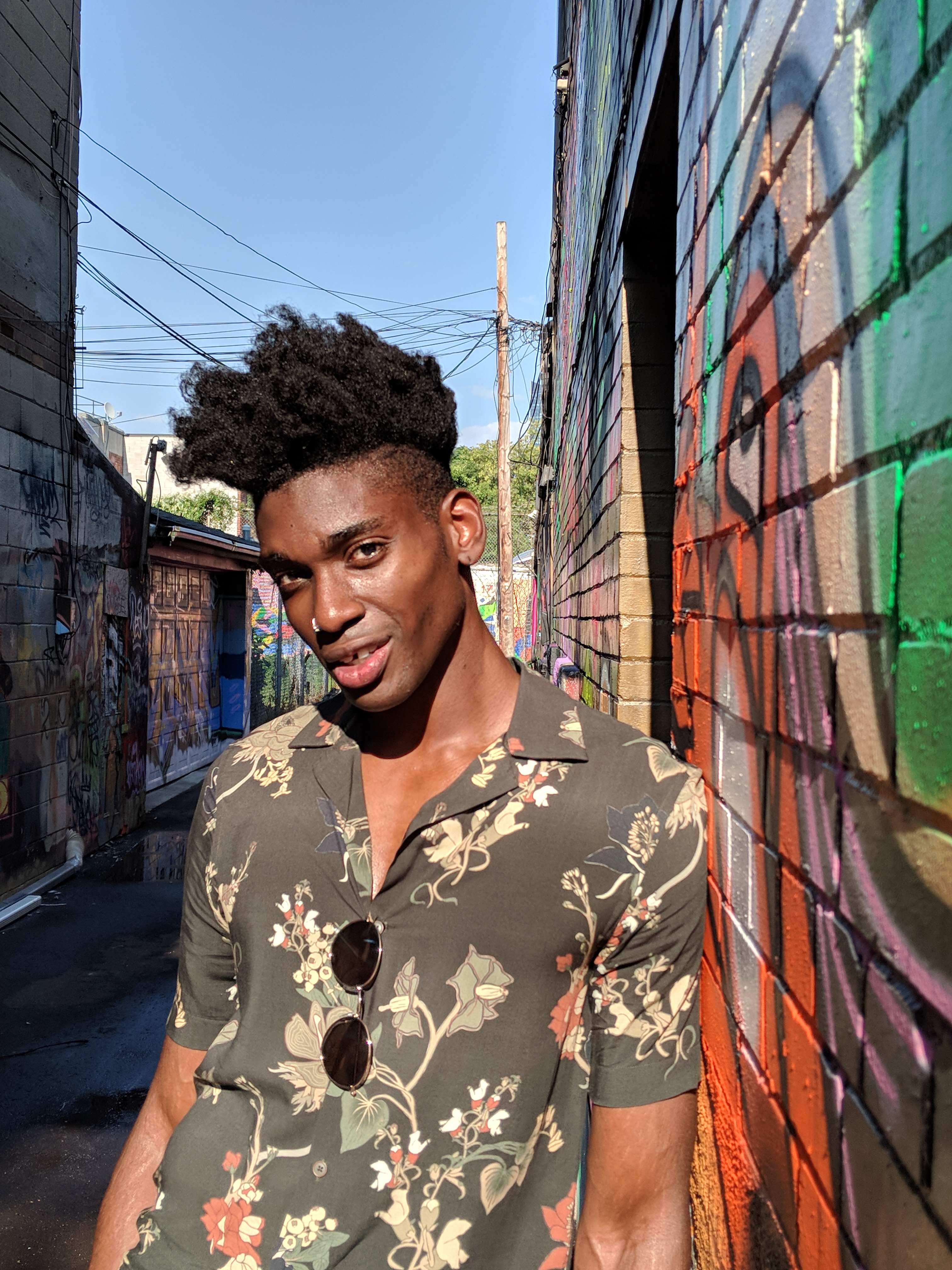}}%
\hspace{1pt}
\subfigure[Coarse Mask]{%
\label{fig:refinement-b}%
\includegraphics[width=0.32\linewidth]{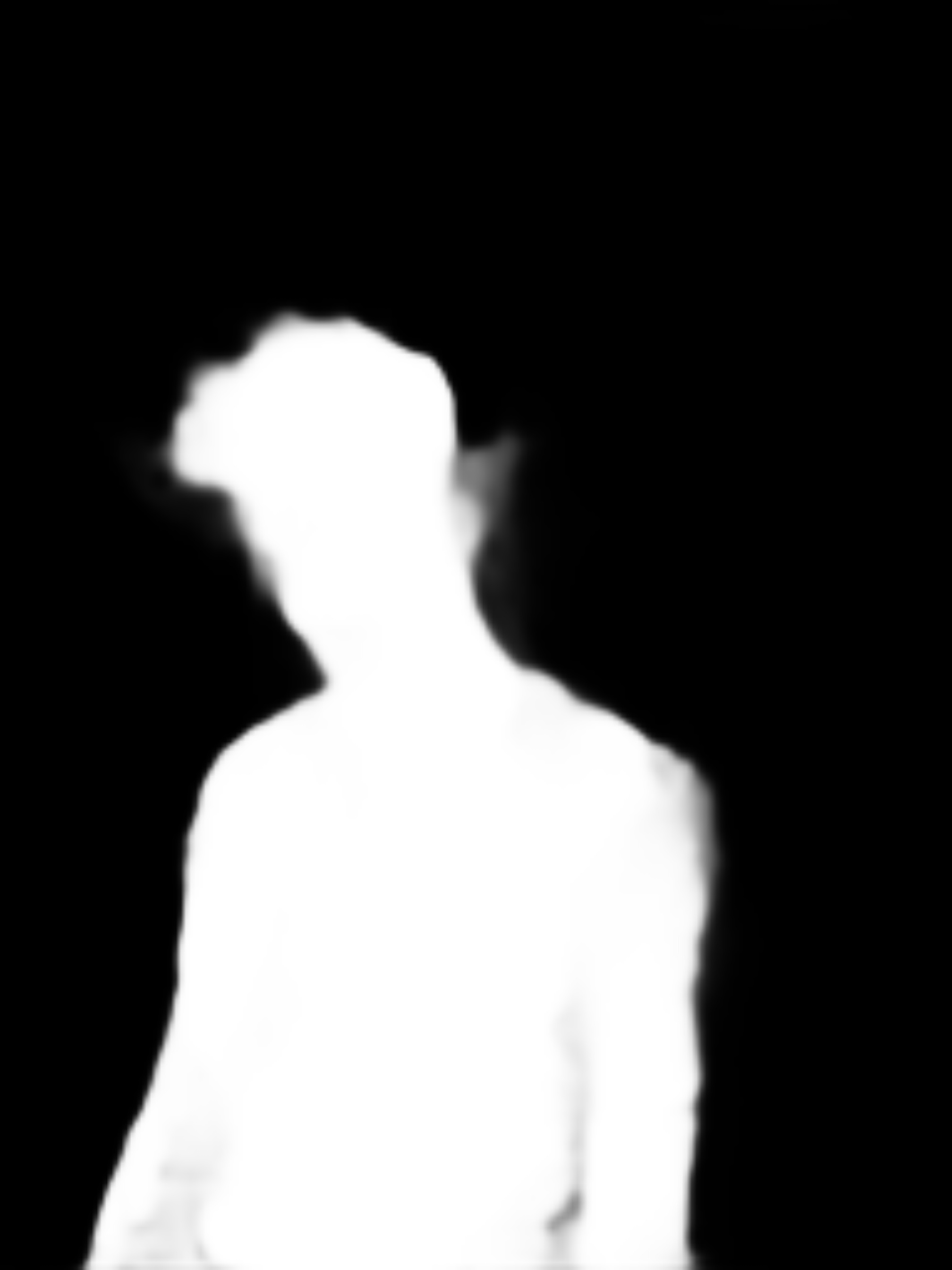}}%
\hspace{1pt}
\subfigure[Filtered Mask]{%
\label{fig:refinement-c}%
\includegraphics[width=0.32\linewidth]{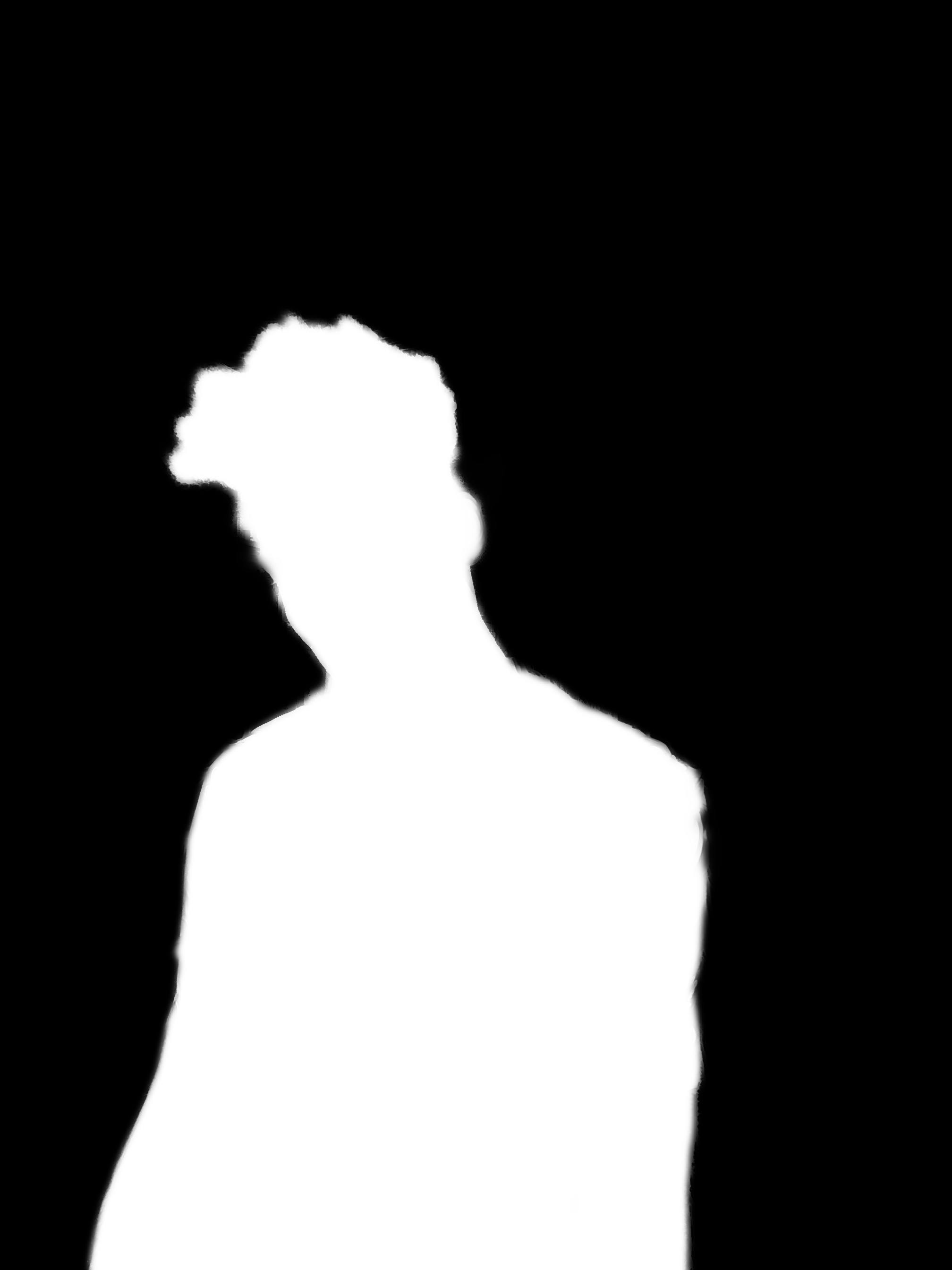}}%
\vspace{-6pt}
\caption{Edge-aware filtering of a segmentation mask.
}
\label{fig:refinement}
\vspace{-6pt}
\end{figure}

Compute and memory requirements make it impractical for a neural network to directly predict a high resolution mask.
Using the prior that mask boundaries are often aligned with image edges, we use an edge-aware filtering approach to upsample the low resolution mask $M(\mathbf{x})$ predicted by the network.
We also use this filtering to refine the ground truth masks used for training---this enables human annotators to only provide approximate mask edges, thus improving the quality of annotation given a fixed human annotation time.

\newcommand{\onehalf}{\nicefrac{1}{2}}

Let $M_c(\mathbf{x})$ denote a coarse segmentation mask to be refined. In the case of a human annotated mask, $M_c(\mathbf{x})$ is the same resolution as the image and is binary valued with pixels set to $1$ inside the supplied mask and $0$ elsewhere.
In the case of the low-resolution predicted mask, we bilinearly upsample $M(\mathbf{x})$ from $256 \times 256$ to image resolution to get $M_c(\mathbf{x})$, which has values between 0 and 1 inclusive (Fig. \ref{fig:refinement-b}). We then compute a confidence map, $C(\mathbf{x})$, from $M_c(\mathbf{x})$ using the heuristic that we have low confidence in a pixel if the predicted value is far from either $0$ or $1$ or the pixel is spatially near the mask boundary. Specifically,
\begin{equation}
C(\mathbf{x}) = \left({M_c(\mathbf{x}) - \onehalf \over \onehalf} \right)^2 \ominus \mathbf{1}_{k \times k}
\end{equation}
where $\ominus$ is morphological erosion and $\mathbf{1}_{k \times k}$ is a $k \times k$ square structuring element of 1's, with $k$ set to $5\%$ of the larger of the dimensions of the image.
Given $M_c(\mathbf{x})$, $C(\mathbf{x})$ and the corresponding RGB image $I(\mathbf{x})$, we compute the filtered segmentation mask $M_f(\mathbf{x})$ by using the fast bilateral solver \cite{barron2016fast}, denoted as $\mathrm{BS}(\cdot)$, to do edge-aware smoothing. We then push the values towards either $0$ or $1$ by applying a sigmoid function. Specifically,
\begin{equation}
M_f(\mathbf{x}) = {1 \over 1 + \exp \left(-k(\mathrm{BS}(M_c( \mathbf{x}), C(\mathbf{x}), I(\mathbf{x})) - \onehalf) \right)}.
\end{equation}
Running the bilateral solver at full resolution is slow and can generate speckling in highly textured regions. Hence, we run the solver at half the size of $I(\mathbf{x})$, smooth any high frequency speckling by applying a Gaussian blur to $M_f(\mathbf{x})$, and upsample via joint bilateral upsampling \cite{kopf2007joint} with $I(\mathbf{x})$ as the guide image to yield the final filtered mask (Fig. \ref{fig:refinement-c}).
We will use the bilateral solver again in Sec.~\ref{sec:bilateral_smoothing} to smooth noisy disparities from DP data.
%\barron{to prevent future confusion, it might be worth mentioning here that we'll revisit the solver later, where it will be a post-processing step and not a training data pre-processing step.}

\subsection{Accuracy and Efficiency}
\label{sec:segmentation_comparisons}

We compare the accuracy of our model against the PortraitFCN+ model from Shen \etal~\shortcite{shen2016automatic} by computing the mean Intersection-over-Union (IoU), i.e.,  area(output $\cap$ ground truth) / area(output $\cup$ ground truth), over their evaluation dataset.
Our model trained on their data has a higher accuracy than their best model, which demonstrates the effectiveness of our model architecture.
Our model trained on only our training data has an even higher accuracy, thereby demonstrating the value of our training data (Table~\ref{tab:compare_portraitfcn}).

\begin{table}
    \centering
    \caption{Comparison of our model with PortraitFCN+ model from \protect\cite{shen2016automatic} on their evaluation data.}
    \vspace{-8pt}
    \begin{tabular}{l | l | c}
        Model & Training data & Mean IoU \\
        \hline \hline
        PortraitFCN+  & \cite{shen2016automatic} & $95.91\%$ \\
        \hline
        \multirow{2}{*}{Our model}  & \cite{shen2016automatic}  & $97.01\%$ \\ \cline{2-3}
                  & Our training data & $97.70\%$
    \end{tabular}
    \label {tab:compare_portraitfcn}
\vspace{0.2in}
    \centering
  \caption{Comparison of our model with Mask-RCNN \protect\cite{maskrcnn2017} on our evaluation dataset.}
  \vspace{-8pt}
    \begin{tabular}{l | l | c}
        Model & Training data & Mean IoU \\
        \hline \hline
         Mask-RCNN & Our training data & $94.63\%$ \\ \hline
         Our model & Our training data & $95.80\%$ \\
    \end{tabular}
    %Comparison result between Mask-RCNN and Portrait Mode Network Model.  Both Network Models have been trained with our internal datasets.  There are 1150 images in the evaluation dataset.}
    \label{tab:compare_maskrcnn}
    \vspace{-8pt}
\end{table}

We also compare against a state-of-the-art semantic segmentation model Mask-RCNN \cite{maskrcnn2017} by training and testing it on our data (Table~\ref{tab:compare_maskrcnn}).
We use our own implementation of Mask-RCNN with a backbone of Resnet-101-C4.
We found that Mask-RCNN gave inferior results when trained and tested on our data while being a significantly larger model.
Mask-RCNN is designed to jointly solve detection and segmentation for multiple classes and may not be suitable for single class segmentation with known face location and high quality boundaries.

Further, our model has orders of magnitude fewer operations per inference --- $3.07$ Giga-flops compared to $607$ for PortraitFCN+ and $3160$  for Mask-RCNN as measured using the Tensorflow Model Benchmark Tool~\shortcite{tensorflow2015-whitepaper}. For PortraitFCN+, we benchmarked the Tensorflow implementation of the FCN-8s model from Long \etal~\shortcite{long2015} on which PortraitFCN+ is based.%(\url{https://github.com/tensorflow/tensorflow/tree/master/tensorflow/tools/benchmark}). 

\section{Depth from a Dual-Pixel Camera}
\label{sec:disparity}

\begin{figure}[t!]
    \centering
    \includegraphics[width=\columnwidth]{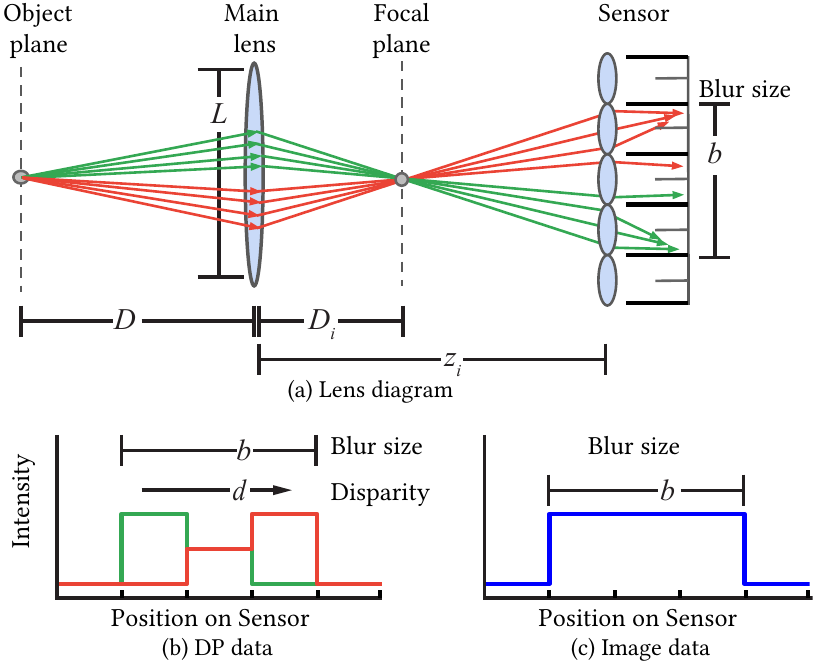}
\caption{A thin lens model showing the relationship between depth $D$, blur diameter $b$, and disparity $d$. An out-of-focus object emits light that travels through the camera's main lens with aperture diameter $L$, focuses in-front of the sensor at distance $D_i$ from the lens and then produces a three-pixel wide blur (a). The left and right half-pixels see light from opposite halves of the lens. The images from the left and right pixels are shifted with disparity proportional to the blur size (b). When summed together, they produce an image that one would expect from a sensor without dual pixels (c).}
    \label{fig:pd_light_field}
    \vspace{-12pt}
\end{figure}

Dual-pixel (DP) auto-focus systems work by splitting pixels in half, such that the left half integrates light over the right half of the aperture and vice versa (Fig.~\ref{fig:pd_light_field}). Because image content is optically blurred based on distance from the focal plane, there is a shift, or disparity, between the two views that depends on depth and on the shape of the blur kernel. This system is normally used for auto-focus, where it is sometimes called {\em phase-detection} auto-focus. In this application, the lens position is iteratively adjusted until the average disparity value within a focus region is zero and, consequently, the focus region is sharp. Many modern sensors split every pixel on the sensor, so the focus region can be of arbitrary size and position. We re-purpose the DP data from these dense split-pixels to compute depth.

\begin{figure*}
    \centering
    \includegraphics[width=\textwidth]{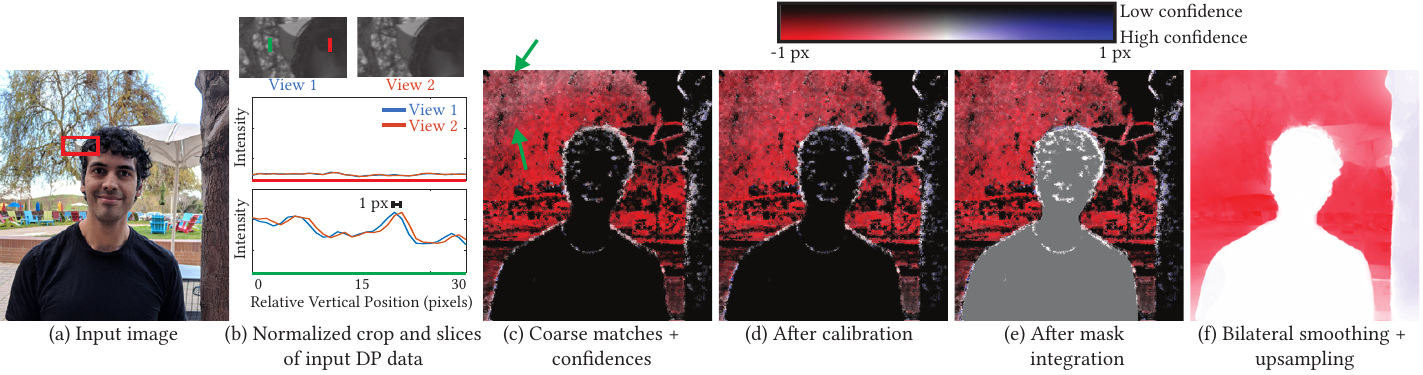}
    \vspace{-18pt}
    \caption{The inputs to and steps of our disparity algorithm. Our input data is a color image (a) and two single-channel DP views that sum to the green channel of the input image. For the purposes of visualization, we normalize the DP data by making the two views have the same local mean and standard deviation. We show pixel intensity vs. vertical position for the two views at two locations marked by the green and red lines in the crops (b). We compute noisy matches and a heuristic confidence (c). Errors due to the lens aberration (highlighted with the green arrows in (c)) are corrected with calibration (d). The segmentation mask is used to assign the disparity of the subject's eyes and mouth to the textureless regions on the subject's shirt (e). We use bilateral space techniques to convert noisy disparities and confidences to an edge-aware dense disparity map (f).}
    \label{fig:disparity_computation}
    \vspace{-6pt}
\end{figure*}

DP sensors effectively create a crude, two-view light field \cite{levoy1996light,gortler1996lumigraph} with a baseline the size of the mobile camera's aperture (${\sim}1$ mm).
It is possible to produce a synthetically defocused image by shearing and integrating a light field that has a sufficient number of views, e.g., one from a Lytro camera~\cite{ng2005light}. However, this technique would not work well for DP data because there are only two samples per pixel and the synthetic aperture size would be limited to the size of the physical aperture. There are also techniques to compute depth from light fields~\cite{adelson1992single,tao2013depth,jeon2015depth}, but these also typically expect more than two views.

Given the two views of our DP sensor, using stereo techniques to compute disparity is a plausible approach. Depth estimation from stereo has been the subject of extensive work (well-surveyed in \cite{Scharstein2002}). Effective techniques exist for producing detailed depth maps from high-resolution image pairs \cite{Sinha2014} and there are even methods that use images from narrow-baseline stereo cameras \cite{micro-baseline-stereo,Yu2014}. However, recent work suggests that standard stereo techniques are prohibitively expensive to run on mobile platforms and often produce artifacts when used for synthetic defocus due to poorly localized edges in their output depth maps \cite{barron2015fast}. We therefore build upon the stereo work of Barron \etal~\shortcite{barron2015fast} and the edge-aware flow work of Anderson~\etal ~\shortcite{anderson2016jump} to construct a stereo algorithm that is both tractable at high resolution and well-suited to the defocus task by virtue of following the edges in the input image.
% and structure from motion \cite{Triggs1999}.

There are several key differences between DP data and stereo pairs from standard cameras. Because the data is coming from a single sensor, the two views have the same exposure and white balance and are perfectly synchronized in time, making them robust to camera and scene motion. In addition, standard stereo rectification is not necessary because the horizontally split pixels are designed to produce purely horizontal disparity in the sensor's reference frame. The baseline between DP views is much smaller than most stereo cameras, which has some benefits: computing correspondence rarely suffers due to occlusion and the search range of possible disparities is small, only a few pixels. However, this small baseline also means that we must compute disparity estimates with sub-pixel precision, relying on fine-image detail that can get lost in image noise, especially in low-light scenes.  While traditional stereo calibration is not needed, the relationship between disparity and depth is affected by lens aberrations and variations in the position of the split between the two halves of each pixel due to lithographic errors. %Finally, because the two DP views are taken through two halves of the same aperture, disparity is exactly proportional to optical defocus. As a result, depth-from-stereo and depth-from-defocus should yield identical disparities \cite{schechner2000depth}. That said, stereo matching is a 1D problem, while analyzing defocus blur is 2D, so exploiting defocus may yield additional robustness at additional expense; exploring this path is future work.

Our algorithm for computing a dense depth map under these challenging conditions is well-suited for synthesizing shallow depth-of-field images.
We first temporally denoise a burst of DP images, using the technique of Hasinoff \etal~\shortcite{hasinoff2016burst}. We then compute correspondences between the two DP views using an extension of Anderson \etal~\shortcite{anderson2016jump} (Fig.~\ref{fig:disparity_computation}(c)). We adjust these disparity values with a spatially varying linear function to correct for lens aberrations, such that any given depth produces the same disparity for all image locations (Fig.~\ref{fig:disparity_computation}(d)). Using the segmentation computed in Sec.~\ref{sec:segmentation}, we flatten disparities within the masked region to bring the entire person in focus and hide errors in disparity (Fig.~\ref{fig:disparity_computation}(e)). Finally, we use bilateral space techniques \cite{kopf2007joint,barron2016fast} to obtain smooth, edge-aware disparity estimates that are suitable for defocus rendering (Fig.~\ref{fig:disparity_computation}(f)).

%We first describe our extension of Anderson \etal~\shortcite{anderson2016jump} to compute correspondences between the two DP views. We then describe the imaging model that relates disparity to depth and provide a calibration procedure to parts of the scene that are at the same depth have the same disparity. We describe how we use the segmentation mask described in the previous network to improve the disparities and then describe an edge-aware smoothing of the disparities, so that they produce a disparity map suitable for synthetic defocus.

%\neal{We may want to mention \cite{jang2016depth}, which proposes something very similar to depth from DP.}

% We compute depth from these split-pixels by computing the disparity between the left and right half images. Note that in-focus points have zero-disparity and therefore the mapping from disparity to depth depends on focus. In addition, the baseline between the two images varies across the field-of-view because of lens aberrations such as field curvature.

\subsection{Computing Disparity}

%Our algorithm to compute noisy disparities consists of two parts. First, we temporally denoise a burst of DP frames using techniques described in Hasinoff \etal~\shortcite{hasinoff2016burst}. Second, we compute correspondences between the two views with an extension of the flow algorithm described in Anderson \etal~ \shortcite{anderson2016jump}.

To get multiple frames for denoising, we keep a circular buffer of the last nine raw and DP frames captured by the camera. When the shutter is pressed, we select a base frame close in time to the shutter press.  We then align the other frames to this base frame and robustly average them using techniques from Hasinoff \etal~\shortcite{hasinoff2016burst}. Like Hasinoff \etal~\shortcite{hasinoff2016burst}, we treat the non-demosaiced raw frames as a four channel image at Bayer plane resolution. The sensor we use downsamples green DP data by $2\times$ horizontally and $4\times$  vertically. Each pixel is split horizontally. That is, a full resolution image of size $2688\times 2016$ has Bayer planes of size $1344 \times 1008$ and DP data of size $1344 \times 504$. We linearly upsample the DP data to Bayer plane resolution and then append it to the four channel Bayer raw image as its fifth and sixth channel.  The alignment and robust averaging is applied to this six channel image. This ensures that the same alignment and averaging is applied to the two DP views. 

\begin{figure}[t]
    \centering
    \includegraphics[width=\columnwidth]{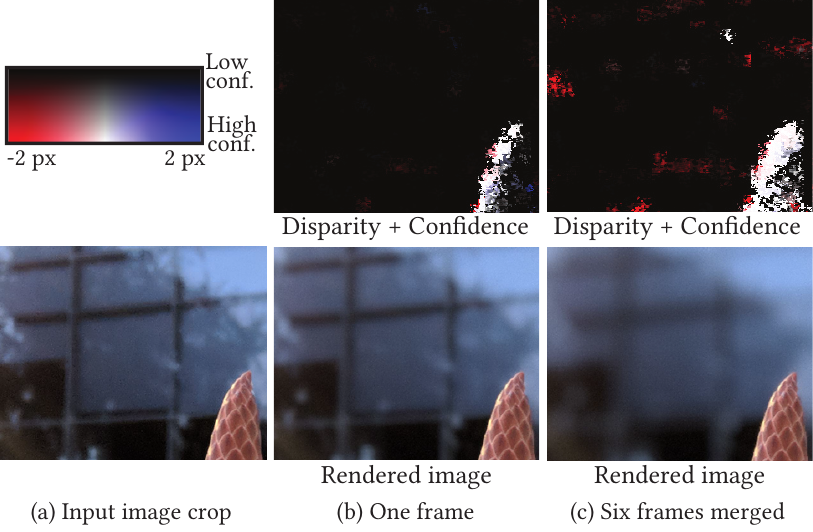}
    \caption{Denoising a burst of DP frames prior to disparity computation improves results in low-light scenes. A crop of a picture taken at dusk (a). The background's disparity is not recoverable from a single frame's DP data, so the background doesn't get blurred (b). When a burst of frames are merged, the SNR is high enough to determine that the background is further than the flower. The background is therefore blurred (c).}
    \label{fig:align_merge}
    \vspace{-12pt}
\end{figure}

This alignment and averaging significantly reduces the noise in the input DP frames and increases the quality of the rendered results, especially in low-light scenes. For example, in an image of a flower taken at dusk (Fig.~\ref{fig:align_merge}), the background behind the flower is too noisy to recover meaningful disparity from a single frame's DP data. However, if we align and robustly average six frames (Fig.~\ref{fig:align_merge}(c)), we are able to recover meaningful disparity values in the background and blur it as it is much further away from the camera than the flower. In the supplemental, we describe an experiment that shows disparity values from six aligned and averaged frames are two times less noisy than disparity values from a single frame for low-light scenes (5 lux).

To compute disparity, we take each non-overlapping $8\times 8$ tile in the first view and search a range of $-3$ pixels to $3$ pixels in the second view at DP resolution. For each integer shift, we compute the sum of squared differences (SSD). We find the minimum of these seven points and fit a quadratic to the SSD value at the minimum and its two surrounding points. We use the location of the quadratic's minimum as our sub-pixel minimum. Our technique differs from Anderson \etal~\shortcite{anderson2016jump} in that we perform a small one dimensional brute-force search, while they do a large two dimensional search over hundreds of pixels that is accelerated by the Fast Fourier Transform (FFT) and sliding-window filtering. For our small search size, the brute-force search is faster than using the FFT (4 ms vs 40 ms). 

For each tile we also compute a confidence value based on several heuristics: the value of the SSD loss, the magnitude of the horizontal gradients in the tile, the presence of a close second minimum, and the agreement of disparities in neighboring tiles.
%For each tile, we also compute a heuristic confidence. This confidence uses the value of the SSD loss, whether there is a close second minimum in the loss and whether the tile agrees with at least one of its neighbors. It also uses the value of the horizontal gradient in the image, which is a notable difference between our method and Anderson \etal~\shortcite{anderson2016jump}, which uses the two dimensional tile variance instead. 
Using only horizontal gradients is a notable departure from Anderson \etal~\shortcite{anderson2016jump}, which uses two dimensional tile variances. For our purely horizontal matching problem the vertical gradients are not informative, due to the aperture problem \cite{adelson1985spatiotemporal}. We upsample the per-tile disparities and confidences to a noisy per-pixel disparities and confidences as described in Anderson \etal~\shortcite{anderson2016jump}.

%Once we have computed the disparities, we apply a calibration to handle lens aberrations (Section~\ref{sec:calibration}) and if the image is of people, we merge the disparities and confidences with the segmentation mask (Section~\ref{sec:merge_mask}).

\subsection{Imaging Model and Calibration}

\label{sec:calibration}

\begin{figure}[t]
    \centering
    \subfigure[Without calibration]{\label{fig:no_calibration_render}\includegraphics[width=0.49\columnwidth]{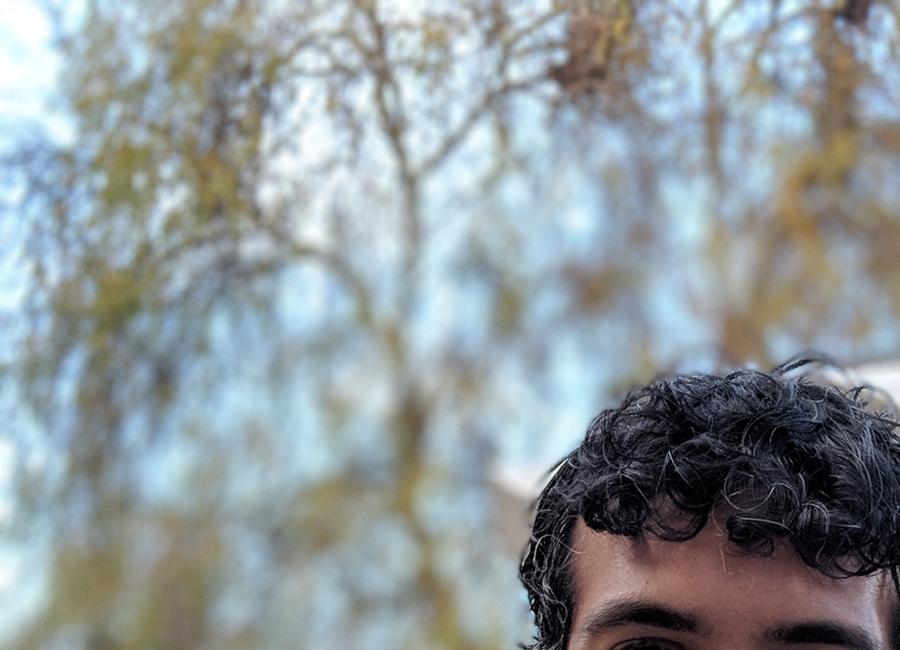}}
    \subfigure[With calibration]{\label{fig:calibration_render}\includegraphics[width=0.49\columnwidth]{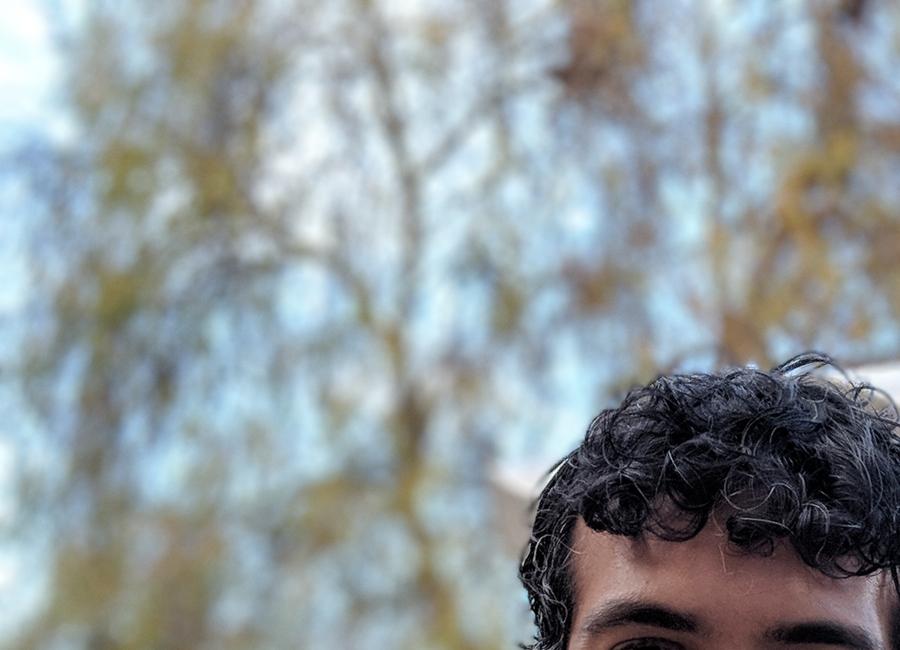}}
    \vspace{-12pt}
    \caption{Synthetic shallow depth-of-field renderings of Fig.~\ref{fig:disparity_computation}(a) without and with calibration. Notice the uneven blur and sharp background in the top left of (a) that is not present in (b).}
    \vspace{-12pt}
\end{figure}

Objects at the same depth, but different spatial locations can have different disparities due to lens aberrations and sensor defects. Uncorrected, this can cause artifacts, such as parts of the background in a synthetically defocused image remaining sharp (Fig.~\ref{fig:no_calibration_render}). We correct for this by applying a calibration procedure (Fig.~\ref{fig:calibration_render}).

% Disparity is proportional to blur size.
To understand how disparity is related to depth, consider imaging an out of focus point light source (Fig.~\ref{fig:pd_light_field}). The light passes through the main lens and focuses in front of the sensor, resulting in an out-of-focus image on the sensor. Light that passes through the left half of the main lens aperture hits the microlens at an angle such that it is directed into the right half-pixel. The same applies to the right half of the aperture and the left half-pixel.
The two images created by the split pixels have viewpoints that are roughly in the centers of these halves, giving a baseline proportional to the diameter of the aperture $L$ and creating disparity $d$ that is proportional to the blur size $b$ (Fig.~\ref{fig:pd_light_field}(b)). That is, there is some $\alpha$ such that $d=\alpha \bar b$, where $\bar b$ is a signed blur size that is positive if the focal plane is in front of the sensor and negative otherwise. 

If we assume the paraxial and thin-lens approximations, there is a straight-forward relationship between signed blur size $\bar b$ and depth $D$. It implies that
\begin{equation} 
d = \alpha \bar b =  \alpha L f\left(\frac{1}{z}-\frac{1}{D}\right)
\label{eq:depth_to_disparity}
\end{equation}
where $z$ is focus distance and $f$ is focal length (details in the supplemental). This equation has two notable consequences. First, disparity depends on focus distance ($z$) and is zero when depth is equal to focus distance ($D=z$). Second, there is a linear relationship between inverse depth and disparity that does not vary spatially. 

\begin{figure}[t]
\centering
\includegraphics[width=\columnwidth]{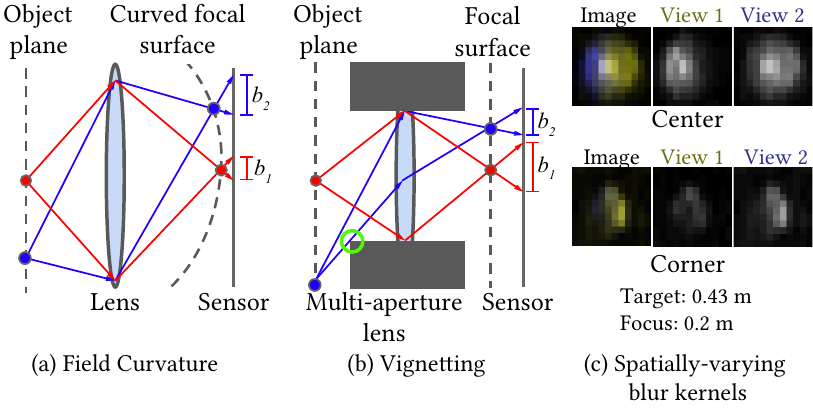}
\caption{Calibration of aberrations, while important for any stereo system, is critical for dual-pixel systems. These aberrations cause blur size $b$ and therefore disparity $d$ to vary spatially for constant depth objects. E.g., field curvature can increase peripheral blur size $b_2$ compared to central blur size $b_1$ (a). In (b), the green circled ray represents the most oblique ray that survives vignetting, thereby limiting peripheral blur size $b_2$.  Real linear-space blur kernels from the center and corner of images captured by a dual-pixel camera (c). The image blur kernel is the sum of the two DP views and is color-coded accordingly.}
\label{fig:calibration_causes}
\vspace{-6pt}
\end{figure}

However, real mobile camera lenses can deviate significantly from the paraxial and thin-lens approximations. This means Eq.~\ref{eq:depth_to_disparity} is only true at the center of the field-of-view. Optical aberrations at the periphery can affect blur size significantly. For example, field curvature is an aberration where a constant depth in the scene focuses to a curved surface behind the lens, resulting in a blur size that varies across the flat sensor (Fig.~\ref{fig:calibration_causes}(a)). 
Optical vignetting blocks some of the light from off-axis points, reducing their blur size (Fig.~\ref{fig:calibration_causes}(b)). In addition to optical aberrations, the exact location of the split between the pixels may vary due to lithographic errors. In Fig.~\ref{fig:calibration_causes}(c), we show optical blur kernels of the views at the center and corner of the frame, which vary significantly. 

\begin{figure}[t]
\centering
\includegraphics[width=\columnwidth]{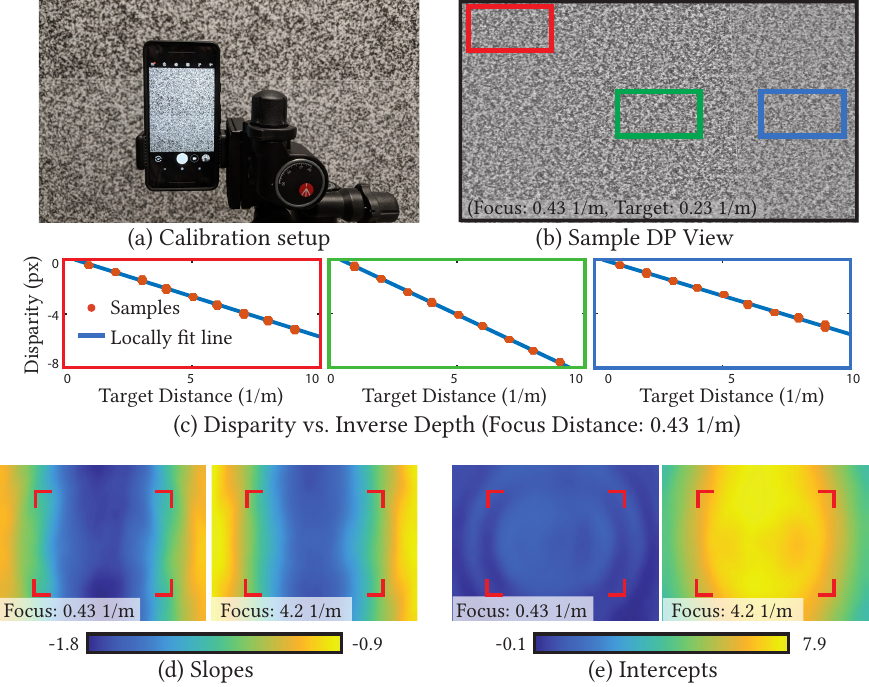}
%\vspace{-12pt}
\caption{Results from our calibration procedure. We use a mobile camera to capture images of a fronto-parallel textured target sweeping through all focus and target distances (a). One such image (b). Disparity vs. target distance and best-fit lines for several spatial locations for one focus setting (c). All disparities are negative for this focus distance of ~2.3m, which is past the farthest target. The largest magnitude disparity of ~8 pixels is an extreme case of a target at 10cm. The best fit slopes (d) and intercepts (e) for linear functions mapping inverse target distance to disparity for every spatial location and two focus distances. The 1.5x crop is marked by the red rectangle corners in each plot.}
\label{fig:calibration_linear}
\vspace{-8pt}
\end{figure}

To calibrate for variations in blur size, we place a mobile phone camera on a tripod in front of a textured fronto-parallel planar test target (Fig.~\ref{fig:calibration_linear}(a-b)) that is at a known constant depth. We capture images spanning a range of focus distances and target distances. We compute disparity on the resulting DP data. For a single focus-distance, we plot the disparities versus inverse depth for several regions in the image (Fig.~\ref{fig:calibration_linear}(c)). We empirically observe that the relationship between disparity and inverse depth is linear, as predicted by Eq.~\ref{eq:depth_to_disparity}. However, the slope and intercept of the line varies spatially. We denote them as $S_z(\mathbf{x})$ and $I_z(\mathbf{x})$ and use least squares to fit them to the data (solid line in Fig.~\ref{fig:calibration_linear}(c)). We show these values for every pixel at several focus distances (Fig.~\ref{fig:calibration_linear}(d-e)). Note that the slopes are roughly constant across focus distances ($z$), while the intercepts vary more strongly. This agrees with the thin-lens model's theoretically predicted slope ($-\alpha L f$), which is independent of focus distance.

To correct peripheral disparities, we use $S_z(\mathbf{x})$ and $I_z(\mathbf{x})$ to solve for inverse depth. Then we apply  $S_z(\mathbf{0})$ and $I_z(\mathbf{0})$, where $\mathbf{0}$ is the image center coordinates. This results in the corrected disparity
\begin{equation}
    d_{\mathit{corrected}}(\mathbf{x}) = I_z(\mathbf{0}) + \frac{S_z(\mathbf{0})(d(\mathbf{x})-I_z(\mathbf{x}))}{S_z(\mathbf{x})}.
\end{equation}
Since focus distance $z$ varies continuously, we calibrate 20 different focus distance and linearly interpolate $S_z$ and $I_z$ between them.

\subsection{Combining Disparity and Segmentation}
\label{sec:merge_mask}

For photos containing people, we combine the corrected disparity with the segmentation mask (Sec.~\ref{sec:segmentation}).
The goal is to bring the entire subject in to focus while still blurring out the background. This requires significant expertise with a DSLR and we wish to make it easy for consumers to do.

Intuitively, we want to map the entire person to a narrow disparity range that will be kept in focus using the segmentation mask as a cue.
To accomplish this, we first compute the weighted average disparity, $d_\mathit{face}$ over the largest face rectangle with confidences as weights.
Then, we set the disparities of all pixels in the \emph{interior} of the person-segmentation mask to $d_\mathit{face}$, while also increasing their confidences. The \emph{interior} is defined as pixels where the CNN output $M_c(\mathbf{x}) > 0.94$. The CNN output is bilinearly upsampled to be at the same resolution as the disparity. We choose a conservative threshold of $0.94$ to avoid including the background as part of the interior and rely on the bilateral smoothing (Sec.~\ref{sec:bilateral_smoothing}) to snap the in-focus region to the subject's boundaries (Fig.~\ref{fig:disparity_computation}).

Our method allows novices to take high quality shallow depth-of-field images where the entire subject is in focus. It also helps hide errors in the computed disparity that can be particularly objectionable when they occur on the subject, e.g., a blurred facial feature.
A possible but rare side effect of our approach is that two people at very different depths will both appear to be in focus, which may look unnatural.
While we do not have a segmentation mask for photos without people, we try to bring the entire subject into focus by other methods (Sec.~\ref{sec:disparity_to_blur_strength}).
\subsection{Edge-Aware Filtering of Disparity}
\label{sec:bilateral_smoothing}

We use the bilateral solver \cite{barron2016fast} to turn the noisy disparities into a smooth edge-aware disparity map suitable for shallow depth-of-field rendering. This is similar to our use of the solver to smooth the segmentation mask (Sec.~\ref{sec:edge_aware_filtering_of_masks}). The bilateral solver produces the smoothest edge-aware output that resembles the input wherever confidence is large, which allows for confident disparity estimates to propagate across the large low-confidence regions that are common in our use case (e.g., the interior of the tree in Fig.~\ref{fig:disparity_computation}(e)).
As in Sec.~\ref{sec:edge_aware_filtering_of_masks}, we apply the solver at one-quarter image resolution. We apply a $3\times3$ median filter to the smoothed disparities to remove speckling artifacts. Then, we use joint bilateral upsampling \cite{kopf2007joint} to upsample the smooth disparity to full resolution (Fig.~\ref{fig:disparity_computation}(f)).

\section{Rendering}
\label{sec:rendering}

%The input to our rendering stage is the final smoothed disparity computed in Sec.~\ref{sec:bilateral_smoothing} and the unblurred linear luminance RGB image. 
The input to our rendering stage is the final smoothed disparity computed in Sec.~\ref{sec:bilateral_smoothing} and the unblurred input image, represented in a linear color space, i.e., each pixel's value is proportional to the count of photons striking the sensor. Rendering in a linear color space helps preserve highlights in defocused regions.

In theory, using these inputs to produce a shallow depth-of-field output is straightforward---an idealized rendering algorithm falls directly out of the image formation model discussed in Sec.~\ref{sec:calibration}. Specifically, each scene point projects to a translucent disk in the image, with closer points occluding those farther away. Producing a synthetically defocused image, then, can be achieved by sorting pixels by depth, blurring them one at a time, and then accumulating the result with the standard ``over'' compositing operator. 
% \david{see latex source}: Decided to ignore it.
% Reviewer comment: The statement on l.537 that "[In theory,] each scene point projects to a translucent disk in the image, with closer points occluding those farther away." can be mis-leading to the non-expert. It suggests that one could take the input unblurred image and use a splatting technique to create the final blurred image. Experts know that splatting leads to severe aliasing, but this is when it's applied to a digitially rasterized image. The statement does say "scene points" not "image pixels", but it might be worth clarifying.
% Possible responses:
% Note that directly splatting each pixel in the unblurred image onto a translucent disk will not produce good results due to aliasing.
% Note that like all post-filtering techniques, this approach will produce poor results if the input image is aliased, so more care would need to be taken if operating on rasterized synthetic images.
% Ignore it.

To achieve acceptable performance on a mobile device, we need to approximate this ideal rendering algorithm. Practical approximations for specific applications are common in the existing literature on synthetic defocus. The most common approximation is to quantize disparity values to produce a small set of constant-depth sub-images \cite{barron2015fast,kraus2007}. Even Jacobs \etal~\shortcite{jacobs2012}, who use real optical blur from a focal stack, decompose the scene into discrete layers. Systems with real-time requirements will also often use approximations to perfect disk kernels to improve processing speed \cite{lee2009}. Our approach borrows parts of these assumptions and builds on them for our specific case.

Our pipeline begins with precomputing the disk blur kernels needed for each pixel. We then apply the kernels to sub-images covering different ranges of disparity values. Finally, we composite the intermediate results, gamma-correct, and add synthetic noise. We provide details for each stage below.

% Our practical rendering pipeline consists of the following steps:
% \begin{enumerate}
% \item Prepare: Determine the function mapping disparity values to blur radii.
% \item Blur: Segment the image into disparity bands and disk blur each band independently.
% \item Finish: Composite the blurred layers, then finish with gamma-correction and synthetic noise.
% \end{enumerate}

%The input to our rendering stage is the computed disparity map and the RGB image that are used to render the synthetic shallow depth-of-field image. We first compute \emph{blur strength map}, i.e., radius of blur kernel to apply at each pixel based on disparity. Given this blur strength map, we then render the final result using back to front rendering \rahul{Use proper nomenclature and references for shallow \DoF rendering}. 
\begin{figure}[t]%
\centering
\subfigure[][Physically correct mapping]{%
\label{fig:half_dof_mapping_correct}%
\begin{tabular}{@{}c@{}}
\includegraphics[width=0.46\columnwidth]{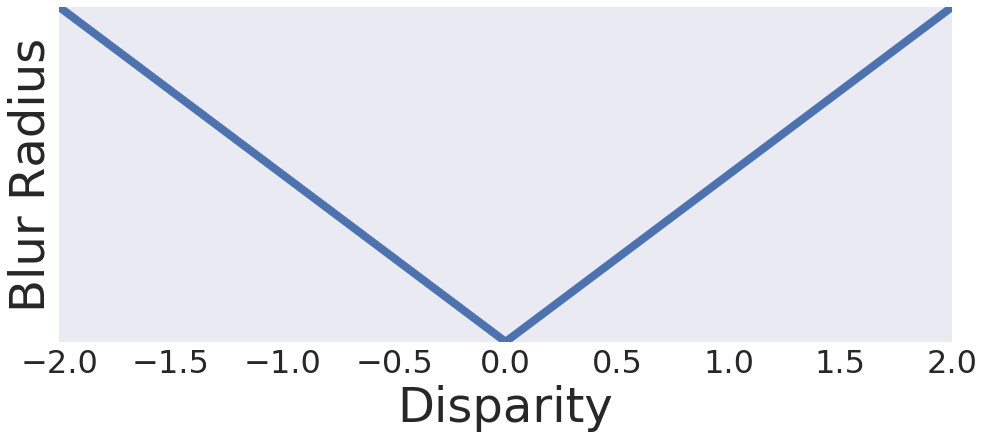} \\
\includegraphics[width=0.46\columnwidth]{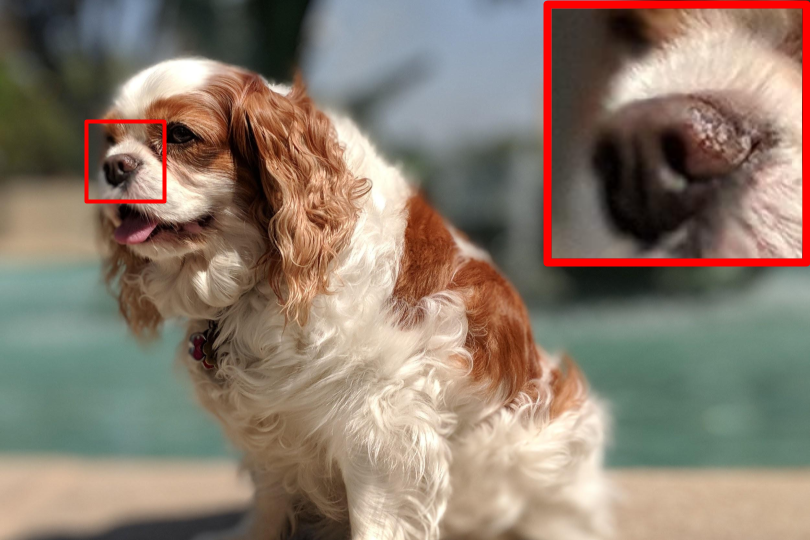}
\end{tabular}
}
\vspace{-0.05in}
\subfigure[][Our mapping]{%
\label{fig:half_dof_mapping_our}%
\begin{tabular}{@{}c@{}}
\includegraphics[width=0.46\columnwidth]{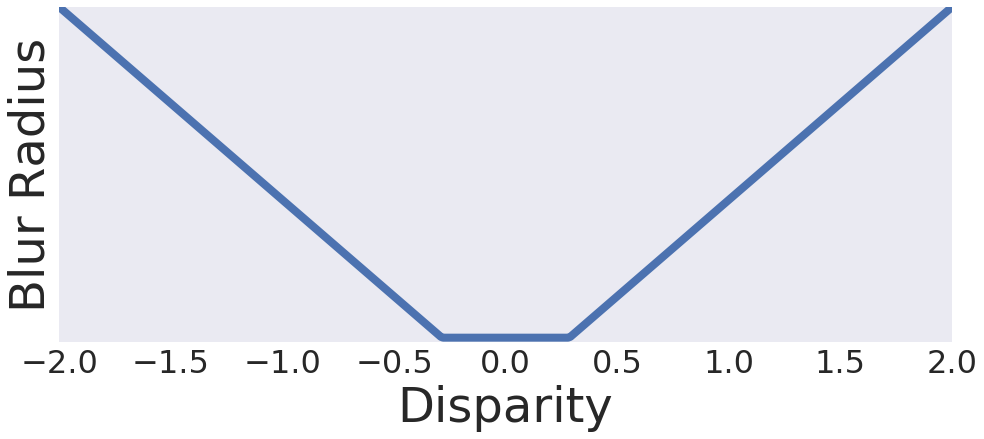} \\
\includegraphics[width=0.46\columnwidth]{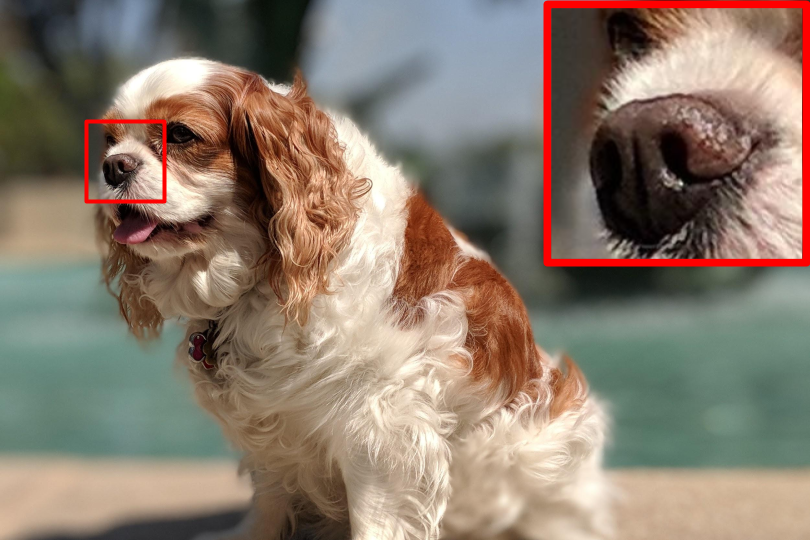}
\end{tabular}
}
\vspace{-0.05in}
\caption[]{
Using a physically correct mapping \subref{fig:half_dof_mapping_correct} keeps the dog's eye in sharp focus but blurs her nose. Our mapping
\subref{fig:half_dof_mapping_our}, which creates an extended region of forced-zero blur radius around the in-focus plane, keeps the entire subject in focus.}
\label{fig:half_dof}%
\vspace{-16pt}
\end{figure}

\subsection{Precomputing the blur parameters}
\label{sec:disparity_to_blur_strength}

As discussed in Sec.~\ref{sec:calibration}, zero disparity corresponds to the focus plane and disparity elsewhere is proportional to the defocus blur size.
In other words, if the photograph is correctly focused on the main subject, then it
already has a defocused background.
However, to produce a shallow depth of
field effect, we need to greatly increase the amount of defocus.  This is
especially true if we wish the effect to be evident and pleasing on the small
screen of a mobile phone.  This implies simulating a larger aperture lens than
the camera's native aperture, hence a larger defocus kernel.  To accomplish
this, we can simply apply  a blur kernel with radius proportional to the computed disparity (Fig.~\ref{fig:half_dof_mapping_correct}).
In practice, the camera may not be
accurately focused on the main subject.  Moreover, if we simulate an aperture
as large as that of an SLR, parts of the subject that were in adequate focus in
the original picture may go out of focus.  Expert users know how to control
such shallow depths-of-field, but novice users do not.  To address both of
these problems, we modify our procedure as follows.

We first compute the disparity to focus at, $d_{\mathit{focus}}$.
%If faces are detected $d_{\mathit{focus}}$ is set to the median disparity over the largest face rectangle, unless a user touch is detected in which case $d_{\mathit{focus}}$ is set to the median disparity over the touch rectangle.
We set it to the median disparity over a subject region of interest.
The region is the largest detected face output by the face detector that is well-focused, i.e., its disparity is close to zero.
In the absence of a usable face, we use a region denoted by the user tapping-to-focus during view-finding.
If no cues exist, we trust the auto-focus and set $d_{\mathit{focus}}=0$.
%If no faces are found and no touch is detected, we set $d_{\mathit{focus}}=0$.
%We would like to transform the disparity map as if the camera had focused at the plane corresponding to $d_{\mathit{focus}}$.
% dejacobs: commenting this out to simplify the story a bit.
% We subtract $d_{\mathit{focus}}$ from the computed disparity map to effectively ``refocus'' it.
% Afterwards, we apply a scale factor $\kappa(\cdot)$ that increases the defocus in proportion to the \emph{refocused} disparity and normalizes the changes in disparity due to focus distance $f$ to yield a pleasing amount of defocus blur in different cases.
% E.g., disparity ranges are smaller for a larger focus distances and require greater scaling.

%and then multiplying that shifted disparity map by a factor $\kappa_1(d_{\mathit{focus}})$.
%= \frac{1}{1 + \nicefrac{d_{\mathit{focus}}}{2}}|_{\left[0.5, 2\right]}$ where $(\cdot)|_{\left[a,b\right]}$ denotes clamping to the range $\left[a,b\right]$.
%Disparity ranges are smaller for larger focus distances and vice versa. In order to normalize for this change, we introduce multiplication by a focus %distance dependent factor $\kappa_2(f)$
%= (0.33f + 0.17)|_{\left[1, 3.5\right]}$ where $f$ is in meters.

To make it easier for novices to make good shallow-depth-of-field pictures, we
artificially keep disparities within $d_\emptyset$ disparity units of $d_\mathit{focus}$
%scaled and shifted disparities 
sharp by mapping them to zero blur radius, i.e., we leave pixels with disparities in $[d_\mathit{focus}-d_\emptyset,d_\mathit{focus}+d_\emptyset]$ unblurred. Combining, we compute blur radius as:
\begin{equation}
\label{eqn:blur_radius}
%r(\mathbf{x}) = \max \left(0, \kappa(f)  \left| d(\mathbf{x}) - d_{\mathit{focus}} \right| - d_\emptyset \right)
r(\mathbf{x}) = \kappa(z) \max \left(0,   \left| d(\mathbf{x}) - d_{\mathit{focus}} \right| - d_\emptyset \right),
\end{equation}
where $\kappa(z)$ controls the overall strength of the blur as a function of focus distance $z$---larger focus distance scenes have smaller disparity ranges, so we increase $\kappa(\cdot)$ to compensate. Fig.~\ref{fig:half_dof_mapping_our} depicts the typical shape of this mapping.
Because rendering large blur radii is computationally expensive, we cap $r(\mathbf{x})$ to $r_\mathit{max}=30$ pixels.

\newcommand{\leafheight}{0.96in}
\newcommand{\petalheight}{0.85in}
\newcommand{\halotablegap}{0.01in}
\begin{figure}[t]
    \centering
    \begin{tabular}{cc@{}c@{}c}
    & \multicolumn{2}{c}{Blurred result, $B$} & $|I - B|$ \\
    \rotatebox[origin=l]{90}{\parbox{\petalheight}{\centering Proposed}} & 
    \includegraphics[height=\petalheight]{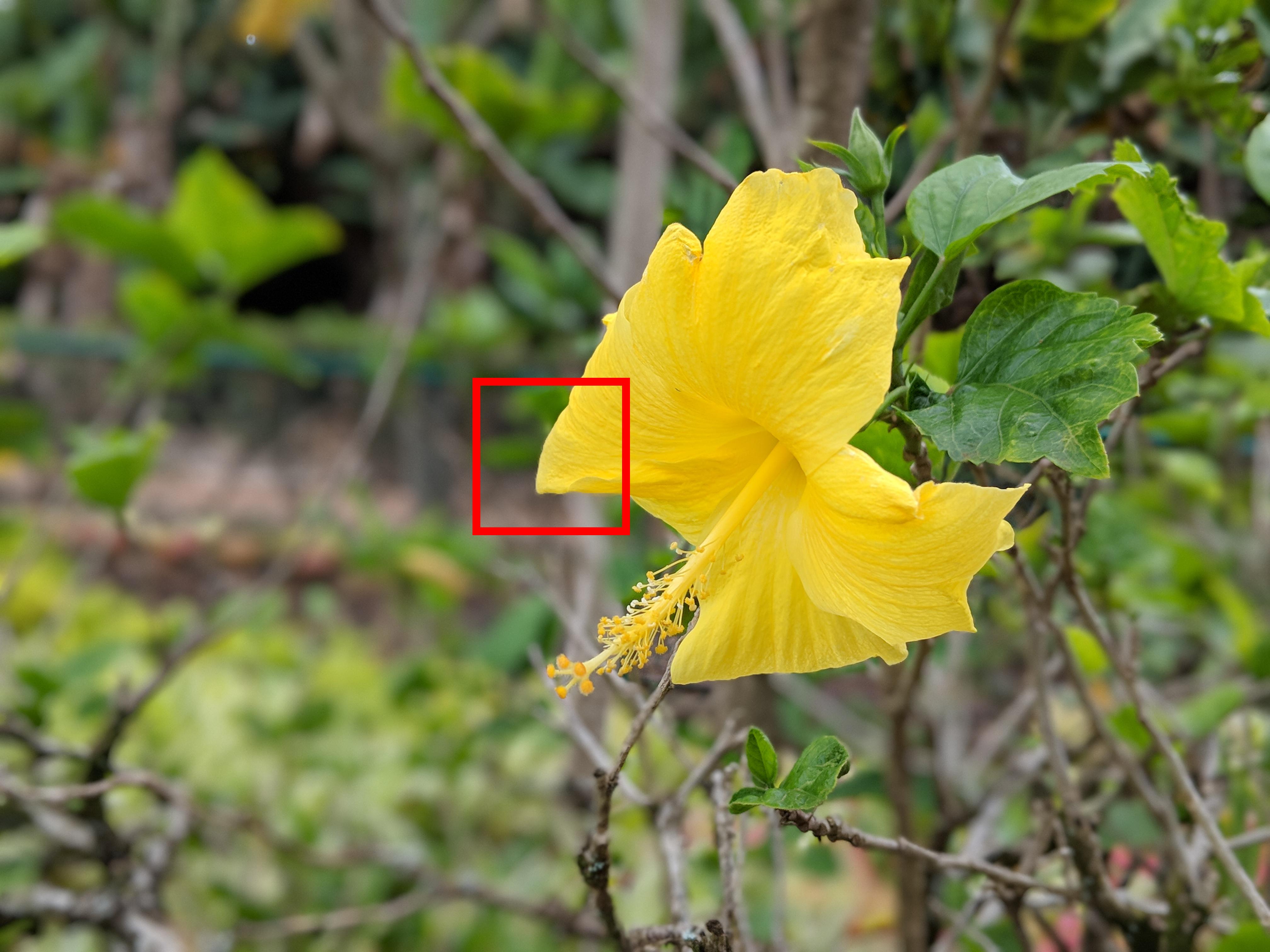} &
    \hspace{\halotablegap}
    \includegraphics[height=\petalheight]{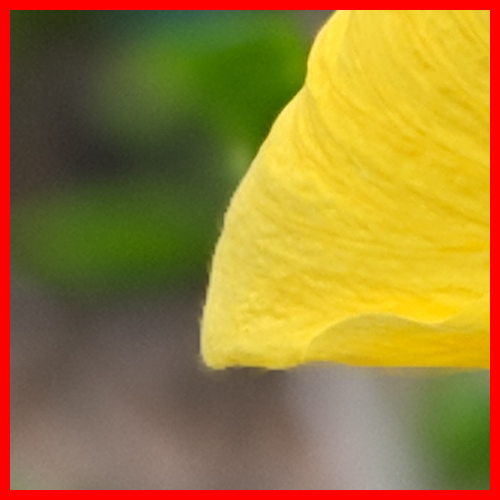} &
    \hspace{\halotablegap}
    \includegraphics[height=\petalheight]{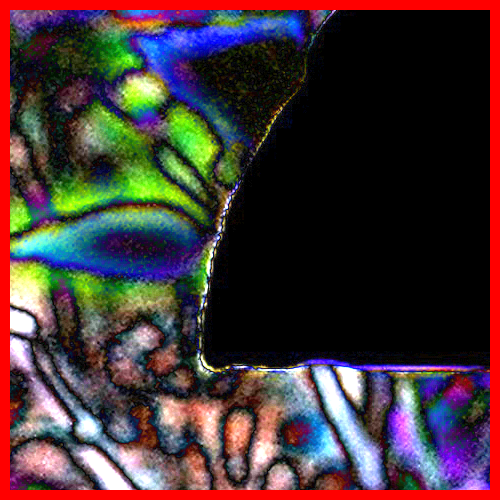} \\
   \rotatebox[origin=l]{90}{\parbox{\petalheight}{\centering Na\"ive
   gather}} & 
    \includegraphics[height=\petalheight]{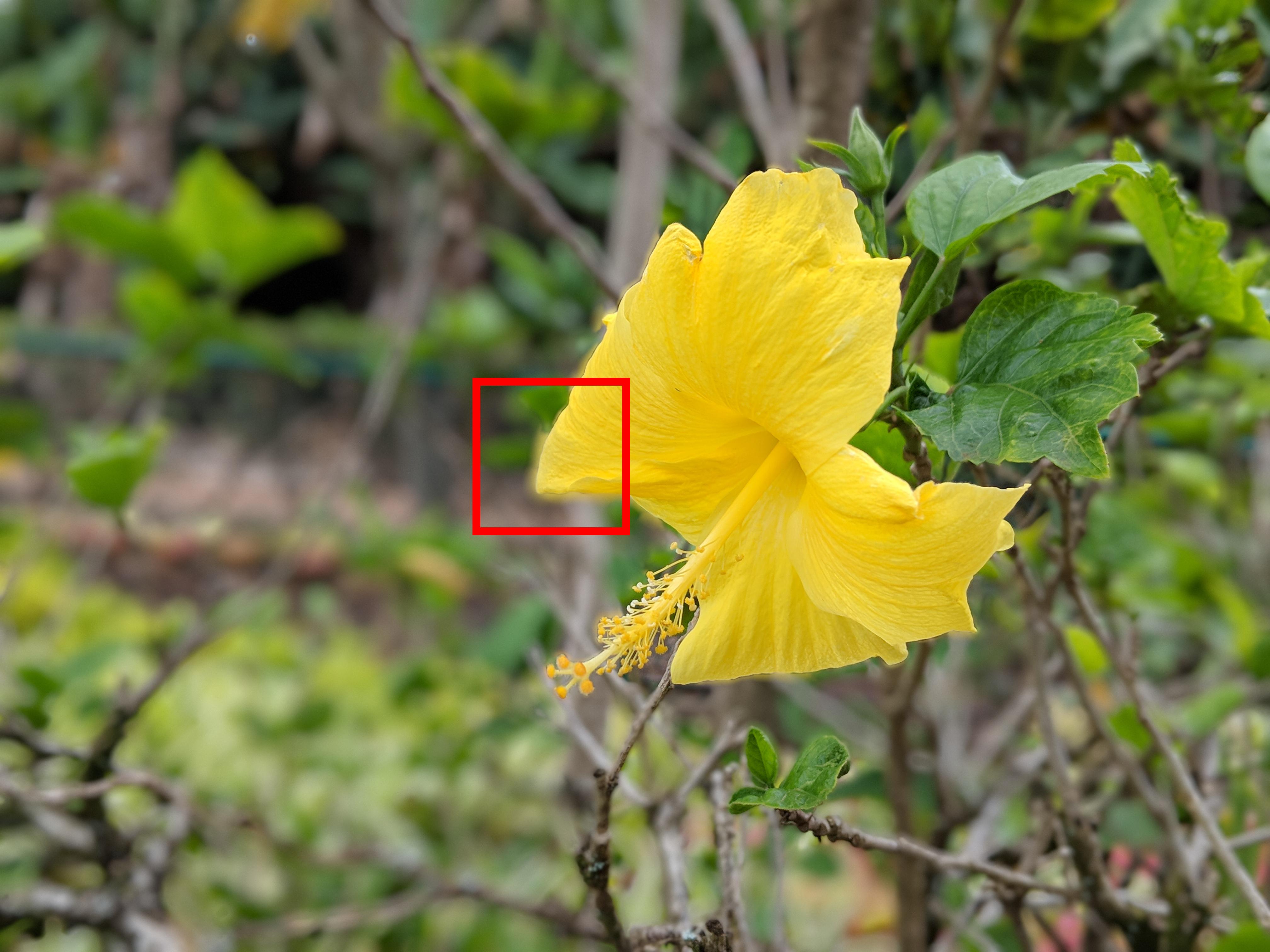} &
    \hspace{\halotablegap}
    \includegraphics[height=\petalheight]{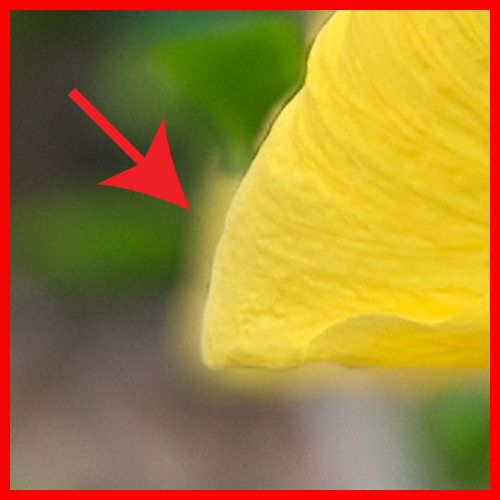} &
    \hspace{\halotablegap}
    \includegraphics[height=\petalheight]{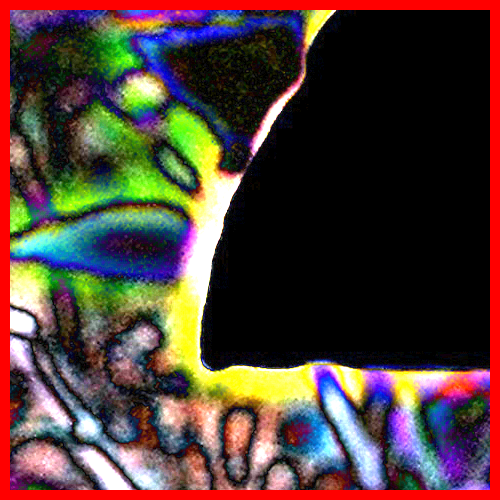} \\
    \rotatebox[origin=l]{90}{\parbox{\petalheight}{\centering Single pass}} &    \includegraphics[height=\petalheight]{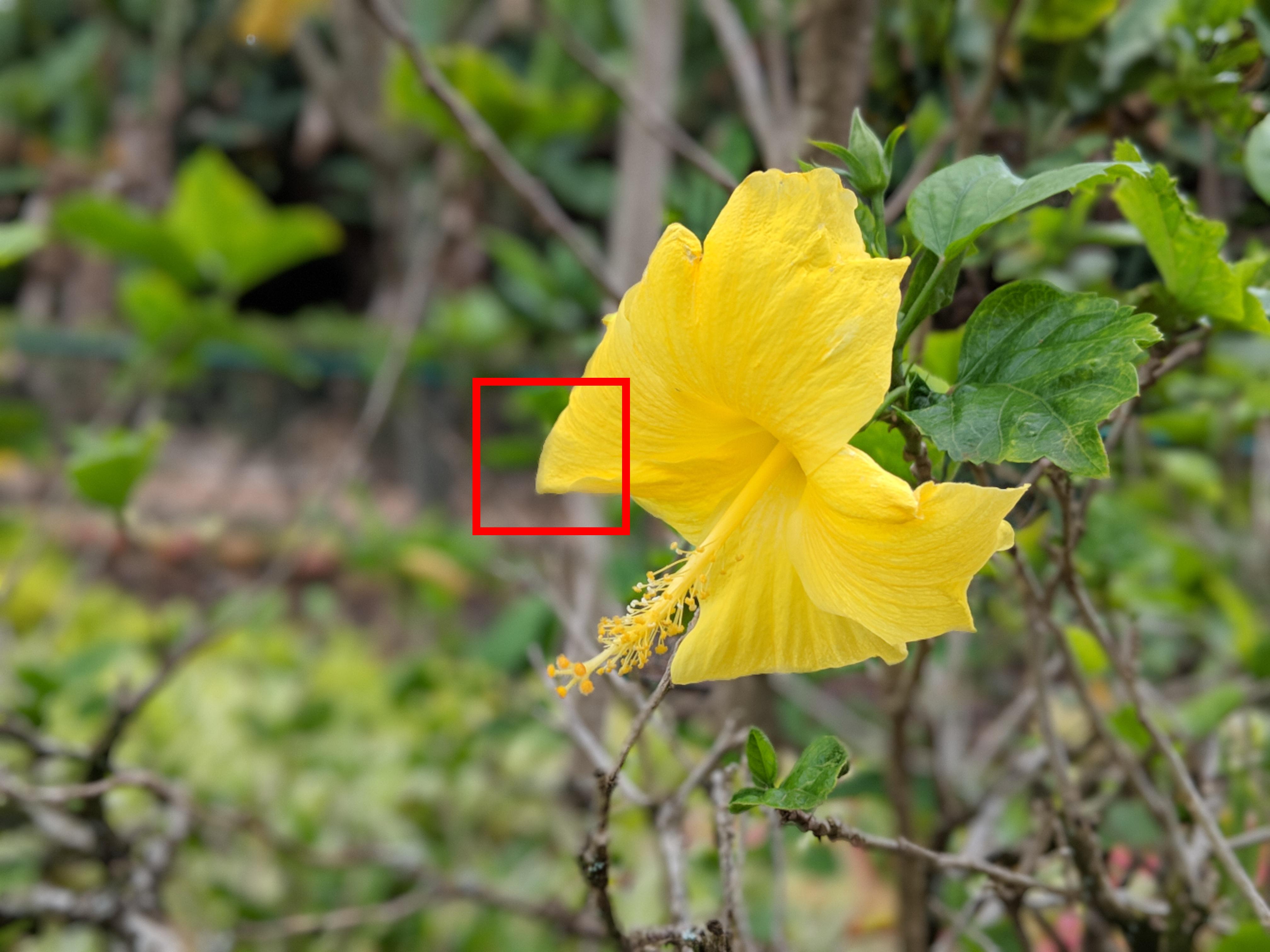} &
    \hspace{\halotablegap}
    \includegraphics[height=\petalheight]{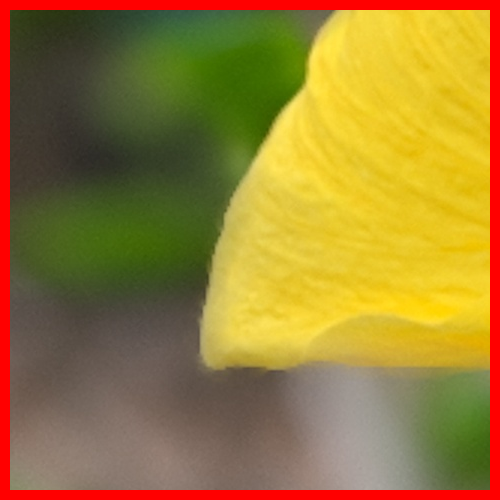} &
    \hspace{\halotablegap}
    \includegraphics[height=\petalheight]{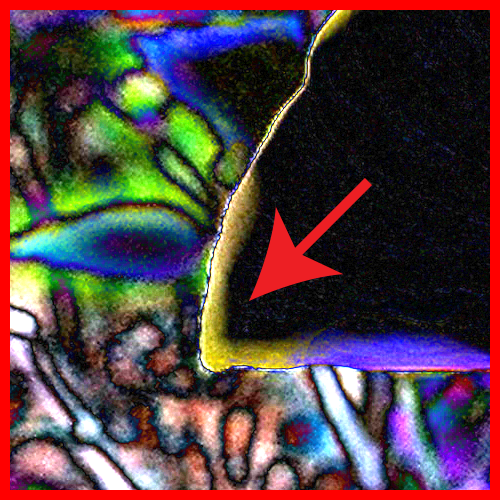} \\
    \end{tabular}
    \caption{Color halos. We illustrate how our proposed method (top row) prevents different types of color halos by selectively removing components of our approach. The right column shows the image difference between each technique's result and the unblurred input image. A na\"ive gather-based blur (middle row) causes the foreground to bleed into the background---note the yellow halo. Blurring the image in a single pass (bottom row) causes the defocused background to blur over the foreground---note the darkened petal edge, most clearly visible in the difference inset.}
    \label{fig:blur_examples}
    \vspace{-6pt}
\end{figure}
\subsection{Applying the blur}
\label{sec:rendering_blur}

%Now that we know how much we want to defocus each pixel, we need to compute the actual blurred image. In designing our approach, there are two visual properties that we want to maintain while achieving fast processing times: disk-shaped blur kernels and a consistent depth relationships between parts of the scene.

When performing the blur, we are faced with a design decision concerning the shape and visual quality of the defocused portions of the scene. This property is known as \emph{bokeh} in the photography community, and is a topic of endless debate about what makes for a good bokeh shape. When produced optically, bokeh is primarily determined by the shape of the aperture in the camera's lens---e.g., a six-bladed aperture would produce a hexagonal bokeh. In our application, however, bokeh is a choice. We choose to simulate an ideal circular bokeh, as it is simple and produces pleasing results.

Efficiently creating a perfect circular bokeh with a disk blur is difficult because of a mismatch between defocus optics and fast convolution techniques. The optics of defocus blur are most easily expressed as \emph{scatter} operations--each pixel of the input scatters its influence onto a disk of pixels around it in the output. Most convolution algorithms, however, are designed around \emph{gather} operations---each pixel of the output gathers influence from nearby pixels in the input. Gather operations are preferred because they are easier to parallelize. If the blur kernel is constant across the image, the two approaches are equivalent, but when the kernel is spatially varying, as it is in depth-dependent blur, naive convolution implementations can create unacceptable visual artifacts, as shown in Fig.~\ref{fig:blur_examples}.

One obvious solution is to simply reexpress the scatter as a gather. For example, a typical convolution approach defines a filtered image $B$ from an input image $I$ and a kernel $K$ as
\begin{equation}
\label{eqn:simple_convolution}
B_\mathit{gather}(\mathbf{x}) = \sum_{\Delta \mathbf{x}}I(\mathbf{x}+\Delta \mathbf{x})K(\Delta \mathbf{x}).
\end{equation}
If we set the kernel to also be a function of the pixel it is sampling, we can express a scatter indirectly:
\begin{equation}
\label{eqn:scatter_as_gather}
B_\mathit{scatter}(\mathbf{x}) = \sum_{\Delta\mathbf{x}}I(\mathbf{x}+\Delta\mathbf{x})K_{\mathbf{x}+\Delta\mathbf{x}}(-\Delta \mathbf{x}).
\end{equation}
The problem with such an approach is that the range of values that $\Delta\mathbf{x}$ must span is the maximum possible kernel size, since we have to iterate over all pixels that could possibly have non-zero weight in the summation. If all the blur kernels are small, this is a reasonable solution, but if any of the blur kernels are large (as is the case for synthetic defocus), this can be prohibitively expensive to compute.

For large blurs, we instead use a technique inspired by summed area tables and the two observations of disk kernels illustrated in Fig.~\ref{fig:blur_kernel_approximations}: 1) large, anti-aliased disks are well-approximated by rasterized, discrete disks, and 2) discrete disks are mostly constant in value, which means they also have a sparse gradient in $y$ (or, without loss of generality, $x$). The sparsity of the $y$-gradient is useful because it means we can perform the blur in the gradient domain with fewer operations. Specifically, each input pixel scatters influence along a circle rather than the solid disk, and that influence consists of a positive or negative gradient, as shown in the rightmost image of Fig.~\ref{fig:blur_kernel_approximations}. For a $w \times h$ pixel image, this accelerates the blur from $\mathcal{O}(w \times h \times r_\mathit{max}^2)$ to $\mathcal{O}(w \times h \times r_\mathit{max})$. % Using mathcal{O} because O is already used in the math.
Once the scatter is complete, we integrate the gradient image along $y$ to produce the blurred result.

Another way we increase speed is to perform the blur at reduced resolution. Before we process the blurs, we downsample the image by a factor of $2\times$ in each dimension, giving an $8\times$ speedup (blur radius is also halved to produce the equivalent appearance) with little perceptible degradation of visual quality. We later upsample back to full resolution during the finish stage described in Sec.~\ref{sec:rendering_finish}.

\newcommand{\diskwidth}{0.22\columnwidth}
\newcommand{\diskspace}{0.01in}
\begin{figure}[t!]
    \centering
    \begin{tabular}{c@{}c@{}c@{}c}
    \parbox[b]{\diskwidth}{\centering Distance} & \parbox[b]{\diskwidth}{\centering Ideal Kernel} & \parbox[b]{\diskwidth}{\centering Discretized Kernel} & \parbox[b]{\diskwidth}{\centering Derivative\\
    in $y$} \vspace{0.025in} \\
        \includegraphics[width=\diskwidth]{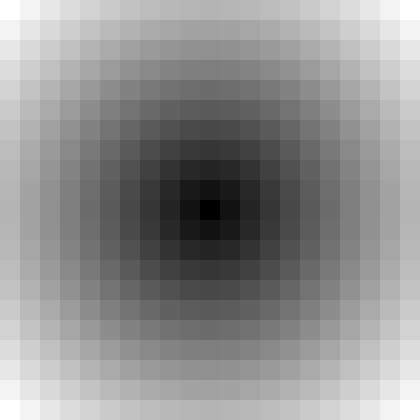} &
        \hspace{\diskspace}
        \includegraphics[width=\diskwidth]{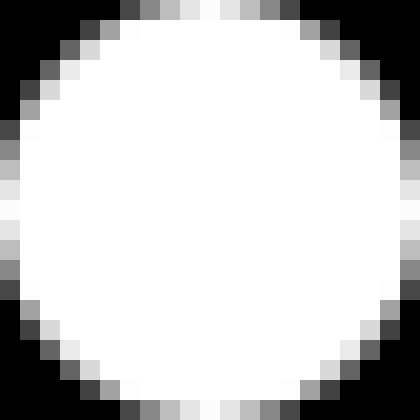} &
        \hspace{\diskspace}
        \includegraphics[width=\diskwidth]{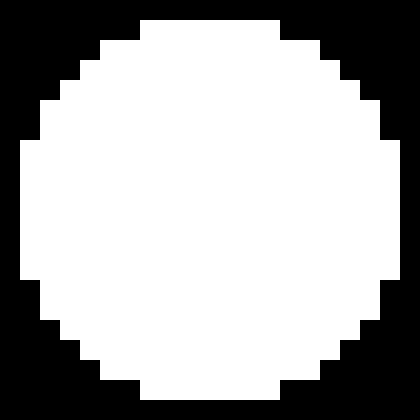} &
        \hspace{\diskspace}
        \includegraphics[width=\diskwidth]{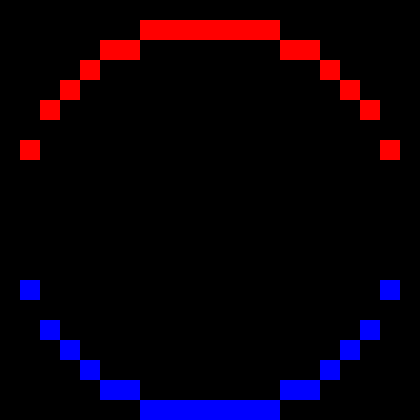}
    \end{tabular}
    \caption{Large disk blur kernels are generated by offseting and truncating a distance function. These ideal kernels can be well approximated by a discretized disk. The sparsity of the discrete kernel's gradient in $y$---shown with red and blue representing positive and negative values, respectively---allows us to perform a scatter blur with far fewer operations per pixel.}
    \label{fig:blur_kernel_approximations}
    \vspace{-6pt}
\end{figure}

An additional aspect of bokeh that we need to tackle is realism. Specifically, we must ensure that colors from the background do not ``leak'' into the foreground. We can guarantee this if we process pixels ordered by their depth as described at the start of Sec.~\ref{sec:rendering}. The computational cost of such an approach can be high, however, so instead we compute the blur in five passes covering different disparity ranges. The input RGB image $I(\mathbf{x}) = [I_R(\mathbf{x}), I_G(\mathbf{x}), I_B(\mathbf{x})]$ is decomposed into a set of premultiplied RGBA sub-images $I_j(\mathbf{x})$. For brevity, we will omit the per-pixel indices for the remainder of this section. Let $\{d_j\}$ be a set of cutoff disparities (defined later) that segment the disparity range into bands such that the $j$-th sub-image, $I_j$, spans disparities $d_{j-1}$ to $d_j$. We can then define 
\begin{equation}
    I_j = \alpha_j[I_R, I_G, I_B, 1],
\end{equation}
where $\alpha_j$ is a truncated tent function on the disparities in that range:
\begin{equation}
    \alpha_j = \left.\left(1 + \frac{1}{\eta}\min(d - d_{j-1}, d_j - d)\right)\right|_{\left[0,1\right]},
\end{equation}
where $\left.\left(\cdot\right)\right|_{\left[a,b\right]}$ signifies clamping a value to be within $[a,b]$. $\alpha_j$ is 1 for pixels with disparities in $[d_{j-1},d_j]$ and tapers down linearly outside the range, hitting 0 after $\eta$ disparity units from the bounds. We typically set $\eta$ so that adjacent bands overlap 25\%, i.e., $\eta = 0.25 \times (d_{j} - d_{j-1})$, to get a smoother transition between them.

The disparity cutoff values $\{d_j\}$ are selected with the following rules. First let us denote the disparity band containing the in-focus parts of the scene as $I_0$. Its bounds, $d_{-1}$ and $d_0$, are calculated to be the disparities at which the blur radius from Eq.~\ref{eqn:blur_radius} is equal to $r_\mathit{brute}$ (see supplemental material for details). The remaining cutoffs are chosen such that the range of possible disparities is evenly divided between the remaining four bands. This approach allows us to optimize for specific blur kernel sizes within each band. For example, $I_0$ can be efficiently computed with a brute-force scatter blur (see Eq.~\ref{eqn:scatter_as_gather}). The other, more strongly blurred sub-images use the accelerated gradient image blur algorithm described earlier.  Fig.~\ref{fig:blur_examples} shows the effect of a single pass blur compared to our proposed method.

The above optimizations are only necessary when we have full disparity information. If we only have a segmentation mask $M$, the blur becomes spatially invariant, so we can get equivalent results with a single-pass, gather-style blur over the background pixel sub-image $I_\mathit{background} = (1-M)[I_R,I_G,I_B,1]$. In this case, we simulate a depth plane by doing a weighted average of the input and blurred images with spatially-varying weight that favors the input image at the bottom and the blurred image at the top.

\subsection{Producing the final image}
\label{sec:rendering_finish}

After the individual layers have been blurred, they are upsampled to full-resolution and composited together, back to front. One special in-focus layer is inserted over the sub-image containing $d_\mathit{focus}$, taken directly from the full resolution input, to avoid any quality loss from the downsample/upsample round trip imposed by the blur pipeline.

The final stage of our rendering pipeline adds synthetic noise to the blurred portions of the image. This is somewhat unusual, as post-processing algorithms are typically designed to reduce or eliminate noise. In our case, we want noise because our blur stages remove the natural noise from the blurry portions of the scene, creating a mismatch in noise levels that yields an unrealistic ``cardboard cutout'' appearance. By adding noise back in, we can reduce this effect. Fig.~\ref{fig:noise_examples} provides a comparison of different noise treatments.

Our noise function is constructed at runtime from a set of periodic noise patches $N_i$ of size $l_i \times l_i$ (see supplemental  material for details). $l_i$ are chosen to be relatively prime, so that we can combine the patches to get noise patterns that are large, non-repeating, and realistic, with little data storage overhead. Let us call our composited, blurred image so far $B_\mathit{comp}$. Our final image with synthetic noise $B_\mathit{noisy}$, then, is
\begin{equation}
B_\mathit{noisy}(\mathbf{x}) = B_\mathit{comp}(\mathbf{x}) + \sigma(\mathbf{x}) M_\mathit{blur}(\mathbf{x}) \sum_{i} N_i(\mathbf{x} \bmod l_i),
\end{equation}
where $\sigma(\mathbf{x})$ is an exposure-dependent estimate of the local noise level in the input image provided by the sensor chip vendor and $M_\mathit{blur}$ is a mask corresponding to all the blurred pixels in the image.

\newcommand{\noiseheight}{1.0in}
\newcommand{\noisetablegap}{0.01in}
\begin{figure}[t]
    \centering
    \begin{tabular}{@{}c@{\hskip3pt}c@{}c@{}c@{}}
    & \parbox{\noiseheight}{\centering Blurred w/o \\ synth. noise} & \parbox{\noiseheight}{\centering Blurred w/ \\
    synth. noise} & \parbox{\noiseheight}{\centering Unblurred \\
    input image} \\
    \rotatebox[origin=l]{90}{\parbox{\noiseheight}{\centering Image}} &
        \includegraphics[width=\noiseheight]{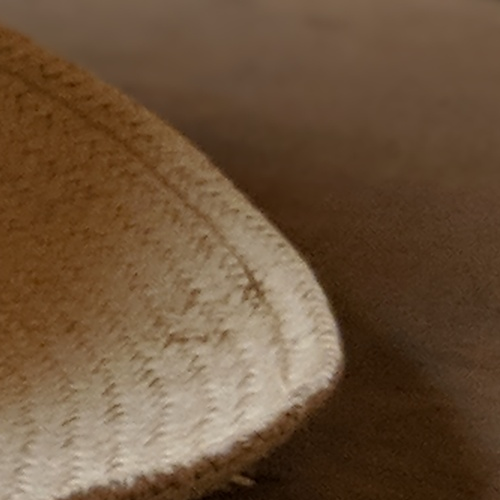} & 
        \hspace{\noisetablegap}
        \includegraphics[width=\noiseheight]{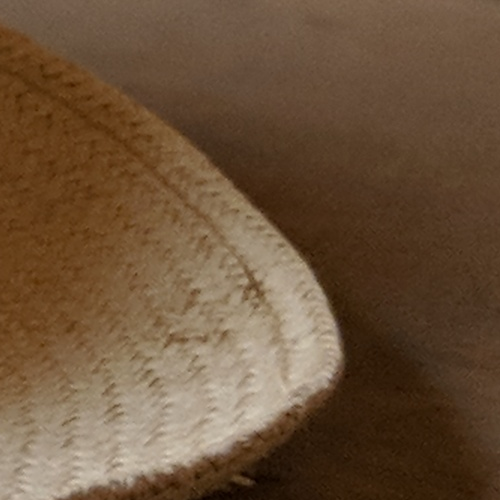} & 
        \hspace{\noisetablegap}
        \includegraphics[width=\noiseheight]{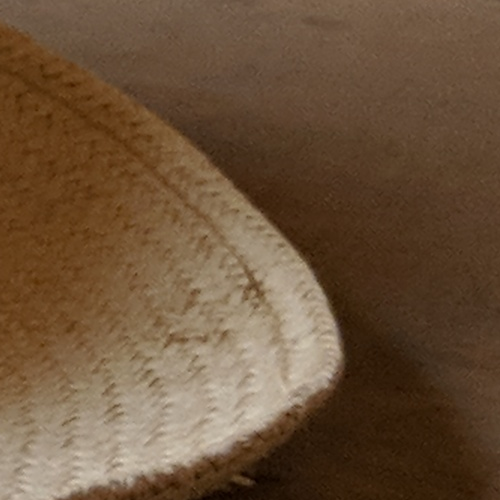} \\
    \rotatebox[origin=l]{90}{\parbox{\noiseheight}{\centering High-pass}} &
        \includegraphics[width=\noiseheight]{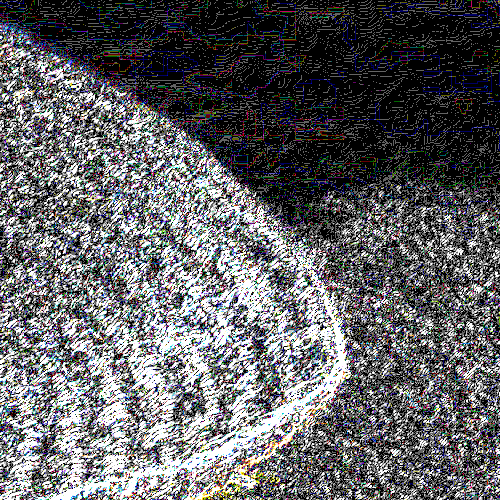} & 
        \hspace{\noisetablegap}
        \includegraphics[width=\noiseheight]{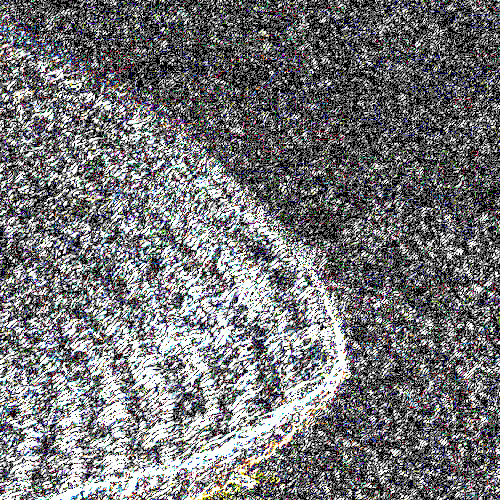} & 
        \hspace{\noisetablegap}
        \includegraphics[width=\noiseheight]{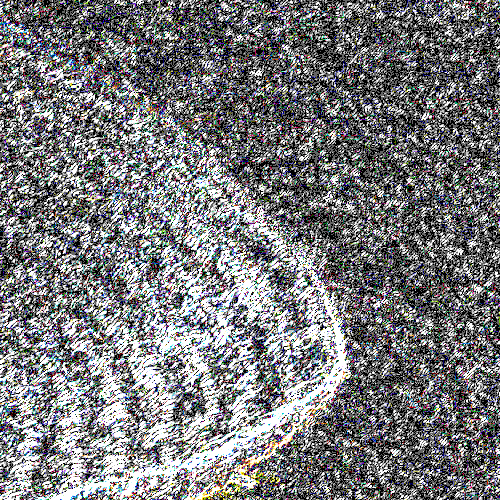} \\
    \end{tabular}
    \caption{Our synthetic defocus removes naturally appearing noise in the image. This causes visual artifacts at the transitions between blurred and sharp parts of the frame. Adding noise back in can hide those transitions, making our results appear more realistic. The effect is subtle in print, so we emphasize the noise in the bottom row with a high-pass filter. This figure is best viewed digitally on a large screen.}
    \label{fig:noise_examples}
    \vspace{-6pt}
\end{figure}

\section{Results}

As described earlier, our system has three pipelines. Our first pipeline is {\bf DP + Segmentation} (Fig.~\ref{fig:teaser}). It applies to scenes with people taken with a camera that has dual-pixel hardware and uses both disparity and a mask from the segmentation network. Our second pipeline, {\bf DP only} (Fig.~\ref{fig:teaser2-b}), applies to scenes of objects taken with a camera that has dual-pixel hardware. It is the same as the first pipeline except there is no segmentation mask to integrate with disparity. Our third pipeline, {\bf Segmentation only} (Fig.~\ref{fig:teaser2-a}), applies to images of people taken with the front-facing (selfie) camera, which typically does not have dual-pixel hardware. The first two pipelines use the full depth renderer (Sec.~\ref{sec:rendering_blur}). The third uses edge-aware filtering to snap the mask to color edges (Sec.~\ref{sec:edge_aware_filtering_of_masks}) and uses the less compute-intensive two-layer mask renderer (end of Sec.~\ref{sec:rendering_blur}). 

%Our third {\bf Segmentation only} pipeline works for images of people on cameras that do not have DP hardware (Fig.~\ref{fig:teaser2}a). This pipeline is well-suited to the front-facing selfie camera on mobile phones. This camera is often fixed-focus because the subject is expected to be a person a fixed-distance from the camera. The controlled nature of image capture means that most photos from this camera are of a few people in front of a distant background. 
%We use the segmentation network to segment out people using the same heuristics as the {\bf DP + Segmentation} pipeline. We use edge-aware filtering of the segmentation mask to make the network's output snap to color edges between the people and the background (Section~\ref{sec:edge_aware_filtering_of_masks}). We use a two-layer version of the renderer, called the {\em mask renderer}, which keeps pixels in the mask sharp and blurs the background by an amount that increases with face size. %Since the bottom of the scene is usually closer to the camera than the top, we employ a heuristic in which we blend the blurred image with the input favoring the input at the bottom and the blurred image at the top. \neal{TODO: Specify the exact details.}

\begin{figure*}[t]%
\centering
\hspace{-12pt}
\subfigure[\small Input w/ face (green)]{%
\begin{tabular}{c}
\includegraphics[width=0.165\textwidth]{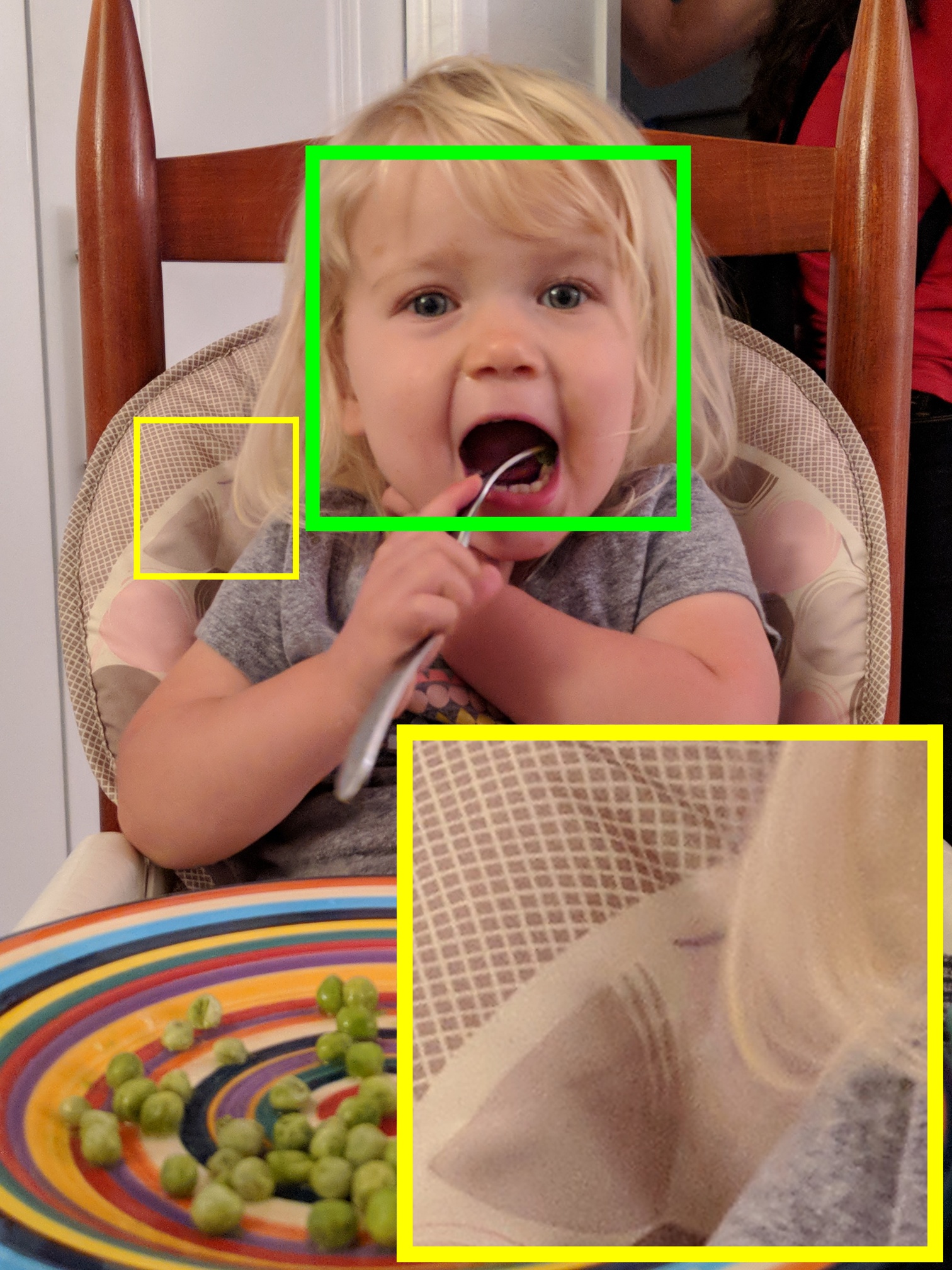}\\
\includegraphics[width=0.165\textwidth]{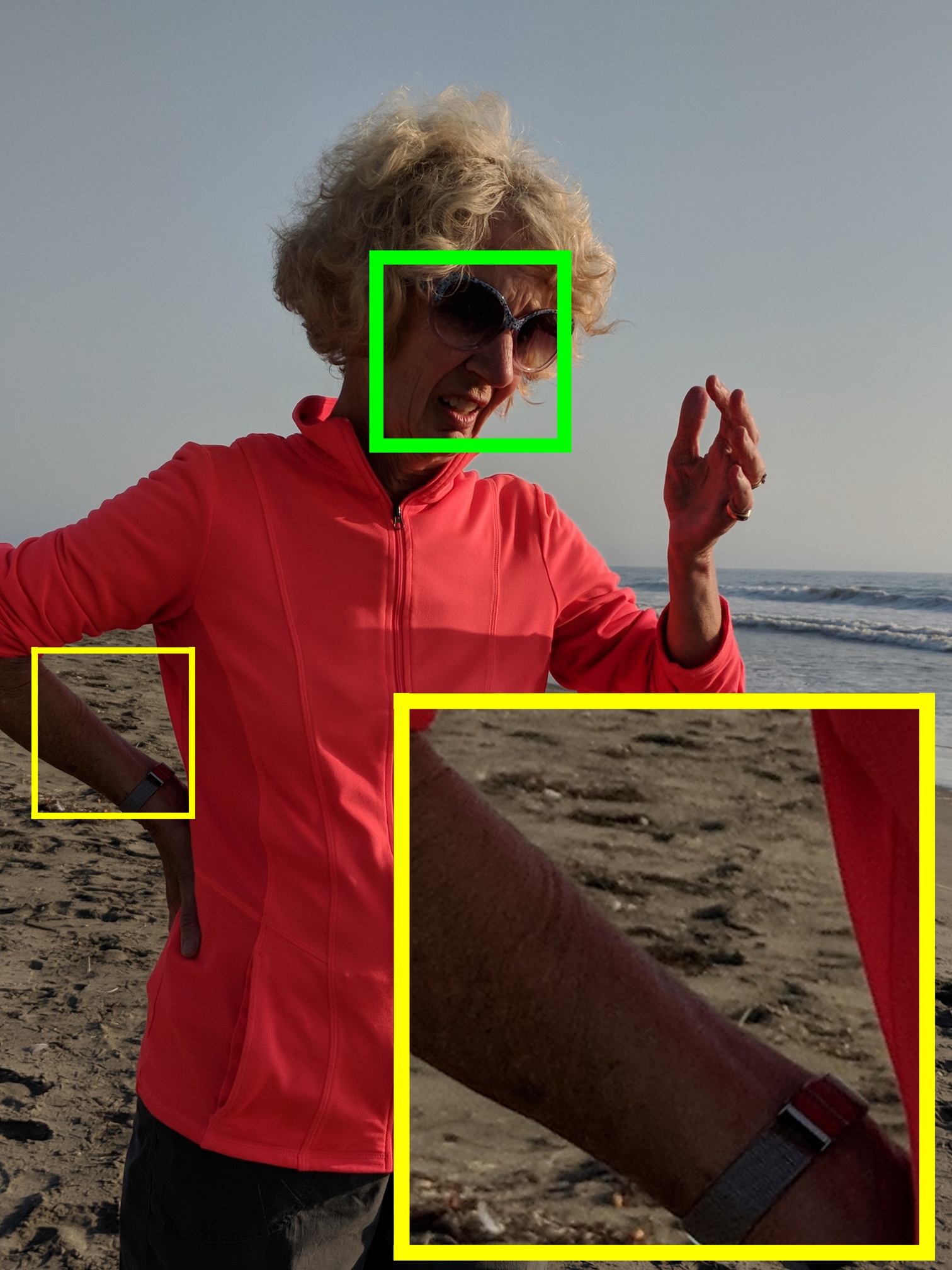}
\end{tabular}}%
\hspace{-12pt}
\subfigure[\small Shen \etal~\protect\shortcite{shen2016automatic}]{%
\begin{tabular}{c}
\includegraphics[width=0.165\textwidth]{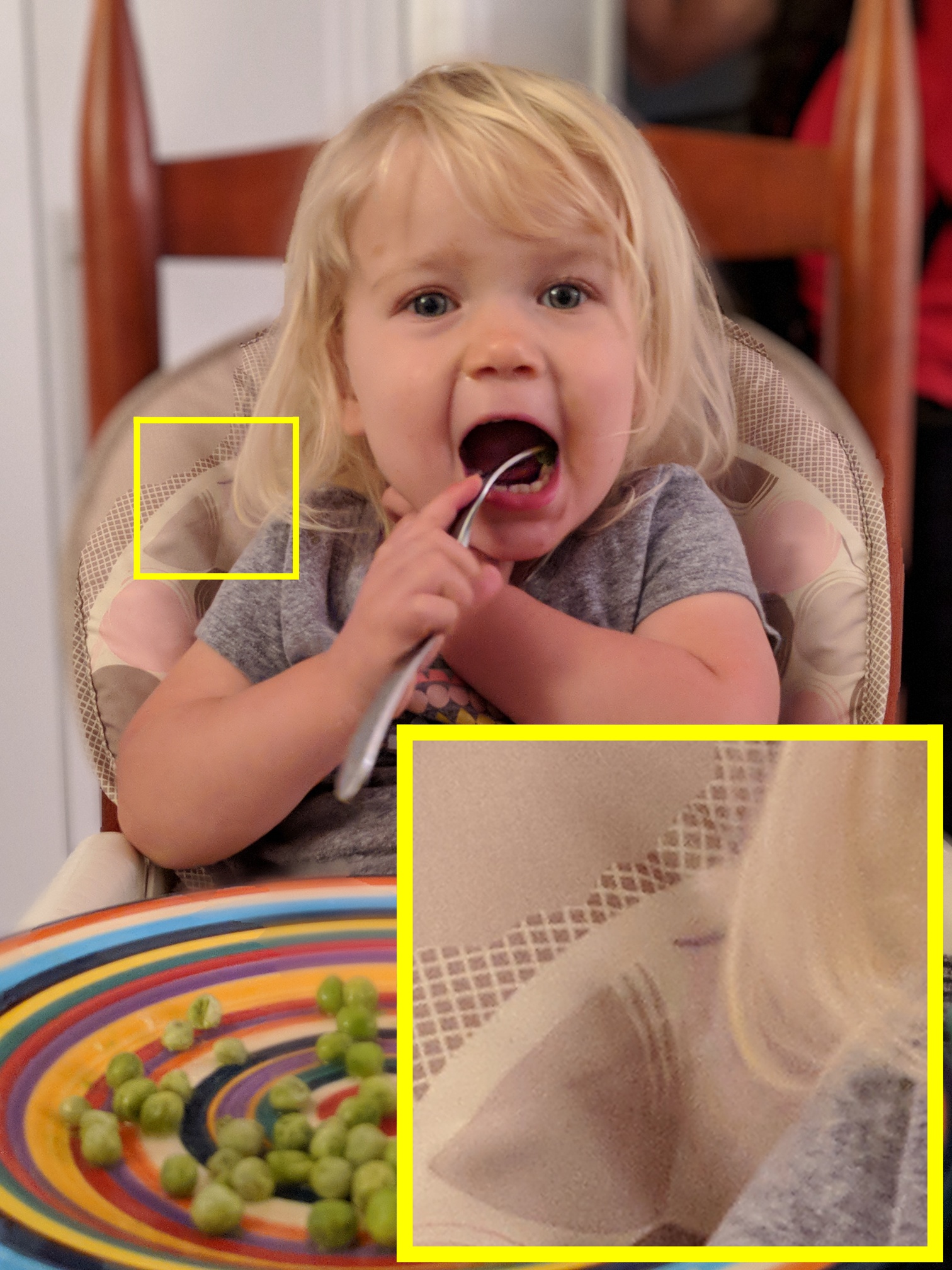}\\
\includegraphics[width=0.165\textwidth]{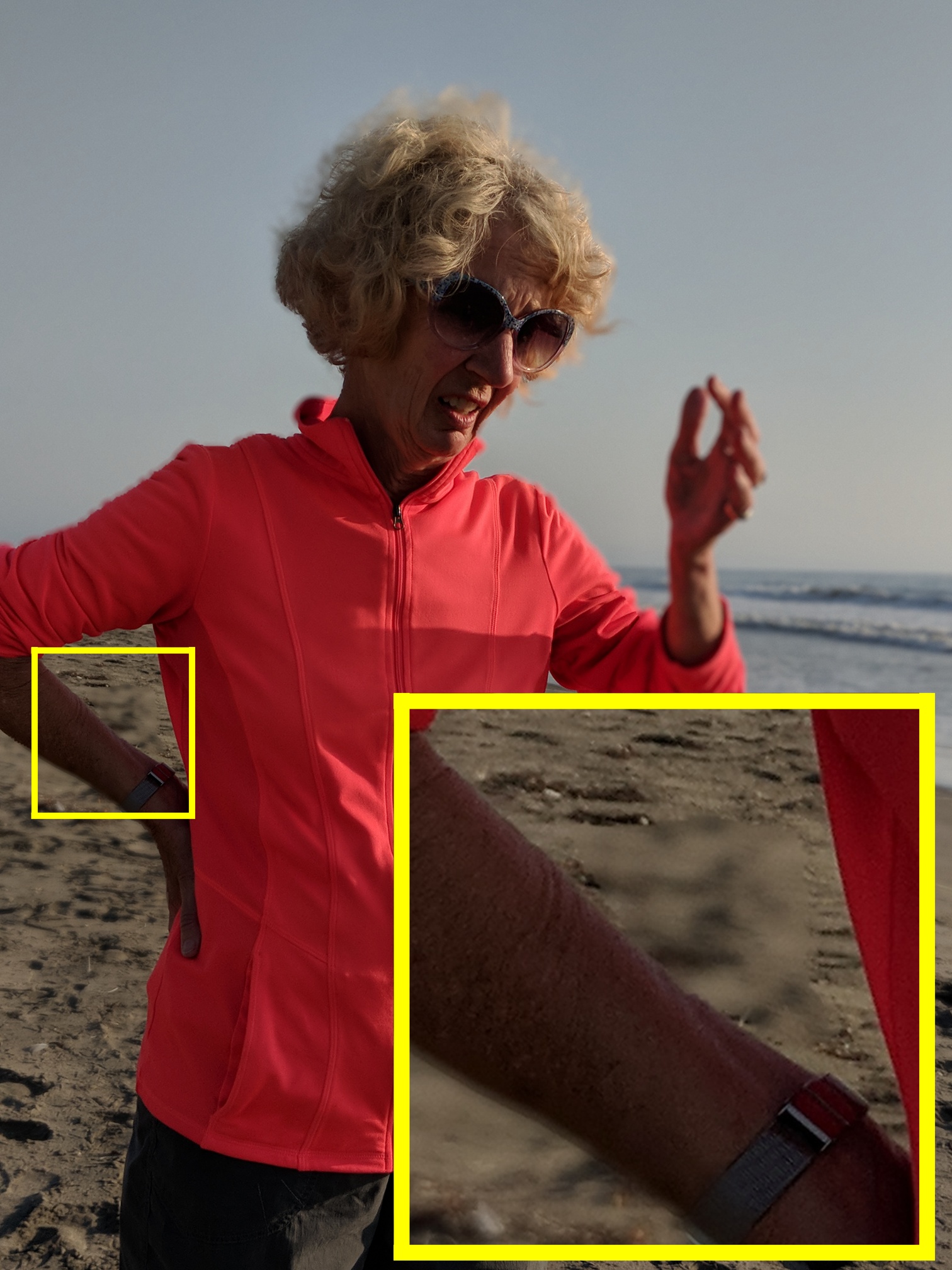}
\end{tabular}}%
\hspace{-12pt}
\subfigure[\small Seg. only (ours)]{%
\begin{tabular}{c}
\includegraphics[width=0.165\textwidth]{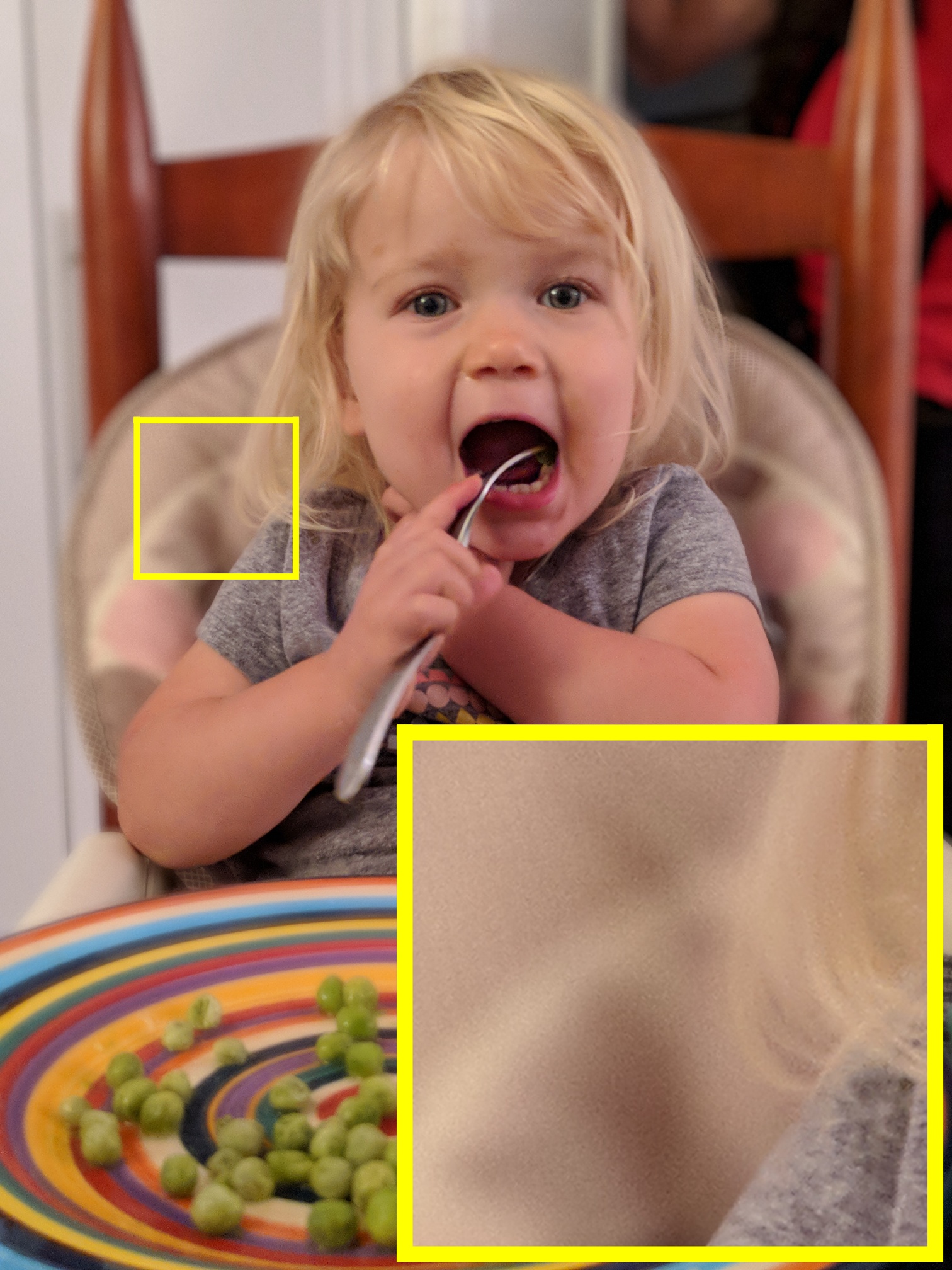}\\
\includegraphics[width=0.165\textwidth]{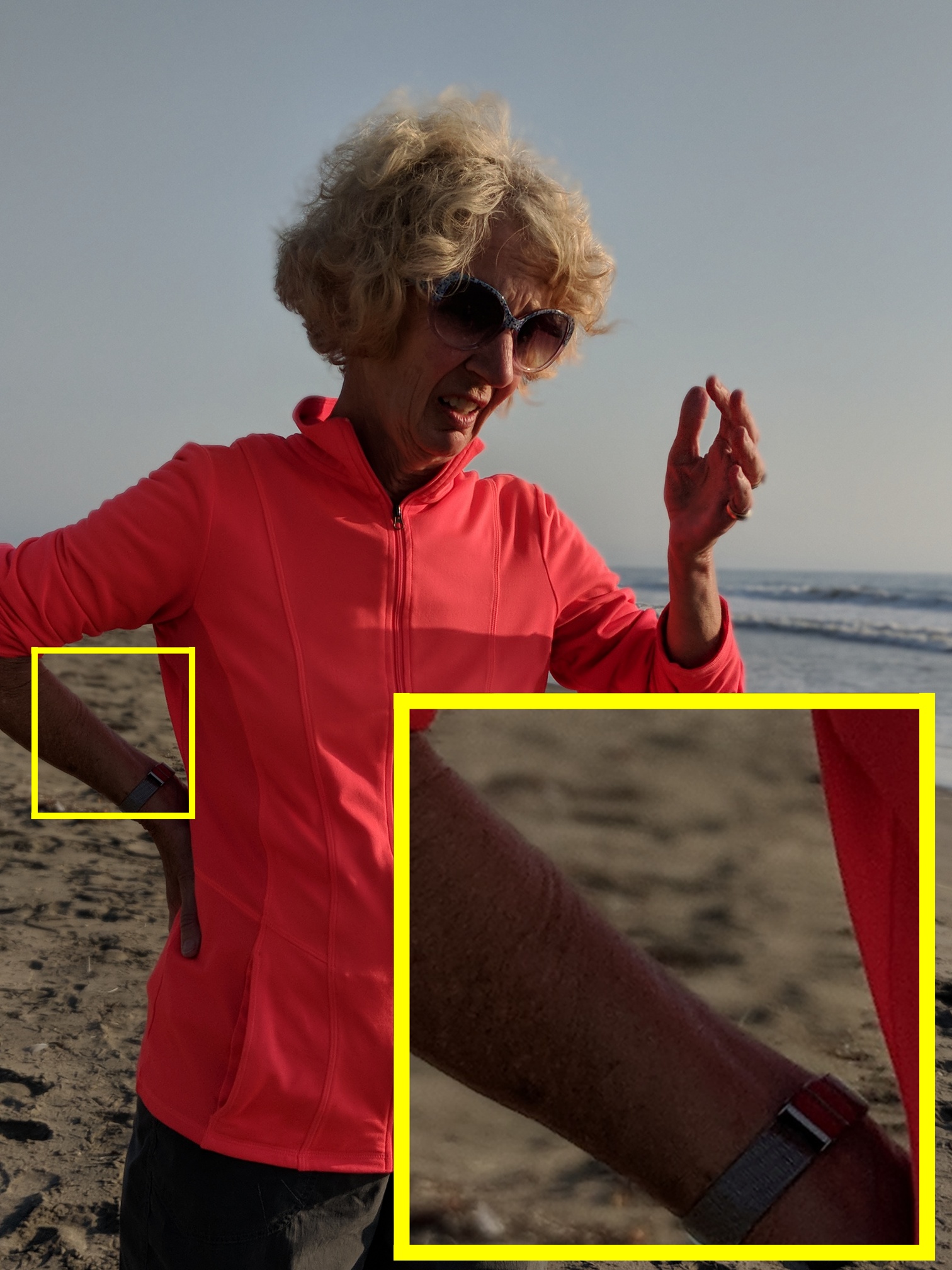}
\end{tabular}}%
\hspace{-12pt}
\subfigure[\small Barron \etal~\protect\shortcite{barron2015fast}]{%
\begin{tabular}{c}
\includegraphics[width=0.165\textwidth]{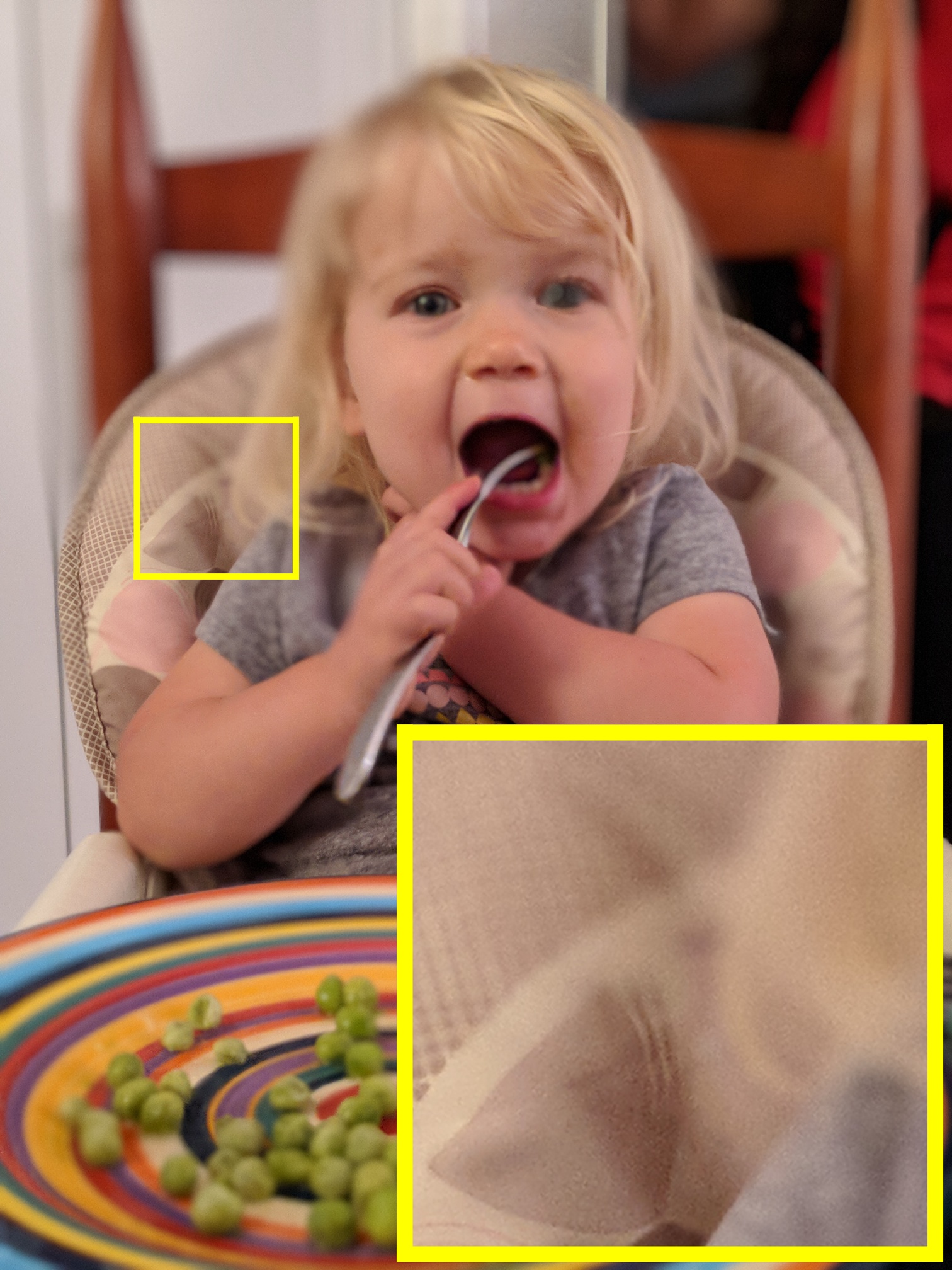}\\
\includegraphics[width=0.165\textwidth]{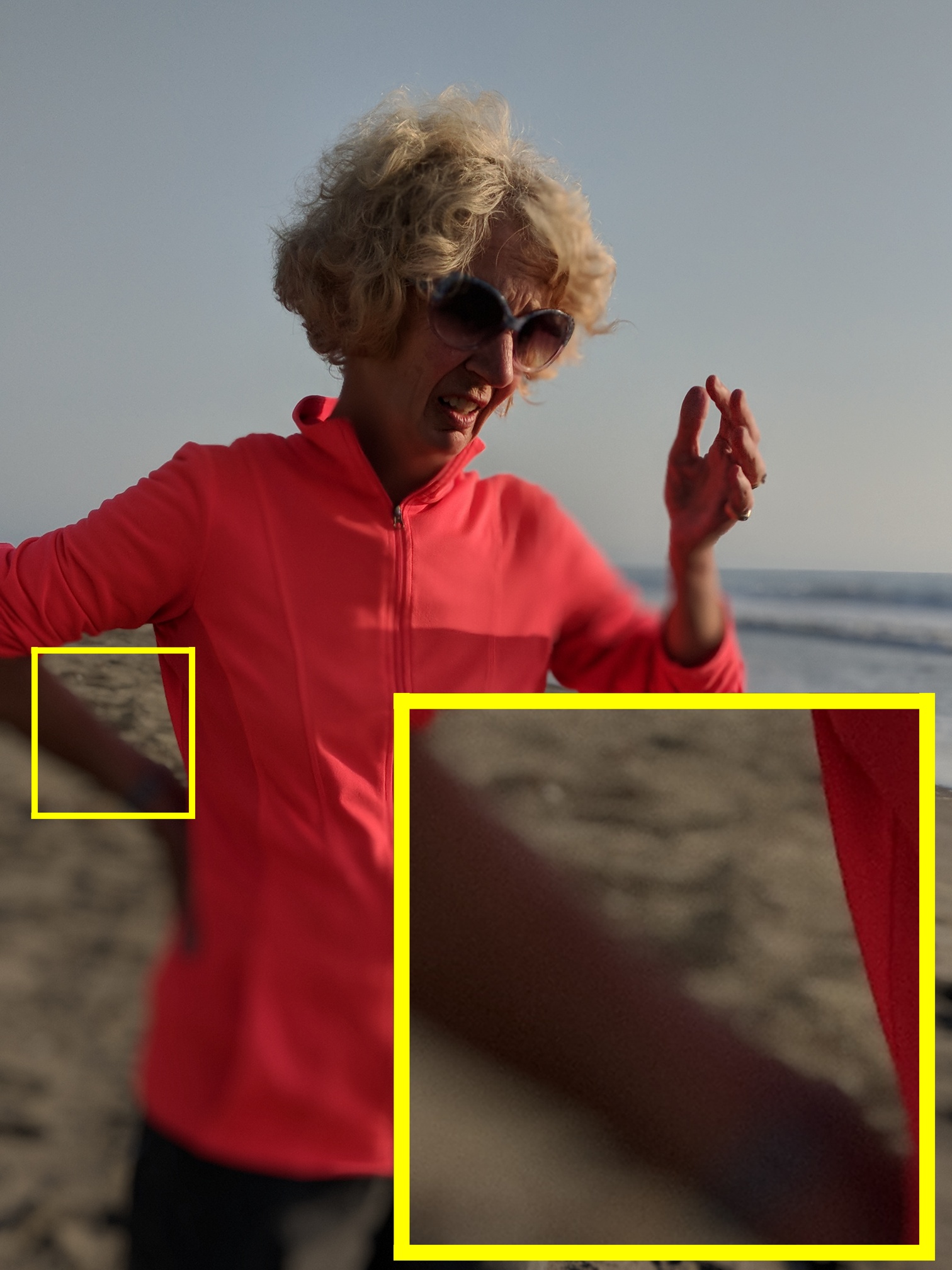}
\end{tabular}}%
\hspace{-12pt}
\subfigure[\small DP only (ours)]{%
\begin{tabular}{c}
\includegraphics[width=0.165\textwidth]{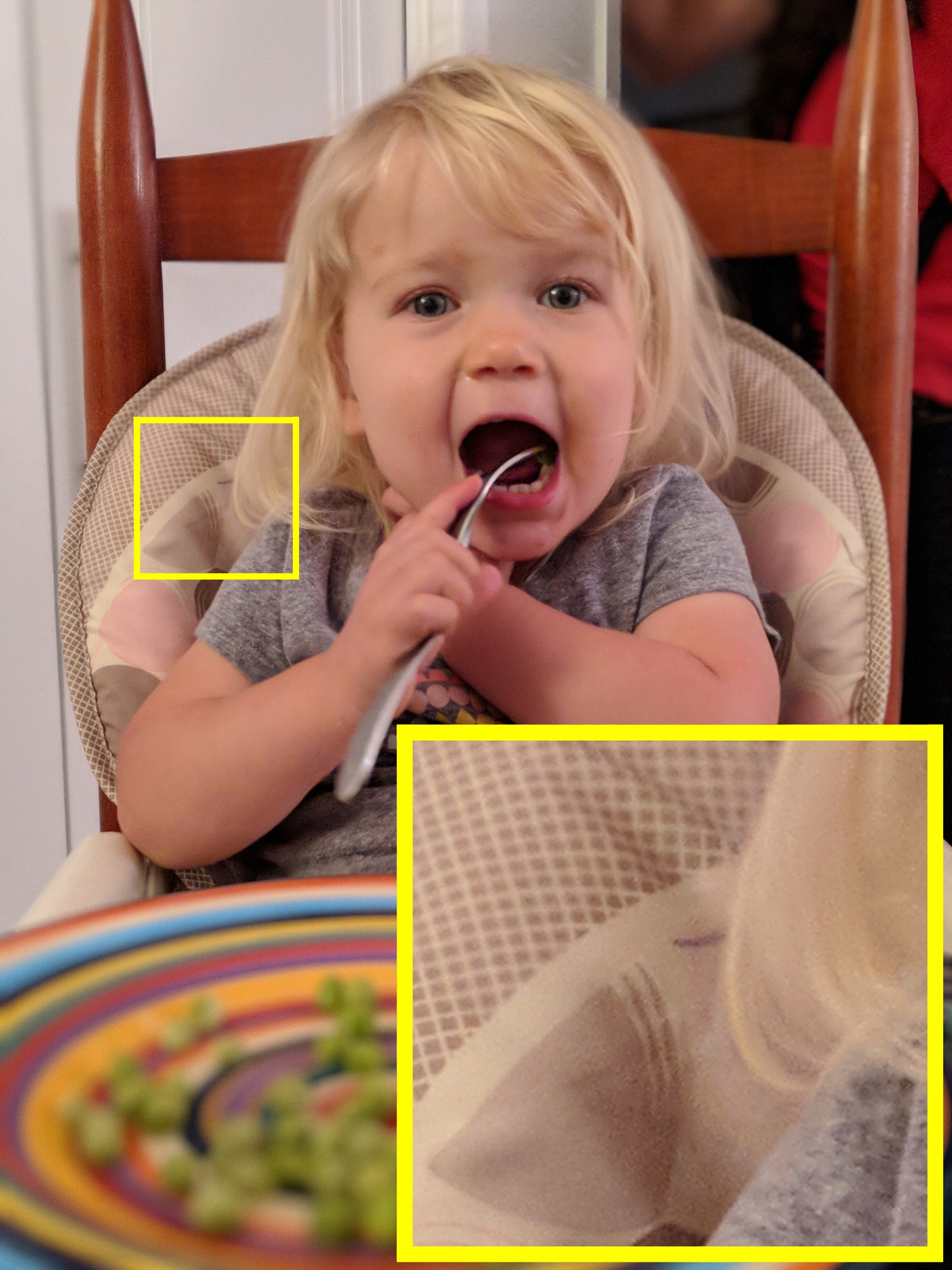}\\
\includegraphics[width=0.165\textwidth]{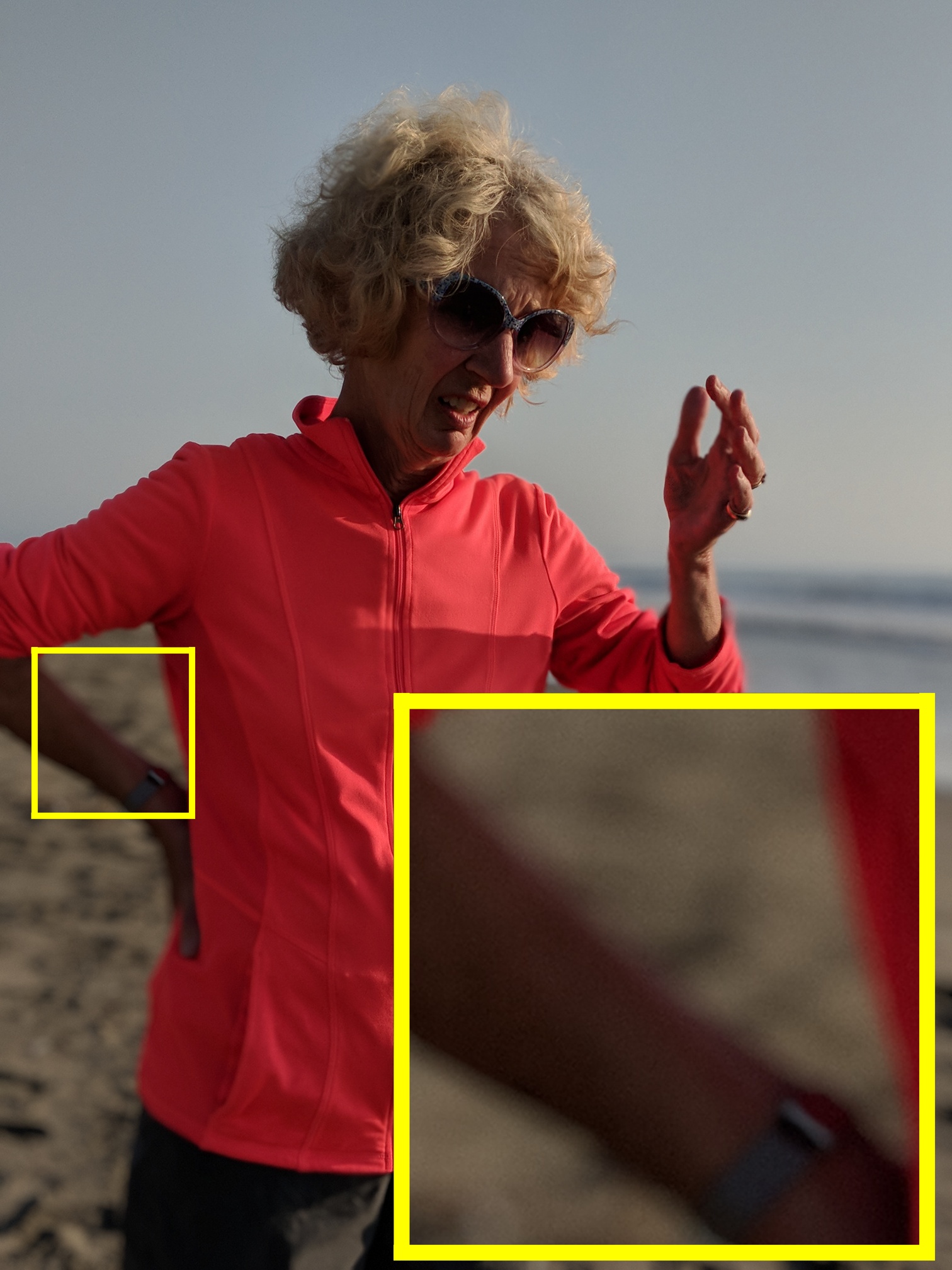}
\end{tabular}}%
\hspace{-12pt}
\subfigure[\small DP + Seg. (ours)]{%
\begin{tabular}{c}
\includegraphics[width=0.165\textwidth]{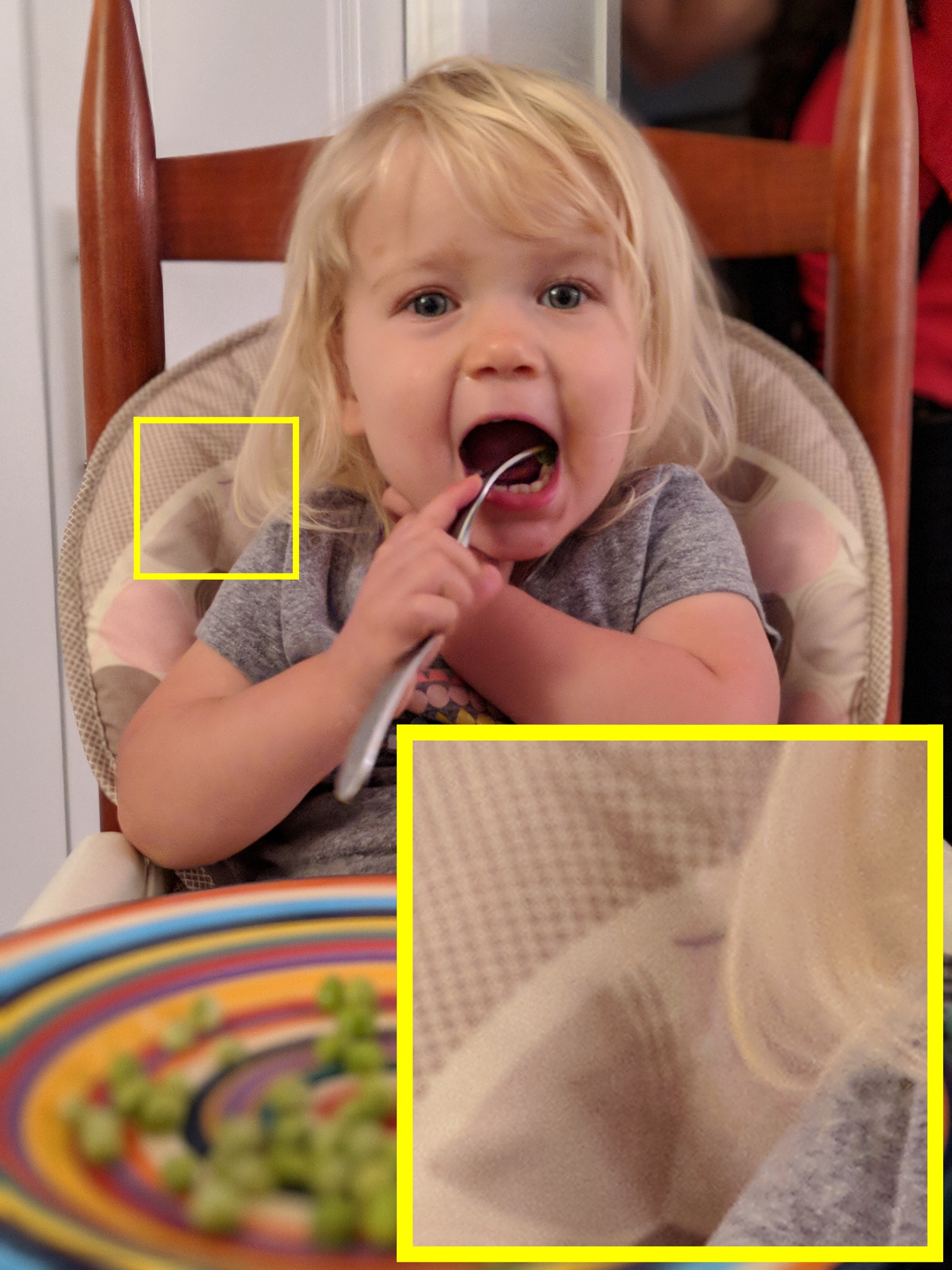}\\
\includegraphics[width=0.165\textwidth]{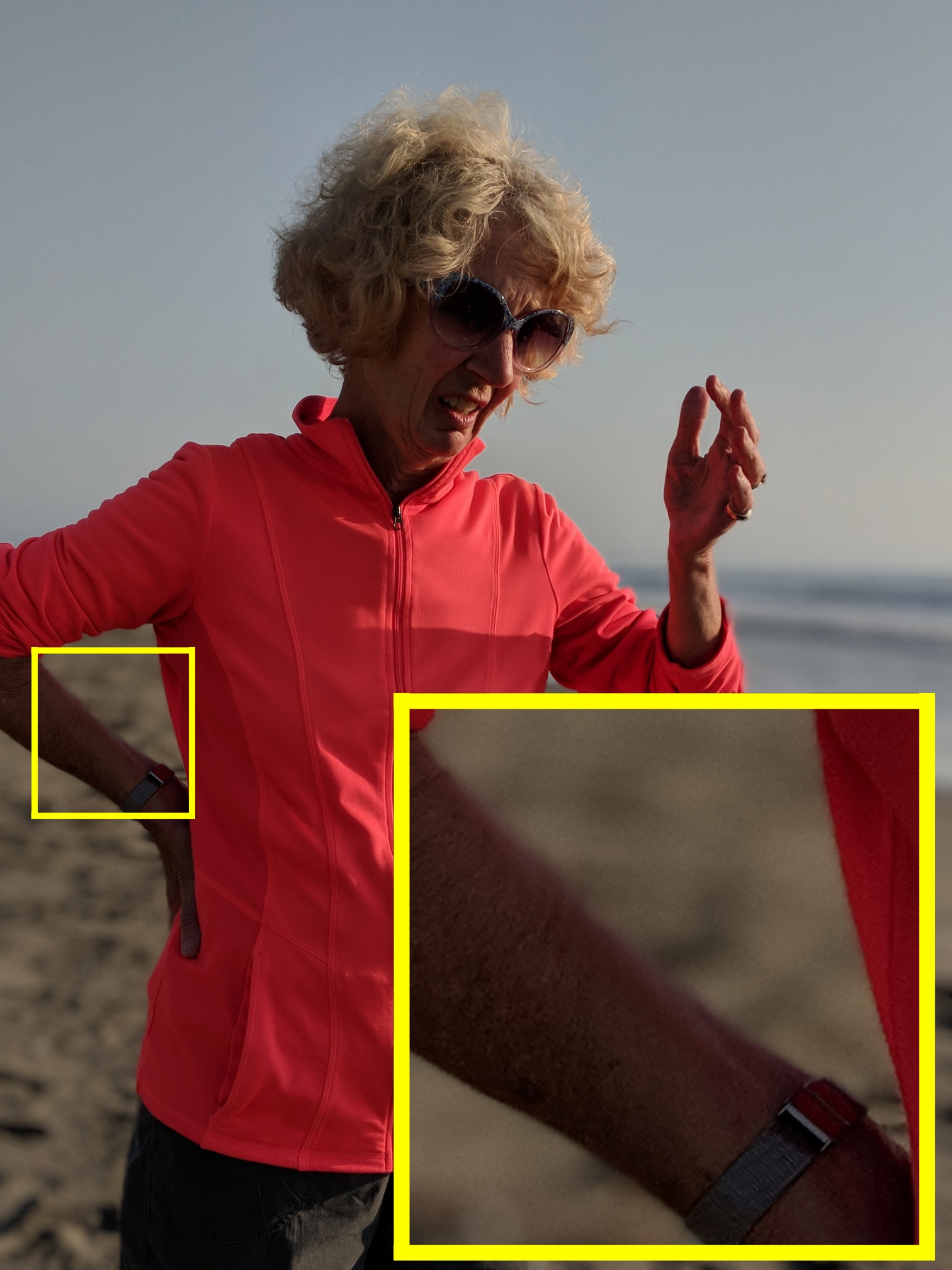}
\end{tabular}}%
\hspace{-12pt}
\vspace{-10pt}
\caption{Output images and insets (in yellow) produced using our three pipelines and two other methods. In all cases, we use our renderer. In the top example, the segmentation networks (b-c) are unable to determine that the cushion is at nearly the same depth as the child and it gets blurred. In the bottom example, the algorithms that use DP (d-e) compute an incorrect disparity on the woman's dark, untextured arm causing it to get blurred with the background. In both cases, our combined result (f) blurs the background and keeps the subject sharp. }
\label{fig:result}
\vspace{-10pt}
\end{figure*}

We show a selection of over one hundred results using all three pipelines in the supplemental materials. We encourage the reader to look at these examples and zoom-in on the images. To show how small disparity between the DP views is, we provide animations that switch between the views. We also show the disparity and segmentation mask for these results as appropriate. 

We also conducted a user-study for images that have both DP data and a person. We ran these images through all three versions of our pipeline as well as the person segmentation network from Shen \etal~\shortcite{shen2016automatic} and the stereo method of Barron \etal~\shortcite{barron2015fast}. We used our renderer in both of the latter cases. We chose these two works in particular to compare against because they explicitly focus on the problem of producing synthetic shallow depth-of-field images. 

In our user study, we asked $6$ people to compare the output of $5$ algorithms (ours, two ablations of ours, and the two aforementioned baseline algorithms) on $64$ images using a similar procedure as Barron \etal~\shortcite{barron2015fast}. All users preferred our method, often by a significant margin, and the ablations of our model were consistently the second and third most preferred. See the supplement for details of our experimental procedure, and selected images from our study.

Two example images run through our pipeline are shown in Fig.~\ref{fig:result}. For the first example in the top row, notice how the cushion the child is resting on is blurred in the segmentation-only methods (Fig.~\ref{fig:result}(b-c)). In the second example, the depth-based methods provide incorrect depth estimates on the textureless arms. Using both the mask and disparity makes us robust to both of these kinds of errors.

\makeatletter
\newcommand{\mypm}{\mathbin{\mathpalette\@mypm\relax}}
\newcommand{\@mypm}[2]{\ooalign{%
  \raisebox{.1\height}{$#1+$}\cr
  \smash{\raisebox{-.6\height}{$#1-$}}\cr}}
\makeatother

\begin{table}
    \begin{center}
    \caption{
    The results of our user study, in which we asked $6$ participants to select the algorithm (or ablation of our algorithm) whose output they preferred. Here we accumulate votes for each algorithm used across $64$ images, and highlight the most-preferred algorithm in red and the second-most preferred algorithm in yellow. All users consistently prefer our proposed algorithm, with the ``segmentation only'' algorithm being the second-most preferred.
    }
    \vspace{-12pt}
    \resizebox{\columnwidth}{!}{%
    \huge
    \begin{tabular}{l || cccccc | c@{\hskip3pt}c@{\hskip3pt}c}
 & \multicolumn{6}{c|}{User} & & & \\
 Method & A  & B  & C & D & E & F & mean & $\mypm$ & std. \\ \hline
Barron \etal~\shortcite{barron2015fast}  & $  0$ & $  2$ & $  0$ & $  0$ & $  0$ & $  0$ &  $ 0.3 $ & $ \mypm $ & $ 0.8 $  \\ 
Shen \etal~\shortcite{shen2016automatic}  & $  7$ & $  6$ & $  9$ & $  3$ & $  0$ & $  7$ &  $ 5.3 $ & $ \mypm $ & $ 3.3 $  \\ 
DP only  (ours)  & $ 10$ & $ 11$ & $ 16$ & $ 15$ & $ 14$ & $  8$ &  $ 12.3 $ & $ \mypm $ & $ 3.1 $  \\ 
Segmentation only  (ours)  & \cellcolor{Yellow} $ 13$ & \cellcolor{Yellow} $ 22$ & \cellcolor{Yellow} $ 17$ & \cellcolor{Yellow} $ 19$ & \cellcolor{Yellow} $ 18$ & \cellcolor{Yellow} $ 22$ & \cellcolor{Yellow}  $ 18.5 $ &\cellcolor{Yellow}  \!$ \mypm $ &\cellcolor{Yellow} \!\!\!  $ 3.4 $  \\ 
DP + Segmentation  (ours)  & \cellcolor{Red} $ 34$ & \cellcolor{Red} $ 23$ & \cellcolor{Red} $ 22$ & \cellcolor{Red} $ 27$ & \cellcolor{Red} $ 32$ & \cellcolor{Red} $ 27$ & \cellcolor{Red}  $ 27.5 $ &\cellcolor{Red}  \!$ \mypm $ &\cellcolor{Red} \!\!\! $ 4.8 $ 
    \end{tabular}%
    }
    
    \label{tab:user_study}
    \end{center}

\end{table}

\begin{figure}[t]
\centering
\includegraphics[width=\columnwidth]{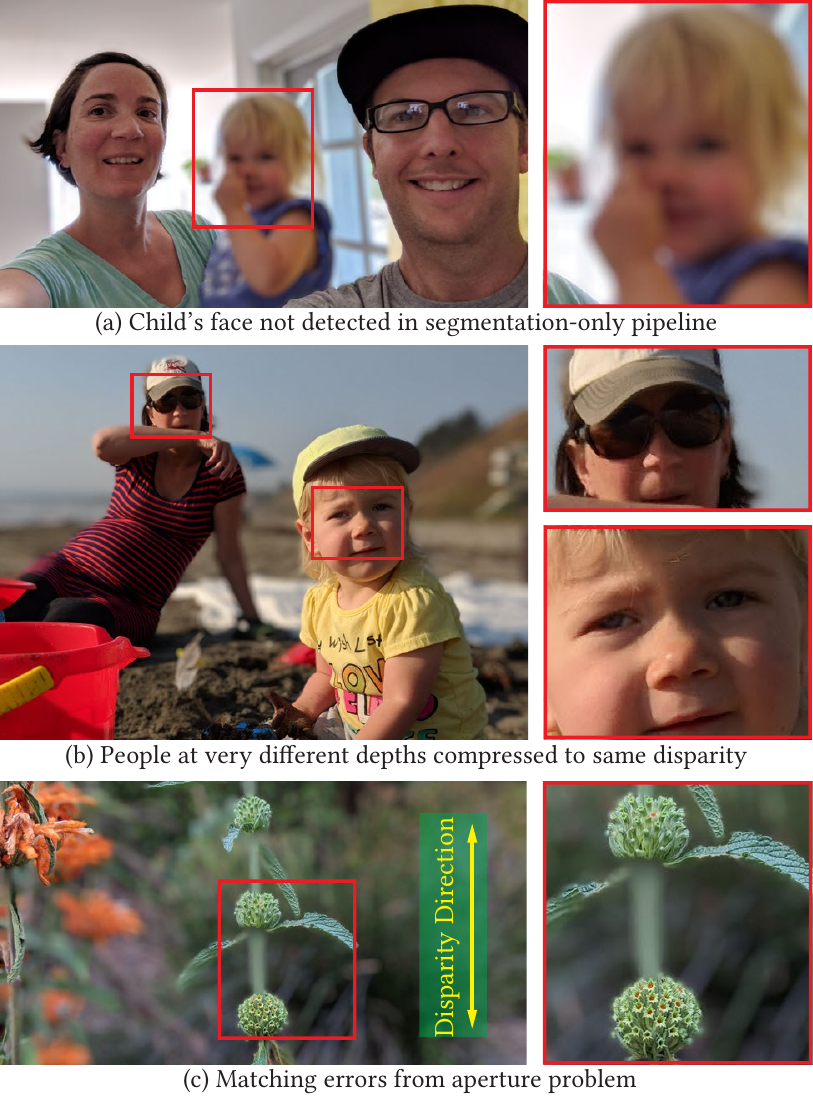}
\caption{Failure modes of our system. Face detection failures in the segmentation-only pipeline can cause people to get blurred with the background (a). Keeping all people with large-enough faces sharp can look unnatural (b). In this extreme case, the child and woman's faces have within a factor of three the same number of pixels, but the woman is much further back. Vertical structures parallel to the disparity direction in (c) are incorrectly assigned the disparity of the background and get blurred (c).}
\label{fig:failures}
\vspace{-10pt}
\end{figure}

Our system is in production and is used daily by millions of people. That said, it has several failure modes. Face-detection failures can cause a person to get blurred with the background (Fig.~\ref{fig:failures}(a)). DP data can help mitigate this failure as we often get reasonable disparities on people at the same depth as the main subject. However, on cameras without DP data, it is important that we get accurate face detections as the input to our system. Our choice to compress the disparities of multiple people to the same value can result in artifacts in some cases. In Fig.~\ref{fig:failures}(b), the woman's face is physically much larger than the child's face. Even though she is further back, both her and the child are selected as people to segment resulting in an unnatural image, in which both people are sharp, but scene content between them is blurred. We are also susceptible to matching errors during the disparity computation, the most prevalent of which are caused by the well-known aperture problem in which it is not possible to find correspondences of image structures parallel to the baseline. The person segmentation mask uses semantics to prevents these errors from causing problem on the subject's body. However, they can still show up on images without people (Fig.~\ref{fig:failures}(c)) or in the backgrounds behind people.

We measured the performance of our system on a modern high-end mobile phone. The CPU had eight cores, four running at 2.35 GHz and the rest at 1.9 GHz. Our code was implemented in Halide \cite{ragan2013halide}, then manually scheduled for the CPU. We ran our system on 325 examples of resolution $2688\times 2016$ using all three pipelines. They all take less than 4 seconds (Table~\ref{tab:running}).

\begin{table}
\centering
\caption{Running time in milliseconds of parts of our algorithm on a high-end mobile phone. The three totals correspond to different variants of our algorithm. The standard deviation $\sigma$ are shown next to the running times. All stages were run on over 100 images.}

\resizebox{\columnwidth}{!}{%
    \Huge
\begin{tabular}{l|c c c l@{\hskip2pt}r}
Stage & DP only & Seg. only & DP + Seg.  \\
\hline \hline
%Burst Denoising \cite{hasinoff2016burst} &  941 & 941 & 941 & $\sigma= $ & $165$ ms\\
%\hline
Face detection    & $143$  & $143$ & $143$ & $\sigma= $ & $46$ ms\\
Person segmentation  & -- & $1612$ & $1612$ & $\sigma= $ & $729$ ms\\
Denoise DP burst & $190$ & -- & $190$ & $\sigma= $ & $45$ ms\\
Disparity computation & $817$ & -- & $817$ & $\sigma= $ & $153$ ms\\
Depth renderer & $1241$ & -- & $1241$ & $\sigma= $ & $152$ ms\\
Edge-aware mask filtering & -- & $1048$ & -- & $\sigma= $ & $339$ ms\\
Mask renderer & -- & $546$ & -- & $\sigma= $ & $67$ ms\\
\hline
Total  & $2134$ & $3343$ & $3854$ & \\
\end{tabular}}
\label{tab:running}
\vspace{-12pt}
\end{table}

\section{Discussion and Future Work}
We have presented a system to compute synthetic shallow depth-of-field images on mobile phones. Our method combines a person segmentation network and depth from dense dual-pixels. This choice of technologies means that our method is able to run on mobile phones that have only a single camera. We show results on a wide variety of examples and compare our method to other papers that produce synthetic shallow depth-of-field images. Our system is marketed as ``Portrait Mode'' on the Google Pixel and Pixel XL smartphones.

One extension to this work is to expand the range of subjects that can
be segmented to pets, food and other objects people photograph.
In addition, since the rendering is already non-photorealistic, and its purpose is to draw
attention to the subject rather than to realistically simulate depth-of-field, it would be interesting to explore alternative ways of separating
foreground and background, such as desaturation or stylization.  
In general, as users accept that computational photography loosens the
bonds tieing us to strict realism, a world of creative image-making awaits us
on the other side.

\begin{acks}
Shipping our system to millions of users would not have been possible without our close collaboration with the Android camera team. We thank them for integrating our system into the Google Camera app and for their product and engineering effort. We also thank Alireza Fathi, Sam Hasinoff and Ben Weiss for their helpful feedback and technical advice. We give a special thanks to photographer Michael Milne for taking thousands of test photographs for us.
\end{acks}

\bibliographystyle{ACM-Reference-Format}
\bibliography{references.bib}

\end{document}

% \author{Anonymous Authors}
% \affiliation{Anonymous Institution}
\author{Neal Wadhwa}
\author{Rahul Garg}
\author{David E. Jacobs}
\author{Bryan E. Feldman} 
\author{Nori Kanazawa}
\author{Robert Carroll}
\author{Yair Movshovitz-Attias}
\author{Jonathan T. Barron}
\author{Yael Pritch}
\author{Marc Levoy}
\affiliation{Google Research}

%%% Replace ``PAPER TEMPLATE TITLE'' with the title of your paper or abstract.

%\title{Synthetic shallow depth-of-field from a single mobile camera with a dual-pixel sensor}
\title{Supplemental Text for Synthetic Depth-of-Field with a Single-Camera Mobile Phone}

\authorsaddresses{}

\renewcommand{\shortauthors}{Wadhwa et al.}

\maketitle

\ExplSyntaxOn
\newcommand\latinabbrev[1]{
  \peek_meaning:NTF . {% Same as \@ifnextchar
    #1\@}%
  { \peek_catcode:NTF a {% Check whether next char has same catcode as \'a, i.e., is a letter
      #1.\@ }%
    {#1.\@}}}
\ExplSyntaxOff
\def\etal{\latinabbrev{et al}}

\section{Usefulness of Synthetic Data}

\begin{figure}[b]
    \centering
    \includegraphics[width=0.9\columnwidth]{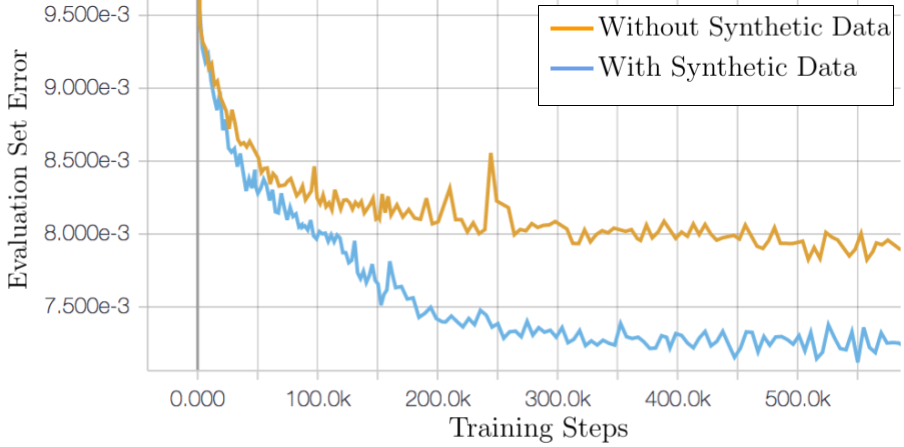}
    \caption{
        Evaluation set error $\operatorname{E}[(\nabla \hat{M} - \nabla M)^2]$ as a function of training steps with and without synthetic training data.
    }
    \label{fig:seg_synthetic_grap}
\end{figure}

We train our segmentation network on synthetic data generated by compositing segmented people onto different backgrounds.
To demonstrate its usefulness, we plot evaluation set accuracy as a function of training iteration with and without synthetic data augmentation in Fig.~\ref{fig:seg_synthetic_grap}. We see that using synthetic data decreases error rates. This evaluation uses a metric that emphasizes the accuracy of mask boundaries: the mean sum of squared error between the gradient of the predicted mask $\hat{M}$ and the gradient of the ground truth mask $M$: $\operatorname{E}[(\nabla \hat{M} - \nabla M)^2]$.

\section{Disparity computation details}
\subsection{DP Normalization}
\begin{figure}
    \centering
    \subfigure[View 1]{\includegraphics[width=0.28\columnwidth]{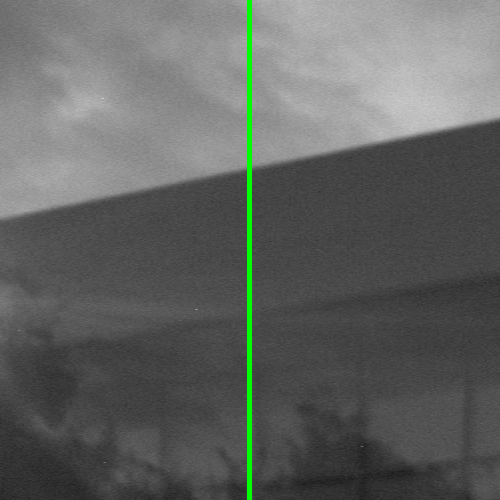}}
    \subfigure[View 2]{\includegraphics[width=0.28\columnwidth]{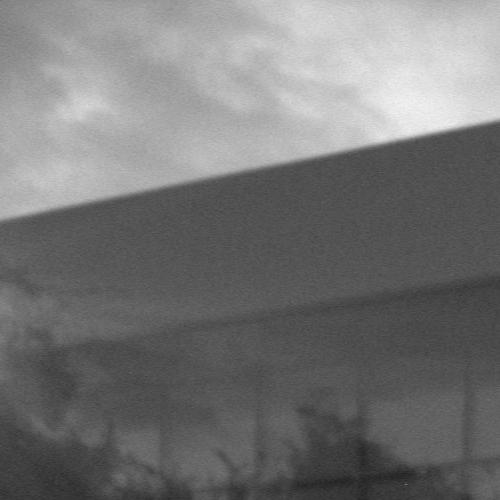}}
    \subfigure[Intensity trace]{\includegraphics[width=0.42\columnwidth]{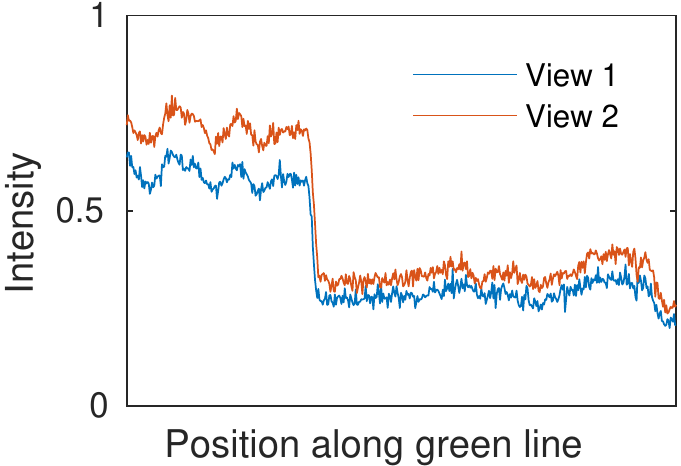}}
    \subfigure[View 1 (normalized)]{\includegraphics[width=0.28\columnwidth]{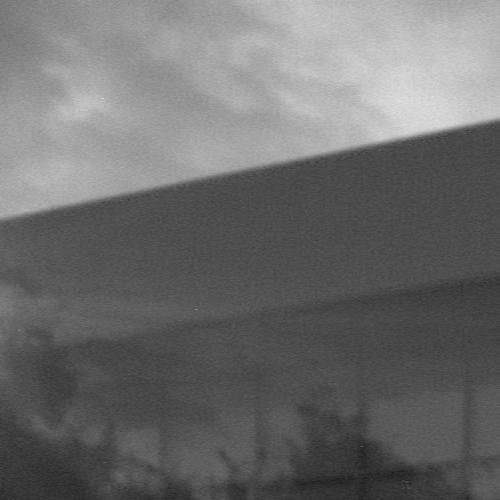}}
    \subfigure[View 2 (normalized)]{\includegraphics[width=0.28\columnwidth]{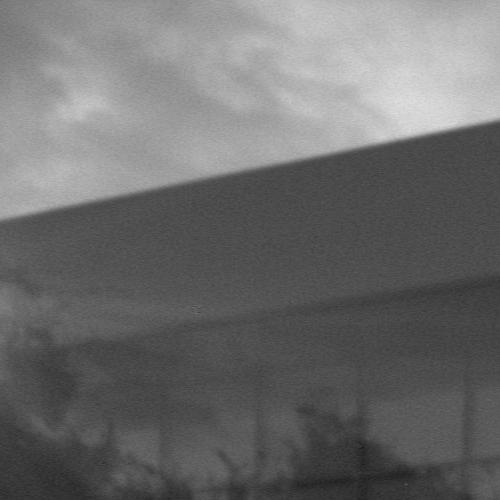}}
    \subfigure[Intensity trace (normalized)]{\includegraphics[width=0.42\columnwidth]{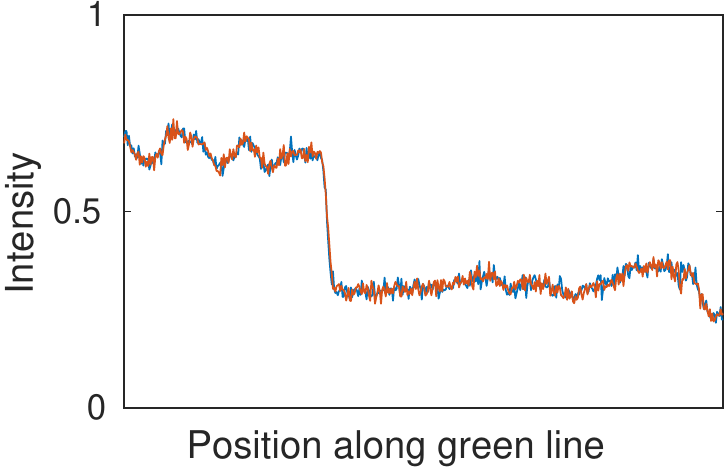}}
    \caption{DP data has significant brightness variations (a-c). We normalize it by making the two views have the same local mean and standard deviation. The normalization results in a reduction in brightness variations as shown by the two images (d-e) and the trace
    of intensities along the green line (f).}
    \label{fig:norm}
\end{figure}

The two DP views have significant brightness variations due to different amounts of lens shading (Fig.~\ref{fig:norm}(a-c)). To compensate for this, we first locally normalize each of the two images by computing their local mean and standard deviation in $9\times 9$ windows and then subtracting the mean and dividing by the standard deviation plus $0.001$ to prevent divide by zero errors. 

To visualize the effectiveness of our normalization, we multiply by the average local standard deviation of the two views and add the average local mean (Fig.~\ref{fig:norm}(d-f)). We use this visualization in the graphs in Fig.~6(b) of the main text.

\subsection{Disparity of Calibration Data}
The images captured during calibration are of a planar test target that occupies the entire field-of-view. Since we know the disparity should be smooth and have no depth-edges, we opt for the optical flow method of Liu~\shortcite{liu2009beyond} as it produces a smooth result with no discontinuities at color-edges. We normalize the DP images for lens shading and then apply Liu's method, modified to only compute horizontal flow, with an L2 smoothness loss function and an L1 data loss function. 

\section{Disparity To Depth Calculation}
We derive the relationship between disparity and depth here (Eq.~3 in the main text).
Disparity $d$ is approximately proportional to signed blur size $\bar b$. That is, there is some $\alpha$, such that
\begin{equation}
    d = \alpha \bar b.
    \label{eq:disparity_to_blur}
\end{equation}
We assume there is an object at distance $D$ from the main lens, which has focal length $f$, aperture size $L$ and is a distance $z_i$ away from the sensor. Let $z$ denote the focus distance.
By the thin lens equation, this object projects a sharp image a distance $D_i$ on the other side of the lens. The lens-sensor distance and the focal distance follow the same relationship. They are given by 
\begin{equation} \frac{1}{D_i} = \frac{1}{f} - \frac{1}{D} \text{  and  } \frac{1}{z_i} = \frac{1}{f} - \frac{1}{z}. \label{eq:thin_lens} \end{equation}
By similar triangles, the object produces a blur of size 
\begin{equation}\bar b = \frac{L(D_i - z_i)}{D_i} = L - \frac{L z_i}{D_i}. \label{eq:blur_size} \end{equation}
Combining Eqs.~\ref{eq:disparity_to_blur}-\ref{eq:blur_size}, we get that the disparity $d$ is related to inverse depth ($\nicefrac{1}{D}$) via a linear relationship
\begin{equation}
    d = \frac{\alpha Lf}{1-f/z}\left(\frac{1}{z}-\frac{1}{D}\right) \approx \alpha L f\left(\frac{1}{z}-\frac{1}{D}\right). \label{eq:depth_to_disparity}
\end{equation}
where the approximation follows because focus distance, $z$, is greater than $10$cm, while focal length, $f$, is roughly $5$mm for a typical smartphone lens.

\section{Controlled Experiments on DP Data}
%\rahul{I was wondering if we should position this section as "Advantages and Limitations of DP Data as source of Depth". Among advantages we can mention that no sync, calib, rectification issues unlike stereo, depth from same viewpoint as RGB image unlike ToF, narrow baseline allows for smaller search range and fewer occlusion issues, and the relationship between blur and disparity makes it ideal for synthetic depth-of-field application. Among limitations, we can mention aperture problem like stereo, doesn't work for large focus distances, calibration issues, and matching metrics like SSD and NCC fail for out of focus regions where effects due to kernel shape dominate}

%As in traditional stereo camera setups, DP imagery suffers from the aperture problem. That is, the shift between the two DP views occurs in only one dimension, and if the image content in a given patch is also oriented along that dimension (or if the image patch is flat) then disparity cannot be directly inferred for that patch.

\begin{figure}[t]
    \centering
        \includegraphics[width=\columnwidth]{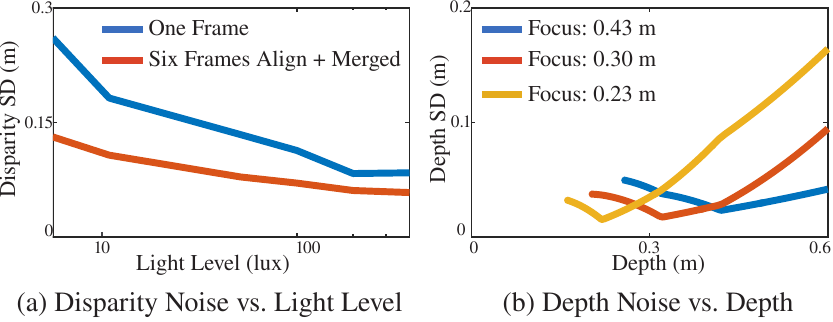}
    \caption{Controlled experiments on DP data. Aligning and merging DP data reduces disparity noise and depth noise is lowest for parts of the scene that are in-focus. (a): Disparity noise vs. light level for a focus distance of 0.95 m and a target distance of 0.88 m. (b): Depth standard deviation vs. depth for several focus distances.}
    \label{fig:pd_limits}
\end{figure}

We did a study to investigate the relationship between noise and light levels and disparity accuracy. We placed a textured, fronto-parallel target in front of a camera and took pictures while varying the overall light-level and focus distance. We only used the central fourth of the field of view to avoid calibration issues. Because the disparity should be constant across all images, we can use the standard deviation of the disparity over the field-of-view as a measure of error (Fig.~\ref{fig:pd_limits}). Since disparities within each tile are correlated, we only used one disparity value per tile.

We found that disparities values from six frames aligned and averaged are two times less noisy for low-light scenes (5 lux) and 1.5 times less noisy for brighter scenes (Fig.~\ref{fig:pd_limits}(a)) compared to disparity values from a single frame. We also found that depth noise is lowest when the target is in focus (Fig.~\ref{fig:pd_limits}(b)). This demonstrates an additional challenge versus wide-baseline stereo, which can rely on coarse image content and is not as sensitive to regions that have less texture due to being out-of-focus.

\section{Blur Parameters}

Here we list the values of the parameters we use in our rendering. These were tuned by visualizing the rendered results on thousands of images, and mostly affect aesthetics like the strength of blur and the depth of field. Our system is not very sensitive to the particular value of these parameters.

The scale factor $\kappa(\cdot)$ (Eq. 5 in main text) is expressed as a multiplication of three different factors, $s$ that scales the blur radius in proportion to the image size and is a constant in our system, $\kappa_1(z)$ that amplifies computed disparities when using a larger focus distance, and $\kappa_2(d_{\mathit{focus}})$ that corrects the scale factor according to the amount of misfocus. It implies that $\kappa(\cdot)$ depends on both $d_{\mathit{focus}}$ and $z$:
\begin{equation}
\kappa(z, d_{\mathit{focus}}) = s\, \kappa_1(z) \kappa_2(d_{\mathit{focus}})
\end{equation}
We set
\begin{equation}
\kappa_1(z) = \left| (0.33z + 0.17) \right|_{\left[1, 3.5\right]}
\end{equation} where $z$ is in meters and  $(\cdot)|_{\left[a,b\right]}$ denotes clamping to the range $\left[a,b\right]$.
We set $\kappa_2(\cdot)$ to be
\begin{equation}
\kappa_2(d_{\mathit{focus}}) = \left| \frac{1}{1 + \nicefrac{d_{\mathit{focus}}}{2}}\right|_{\left[0.5, 2\right]}
\end{equation}
Abrupt transitions may occur near the boundaries of the segmentation mask, especially for frontal objects, e.g., a person resting their hands on the table. To ameliorate artifacts at such transitions, we reduce the amount of blur for the frontal part of the scene, i.e., where $d(\mathbf{x}) - d_{\mathit{focus}} > 0$, by multiplying $\kappa(\cdot)$ by $0.6$ for such pixels.

As described in the main text, we keep pixels with disparities 
$[d_\mathit{focus}-d_\emptyset,d_\mathit{focus}+d_\emptyset]$ unblurred.
We would like to make this range independent of changes in disparity as well, e.g.,
use a smaller range for larger focus distances since the range of disparities is smaller.
We achieve this by setting $d_\emptyset = \nicefrac{0.56}{\kappa(z, d_{\mathit{focus}})}$.
For photos with people, combining disparity with the segmentation mask already achieves the effect of bringing the entire subject into focus, so we use a smaller value of $d_\emptyset=\nicefrac{0.19}{\kappa(z, d_{\mathit{focus}})}$ pixels.

The blur radius, $r_\mathit{brute}$, at which we transition between brute force convolution and our accelerated gradient domain blur, is set to  $2.75$ pixels (at the blur's working resolution, so $5.5$ pixels at the original image resolution). We find this value to be a good balance between the higher computational cost of brute force blurs and the higher approximation error introduced when discretizing smaller kernels (e.g., a rasterized circle with radius $2$ is just a $3\times3$ pixel square).

\section{Noise generation}

As mentioned in the paper, producing natural looking synthetically defocused images requires that we inject noise into the blurred regions of our output rendering.
Unfortunately, the complexity of our camera's imaging system and its post-processing pipeline makes it difficult to describe the naturally occurring noise analytically. This makes it challenging to construct an accurate generative noise model  (e.g., Gaussian noise), and we instead adopt a non-parametric approach. We start by collecting photos of a white computer monitor through a diffuser. We then high-pass filter these images and blend them to produce small, periodic noise patches, which are added to the defocused regions of our output renderings. This procedure is illustrated in Fig.~\ref{fig:noise_patch_synthesis}.

\newcommand{\diskwidth}{0.22\columnwidth}
\begin{figure}[t]
    \centering
    \subfigure[][Raw image]{
    \label{fig:noise_patch_input}
    \includegraphics[width=\diskwidth]{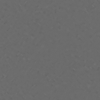}
    }
    \subfigure[][High-passed noise]{
    \label{fig:noise_patch_highpass}
    \includegraphics[width=\diskwidth]{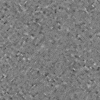}
    }
    \subfigure[][Feathered noise]{
    \label{fig:noise_patch_feathered}
    \includegraphics[width=\diskwidth]{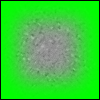}
    }
    \subfigure[][Overlapped feathered patches]{
    % 0.543 is 1.81 (ratio of this image size to above ones) * 0.3 (see def of diskwidth).
    \label{fig:noise_patch_overlap}
    \includegraphics[width=0.543\columnwidth]{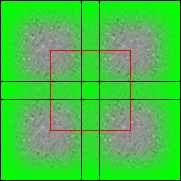}
    }
    \subfigure[][Periodic noise]{
    \label{fig:noise_patch_result}
    \includegraphics[width=\diskwidth]{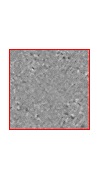}
    }
    \caption{Periodic noise construction steps. \subref{fig:noise_patch_input} Capture a raw image of a uniform white target and extract an $L \times L$ patch from the center. \subref{fig:noise_patch_highpass} Apply a high-pass filter to isolate the noise. \subref{fig:noise_patch_feathered} Feather the alpha channel near the edges of the patch, leaving an $l \times l$ region mostly opaque. \subref{fig:noise_patch_overlap} Replicate \subref{fig:noise_patch_feathered} four times in an overlapping pattern, with a spacing of $l$. \subref{fig:noise_patch_result} Crop out the center $l \times l$ region and normalize by the alpha channel to get a noise pattern with period $lß$.}
    \label{fig:noise_patch_synthesis}
    \vspace{-6pt}
\end{figure}

\section{User Study}

Our user study was designed similarly to the user study performed in Barron \etal~\shortcite{barron2015fast}.
We asked $6$ people who are not affiliated with this paper to compare the output of our ``DP + Segmentation'' algorithm, its ``Segmentation Only'' and ``DP Only'' ablations, and the techniques of Barron \etal~\shortcite{barron2015fast} and Shen \etal~\shortcite{shen2016automatic}. The images used in this study were a random sampling of the casual photographs taken by someone who is not affiliated with this paper, nor a participant in the user study. These images contain a mixture of shots with one or more human subjects and with non-human subjects (which may still incidentally contain a human).
To ascertain the user's preference, for each scene we first presented two (randomly selected) algorithm's renderings of the same scene side-by-side and asked the user to select the rendering they preferred. All participants were instructed to select the rendering they would prefer to have been produced by their camera. The winning algorithm's rendering was then presented side-by-side with another (randomly selected) algorithm's rendering of the same scene, and this process was repeated until all algorithms had been compared. For each scene, the order in which algorithms were presented to the user was randomized, and each algorithm was presented ``blind'' with the participant having no way to associate the renderings being compared with a specific algorithm or a past rendering. 

When presenting each rendering to the user, we presented the entire rendering alongside three cropped sub-images of the rendering, which (similar to the procedure in Barron \etal~\shortcite{barron2015fast}) were algorithmically selected as the three non-overlapping $128 \times 128$ sub-images with the largest inter-image variance across the five algorithms, that also lay on the boundary of the human subject (if present). Presenting images in this way allowed the participant to evaluate the composition of the scene holistically, while also allowing them to easily compare detail on the human subject's silhouette.

Images from our user study, presented in the same format as was used in our user study, can be found in Figs.~\ref{fig:userstudy1} and \ref{fig:userstudy2}.

\newcommand{\studyfigwidth}{0.18\textwidth}

\begin{figure*}
    \centering
    \subfigure[Barron \etal~\shortcite{barron2015fast}]{
    \begin{tabular}{@{}c@{}}
    \includegraphics[width=\studyfigwidth]{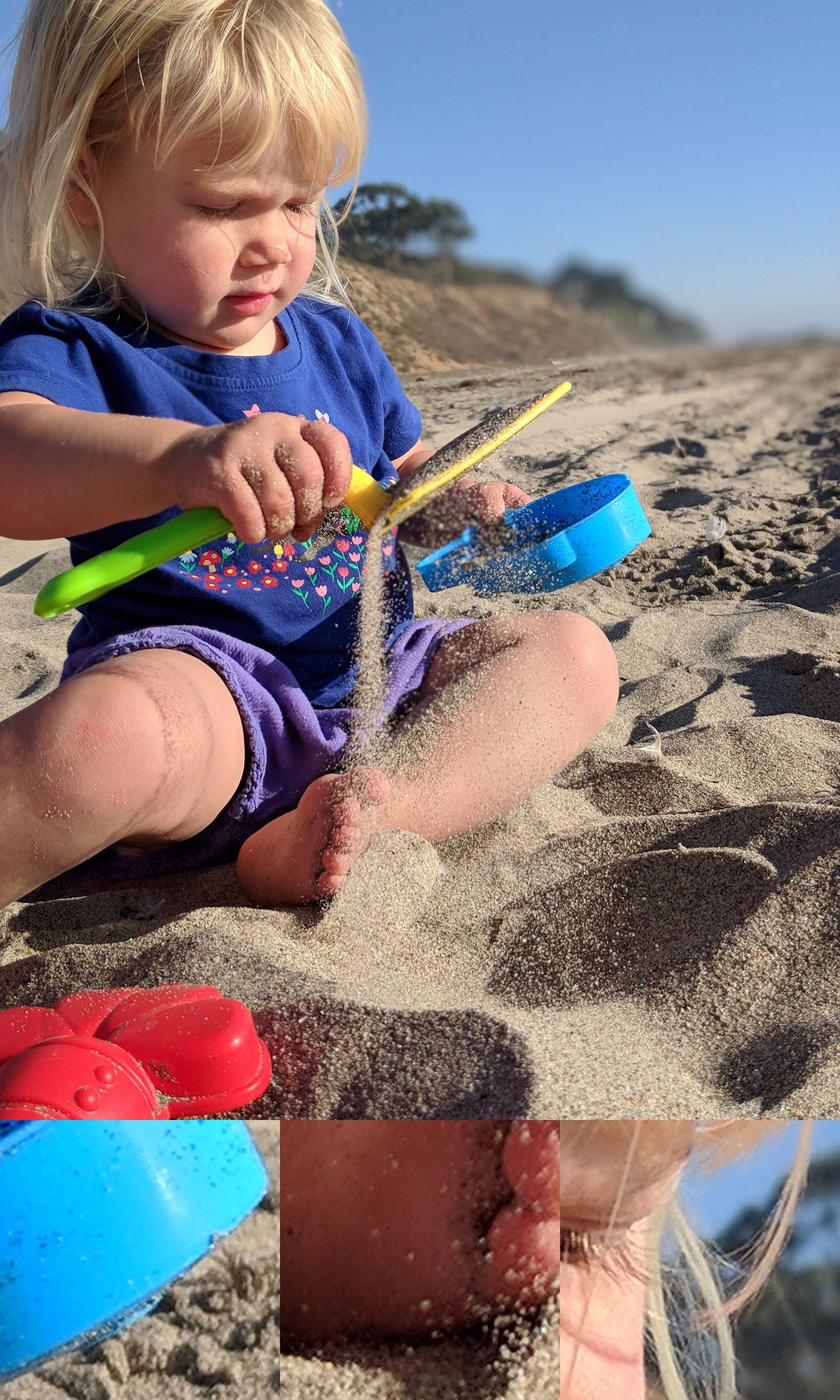} \\
    \includegraphics[width=\studyfigwidth]{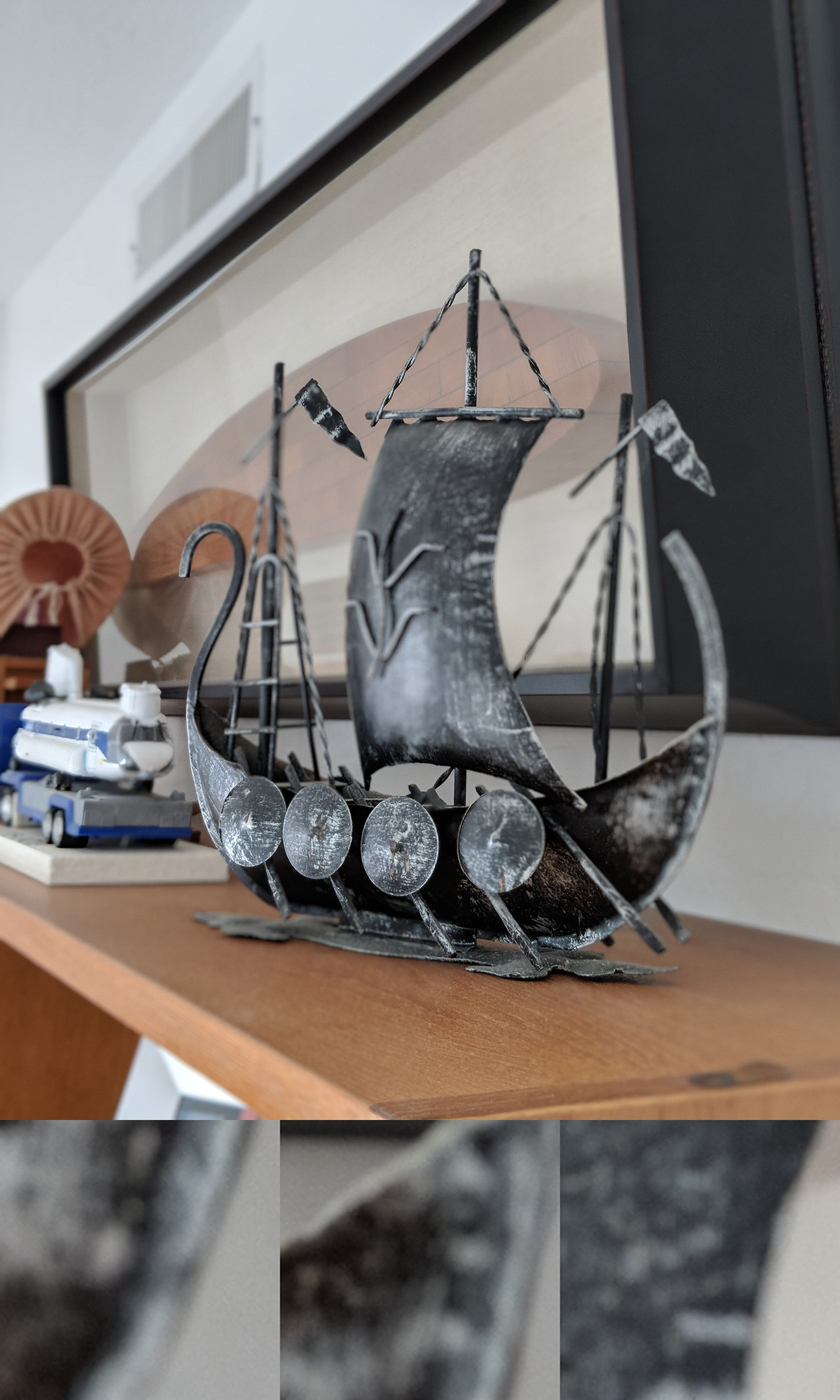} \\
    \includegraphics[width=\studyfigwidth]{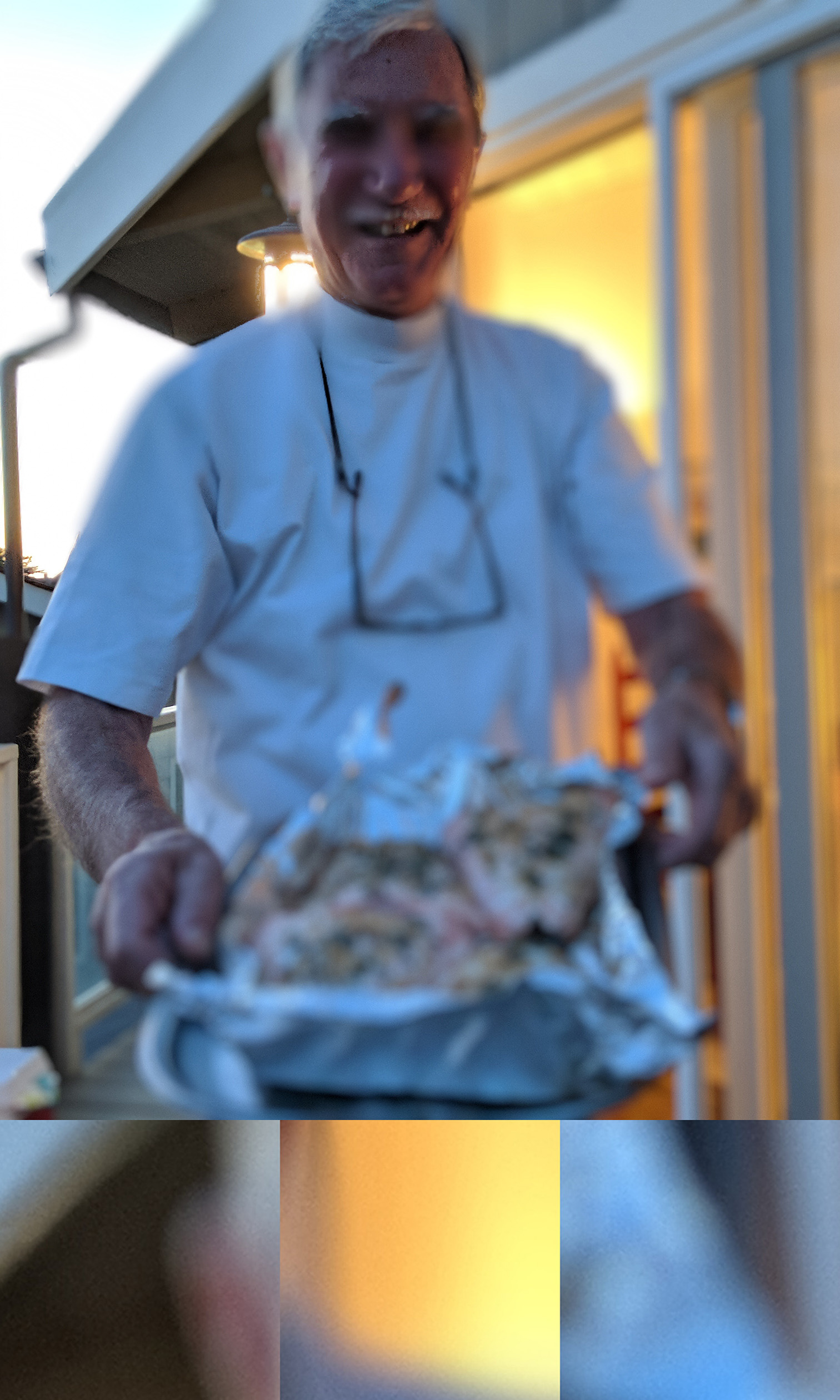} \\
    \includegraphics[width=\studyfigwidth]{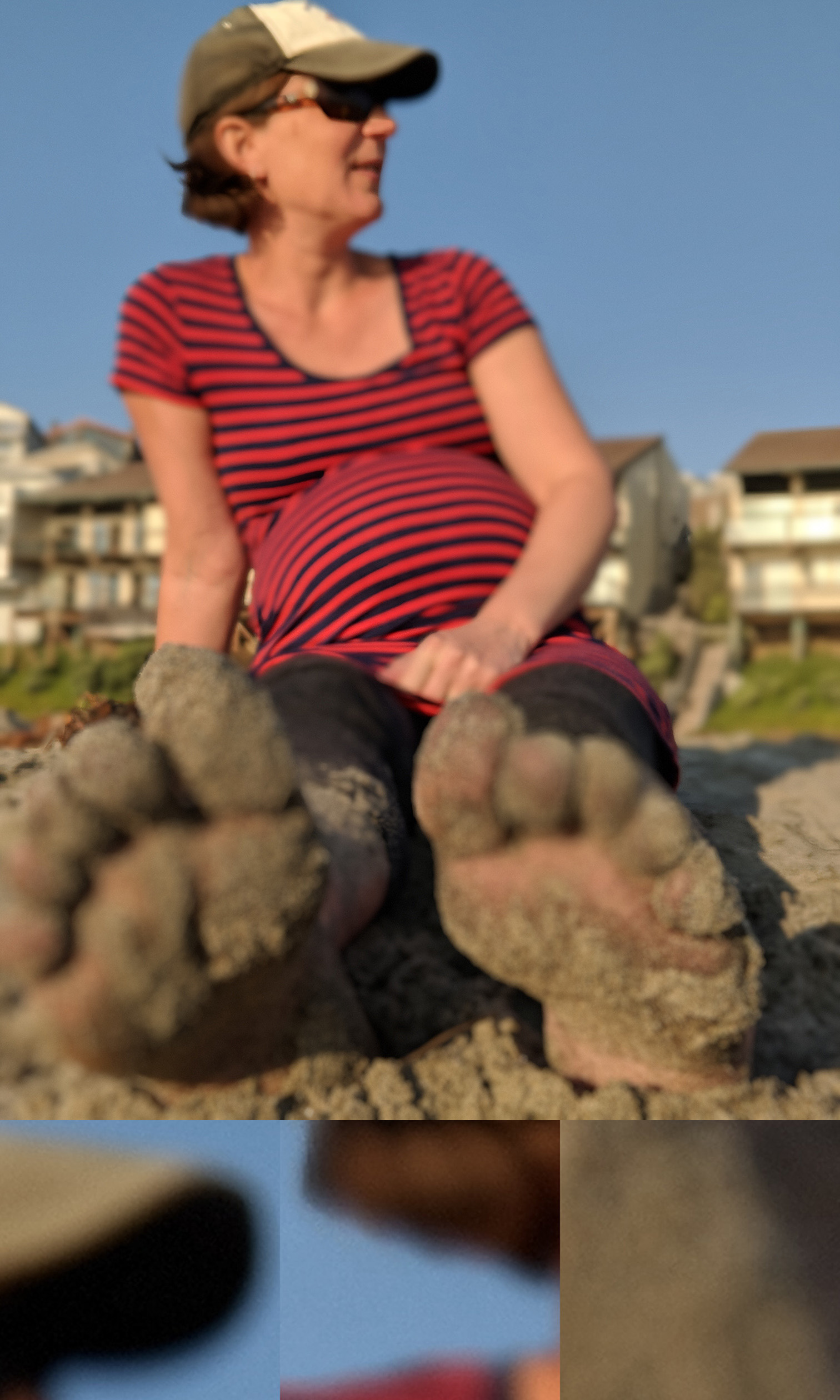}
    \end{tabular}
    }
    \subfigure[Shen \etal~\shortcite{shen2016automatic}]{
    \begin{tabular}{@{}c@{}}
    \includegraphics[width=\studyfigwidth]{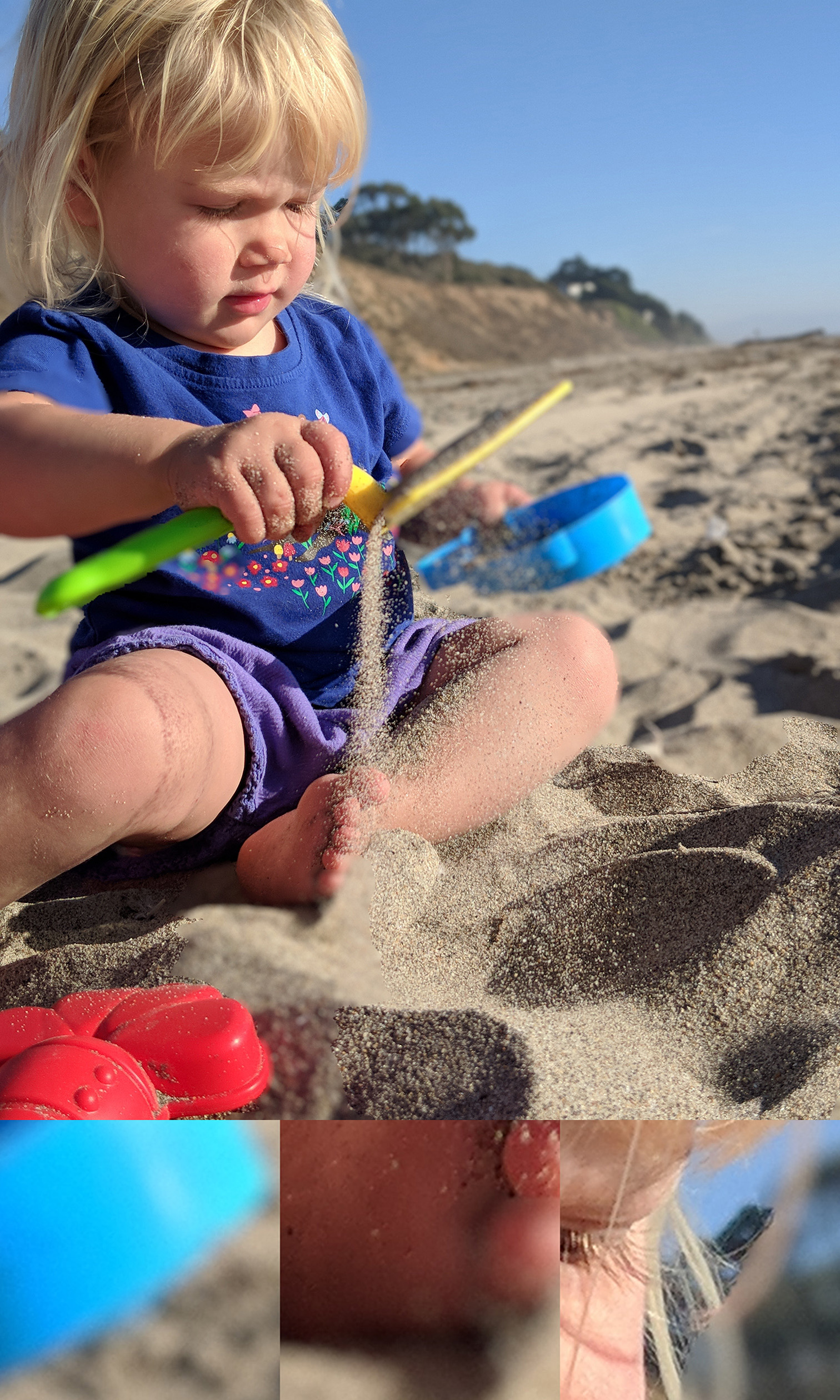} \\
    \includegraphics[width=\studyfigwidth]{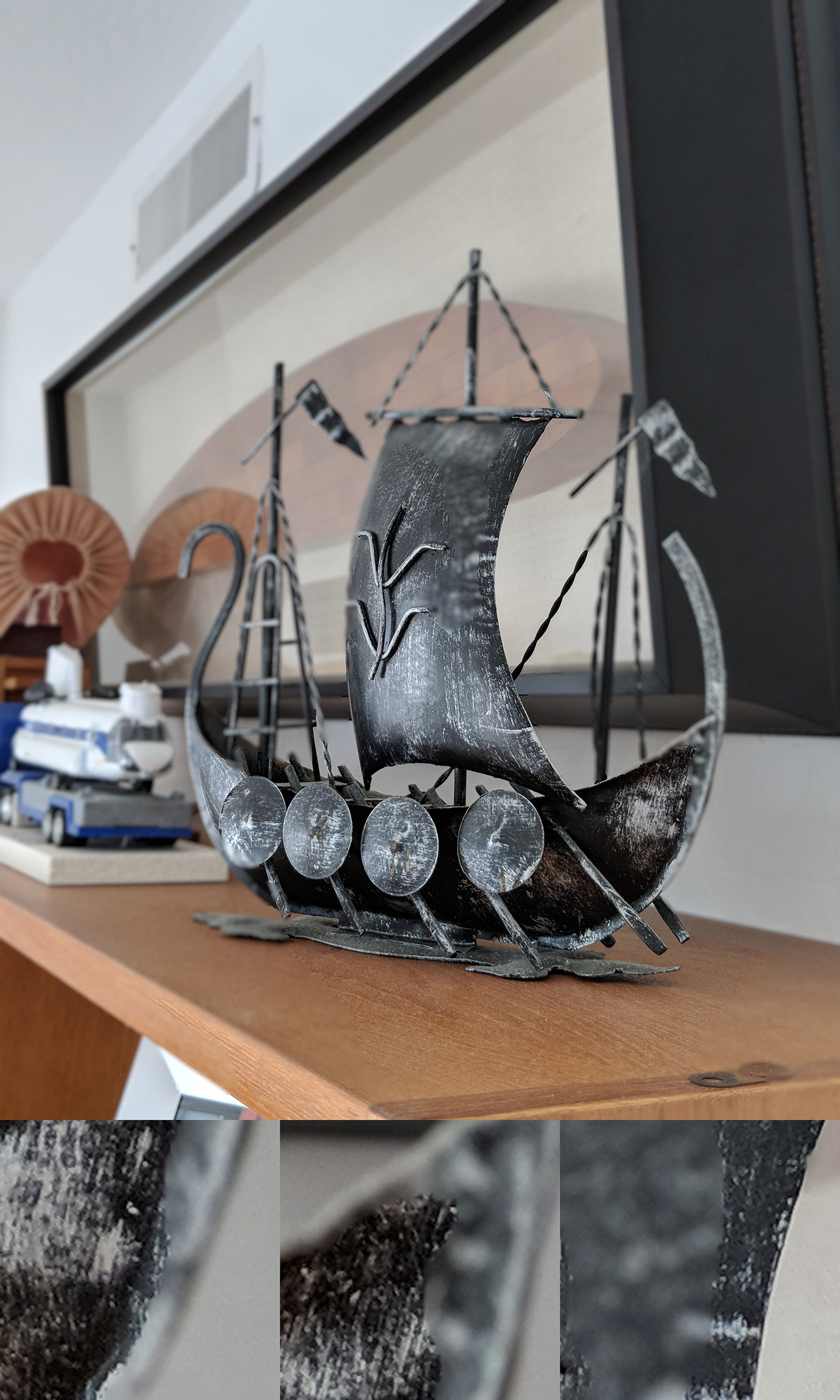} \\
    \includegraphics[width=\studyfigwidth]{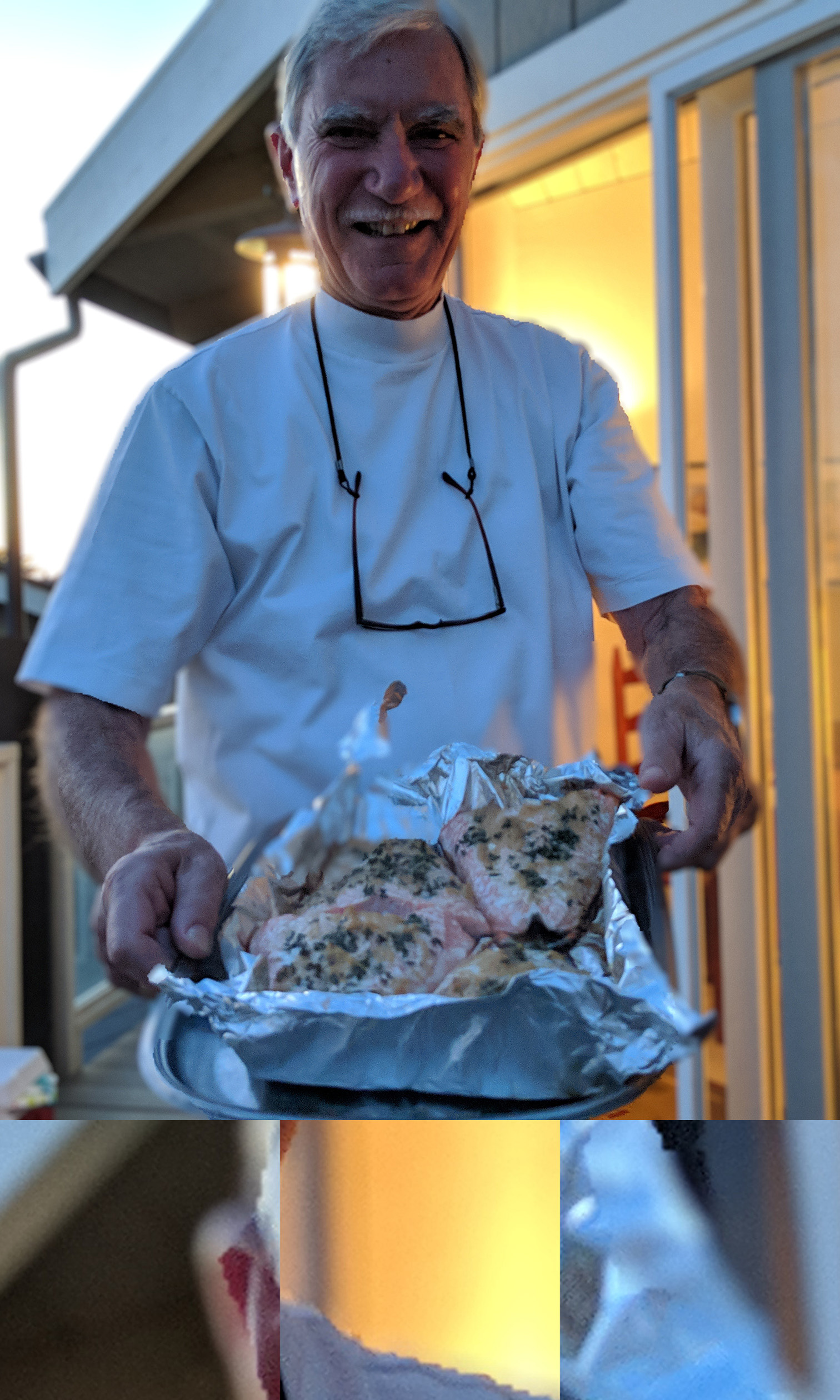} \\
    \includegraphics[width=\studyfigwidth]{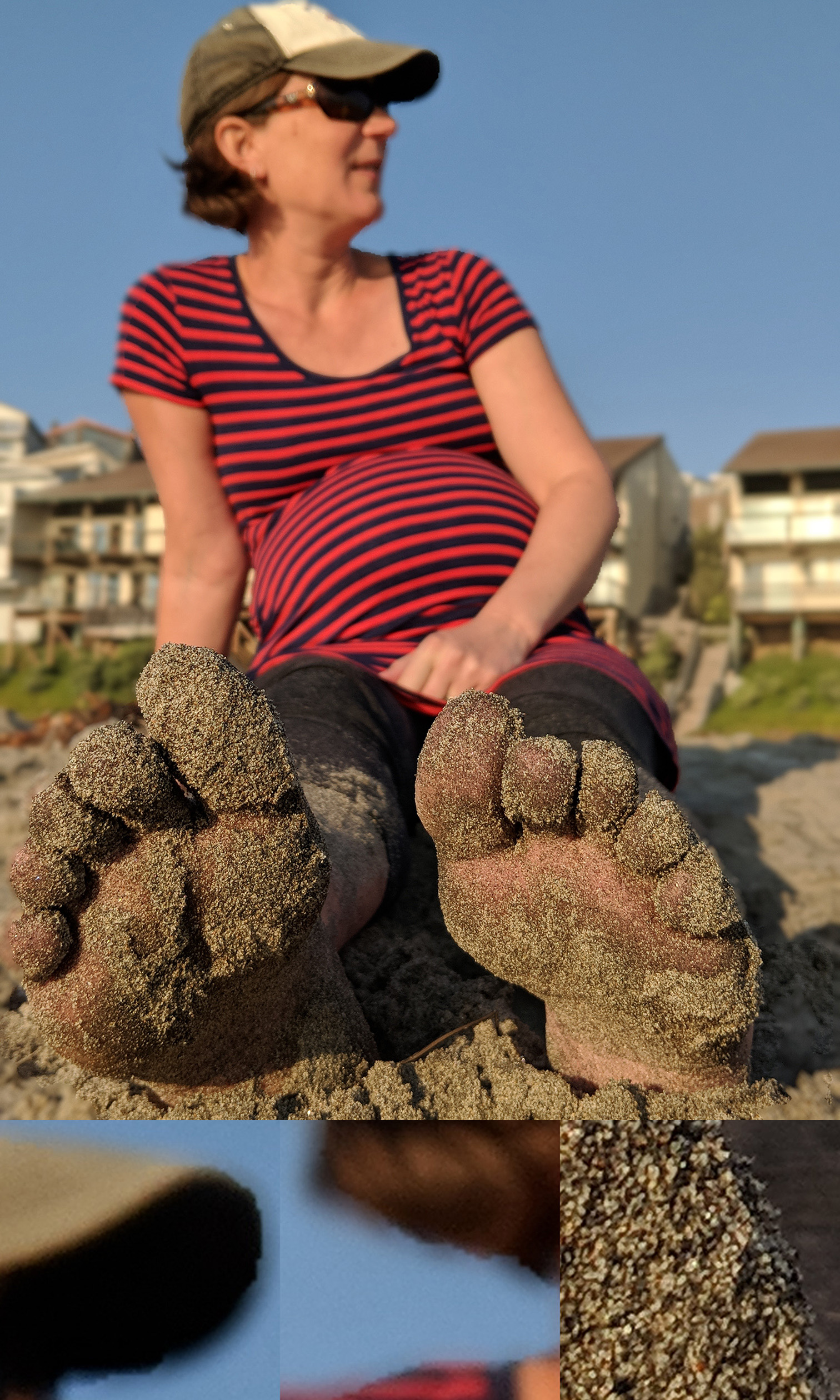}
    \end{tabular}
    }
    \subfigure[DP Only (ours)]{
    \begin{tabular}{@{}c@{}}
    \includegraphics[width=\studyfigwidth]{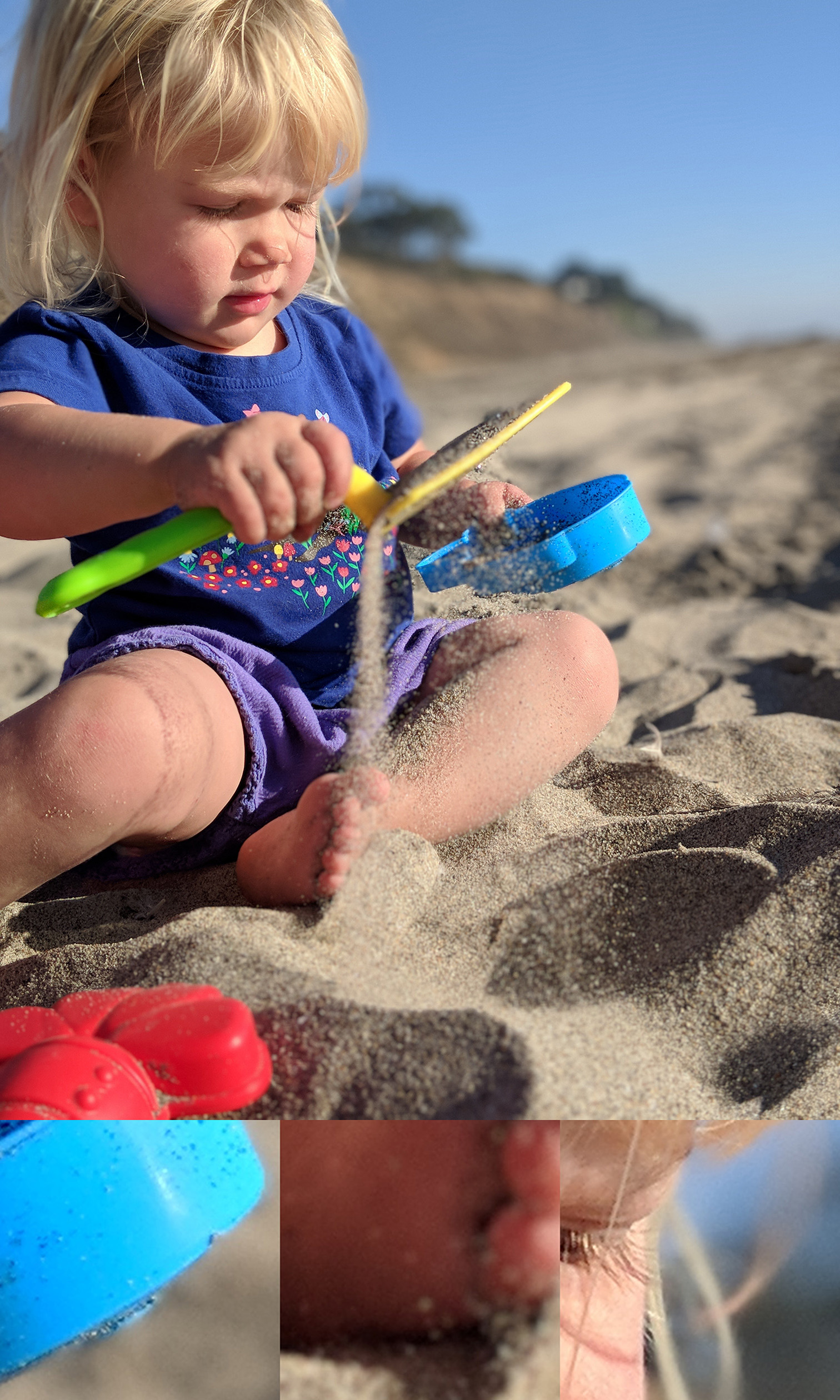} \\
    \includegraphics[width=\studyfigwidth]{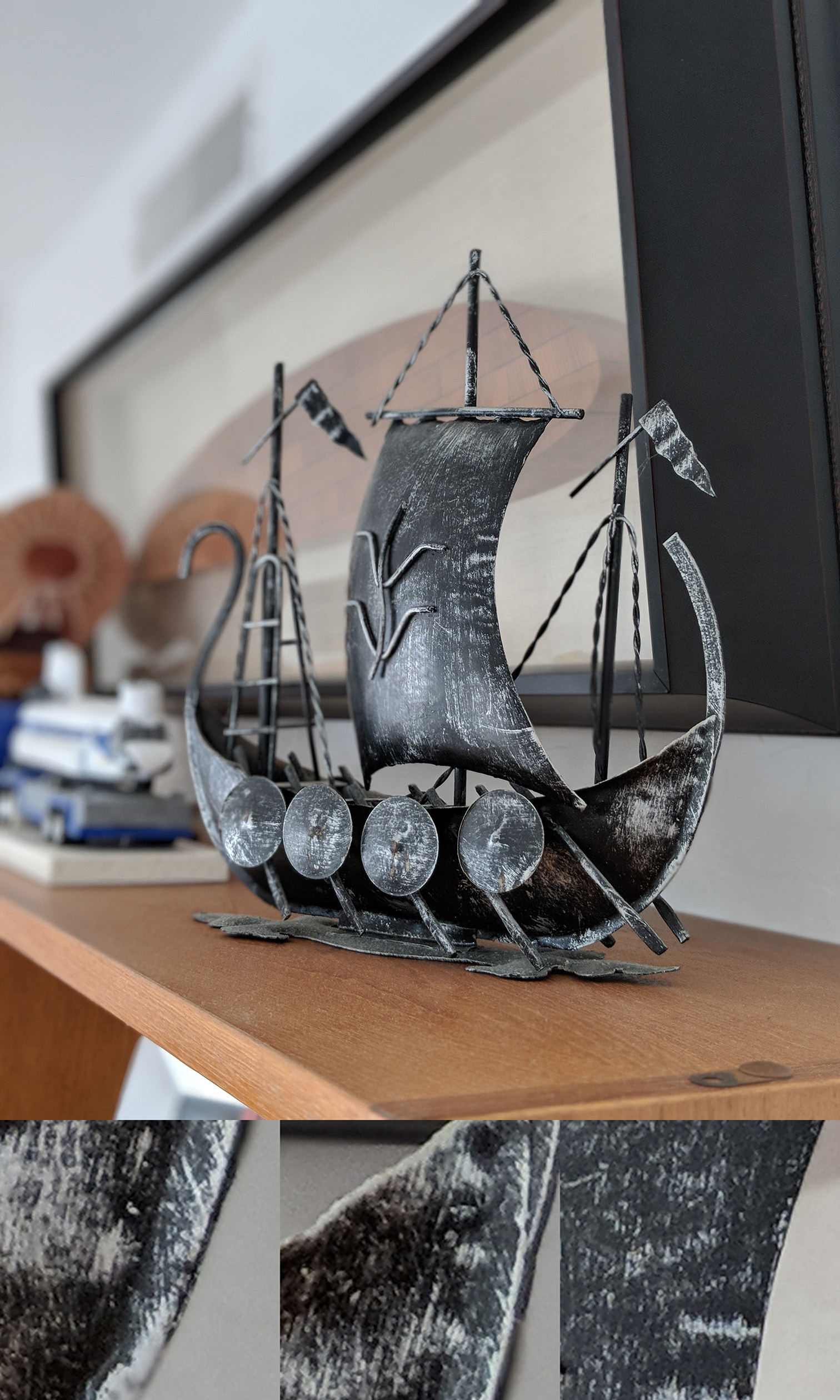} \\
    \includegraphics[width=\studyfigwidth]{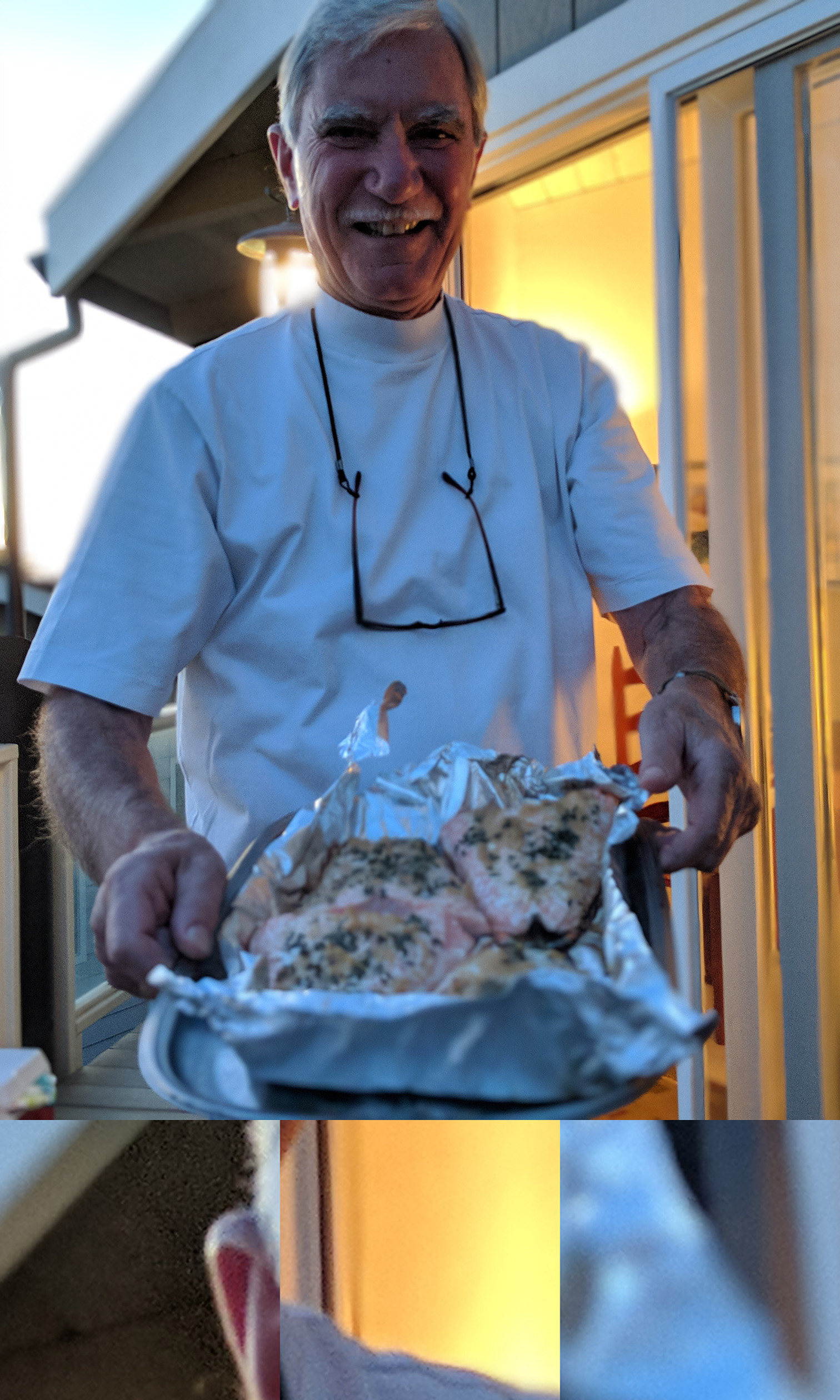} \\
    \includegraphics[width=\studyfigwidth]{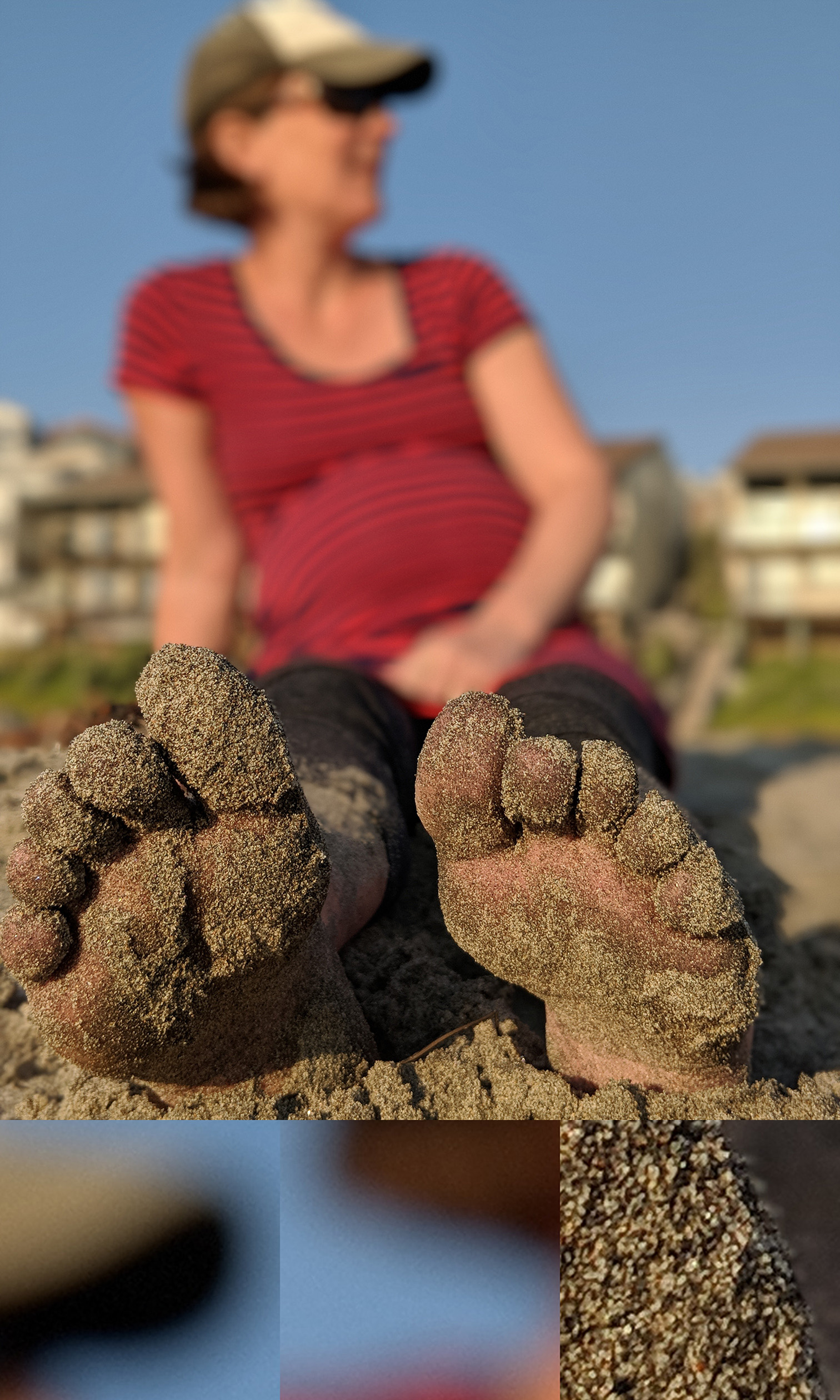}
    \end{tabular}
    }
    \subfigure[Segmentation Only (ours)]{
    \begin{tabular}{@{}c@{}}
    \includegraphics[width=\studyfigwidth]{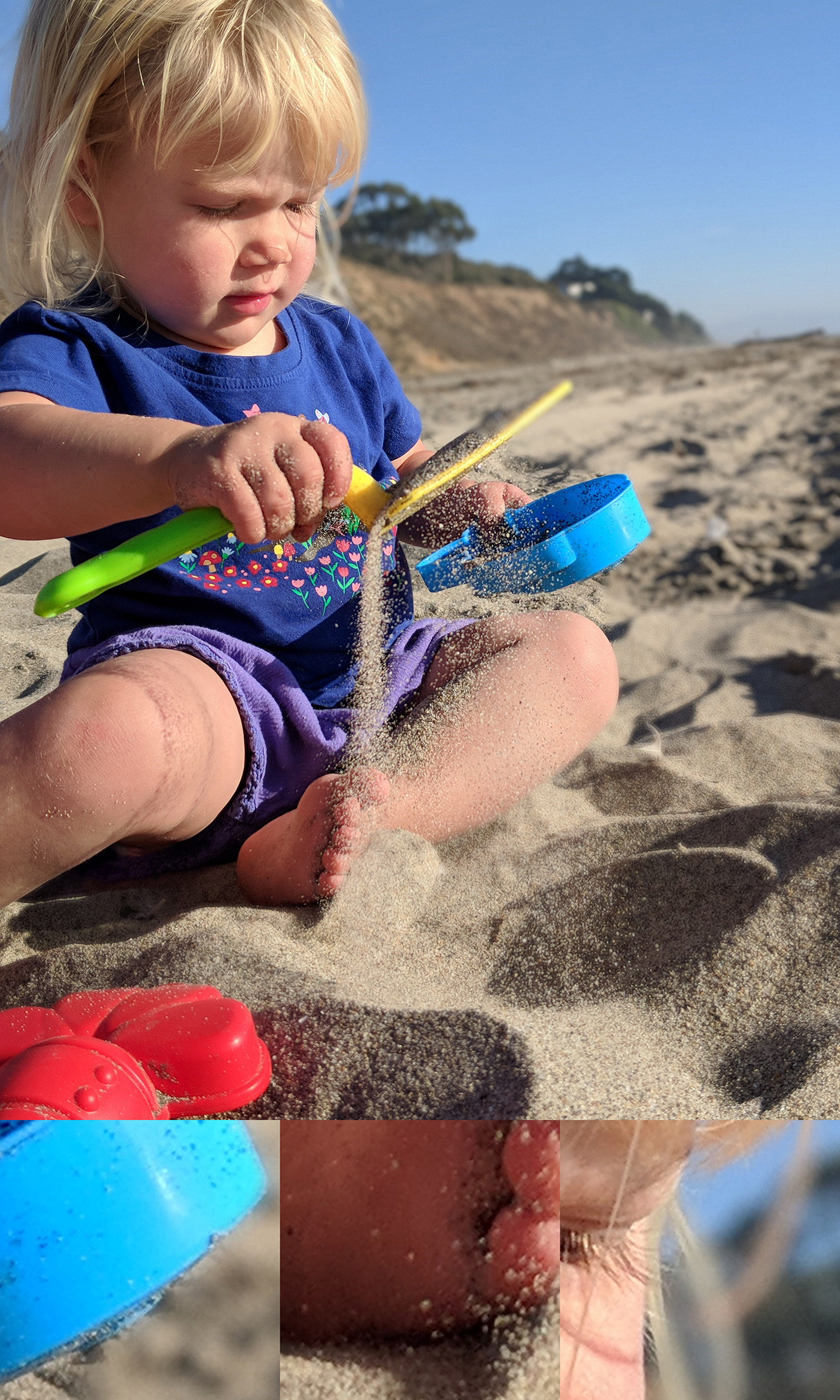} \\
    \includegraphics[width=\studyfigwidth]{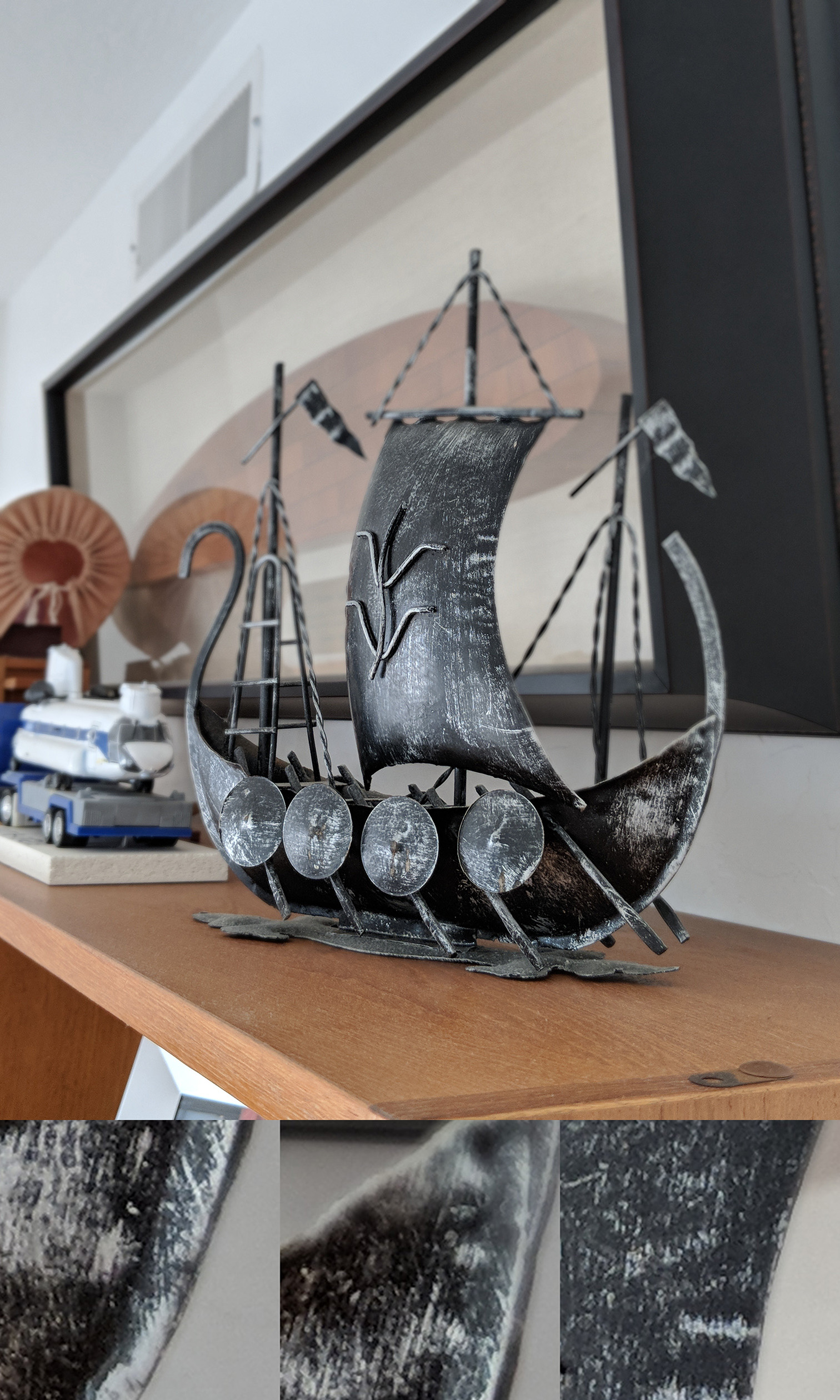} \\
    \includegraphics[width=\studyfigwidth]{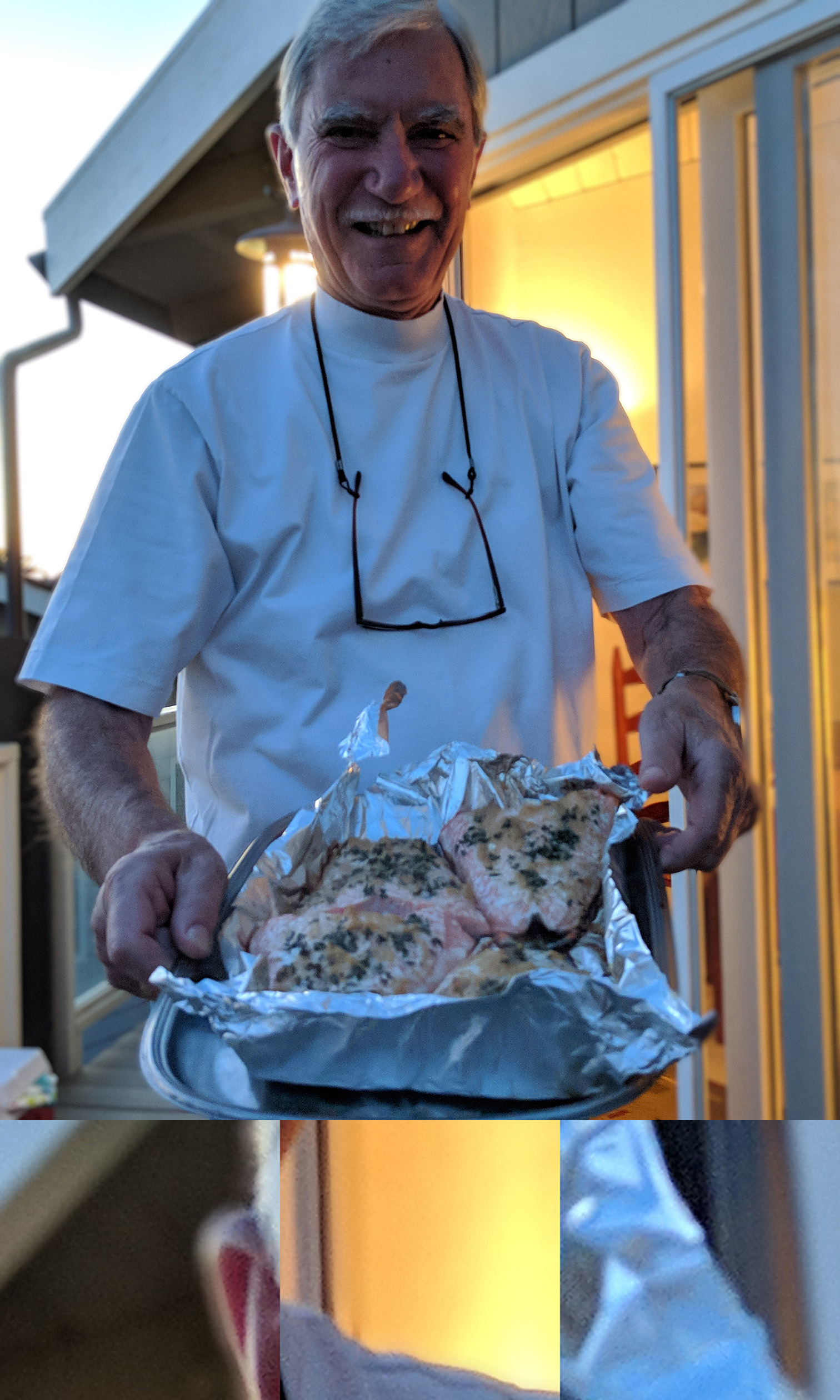} \\
    \includegraphics[width=\studyfigwidth]{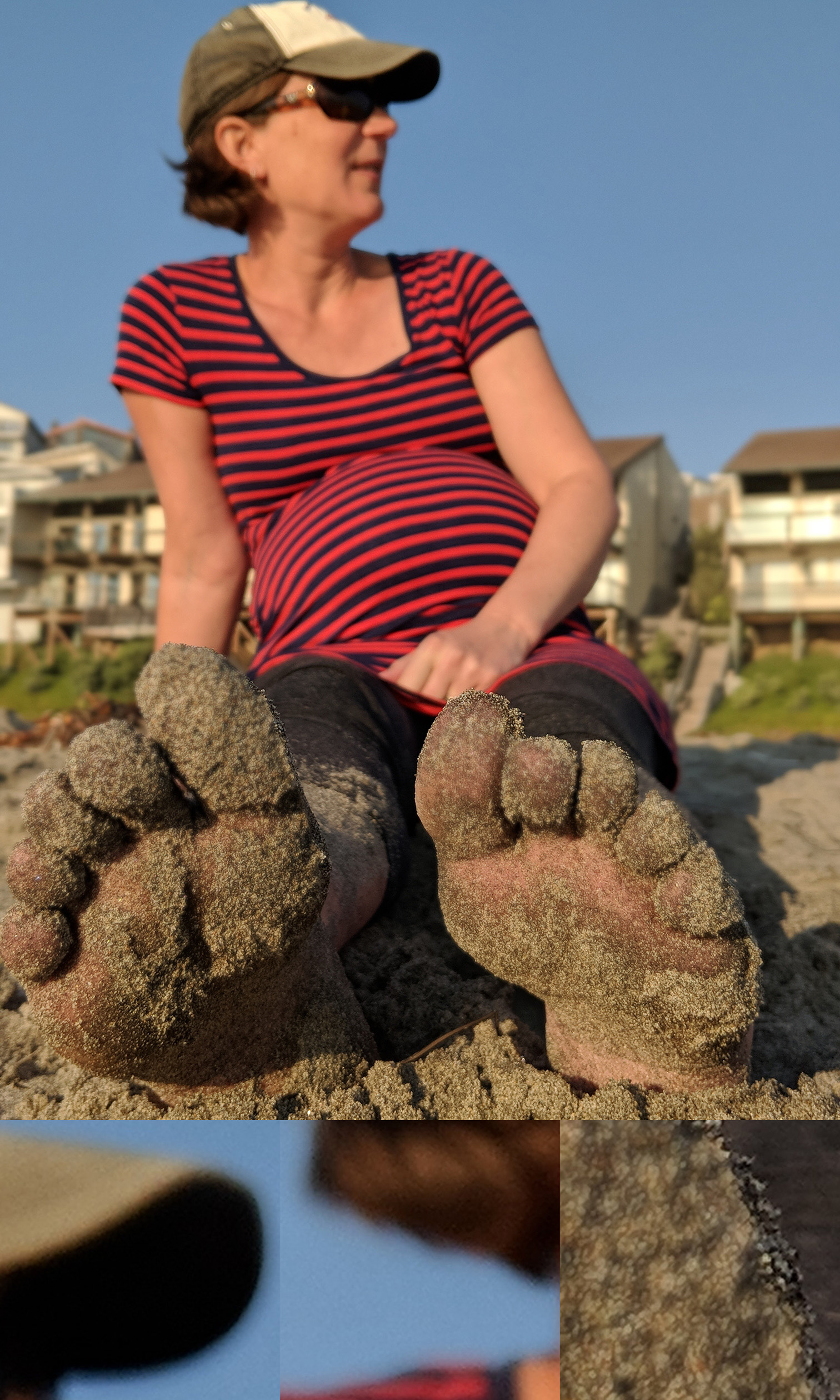}
    \end{tabular}
    }
    \subfigure[DP + Segmentation (ours)]{
    \begin{tabular}{@{}c@{}}
    \includegraphics[width=\studyfigwidth]{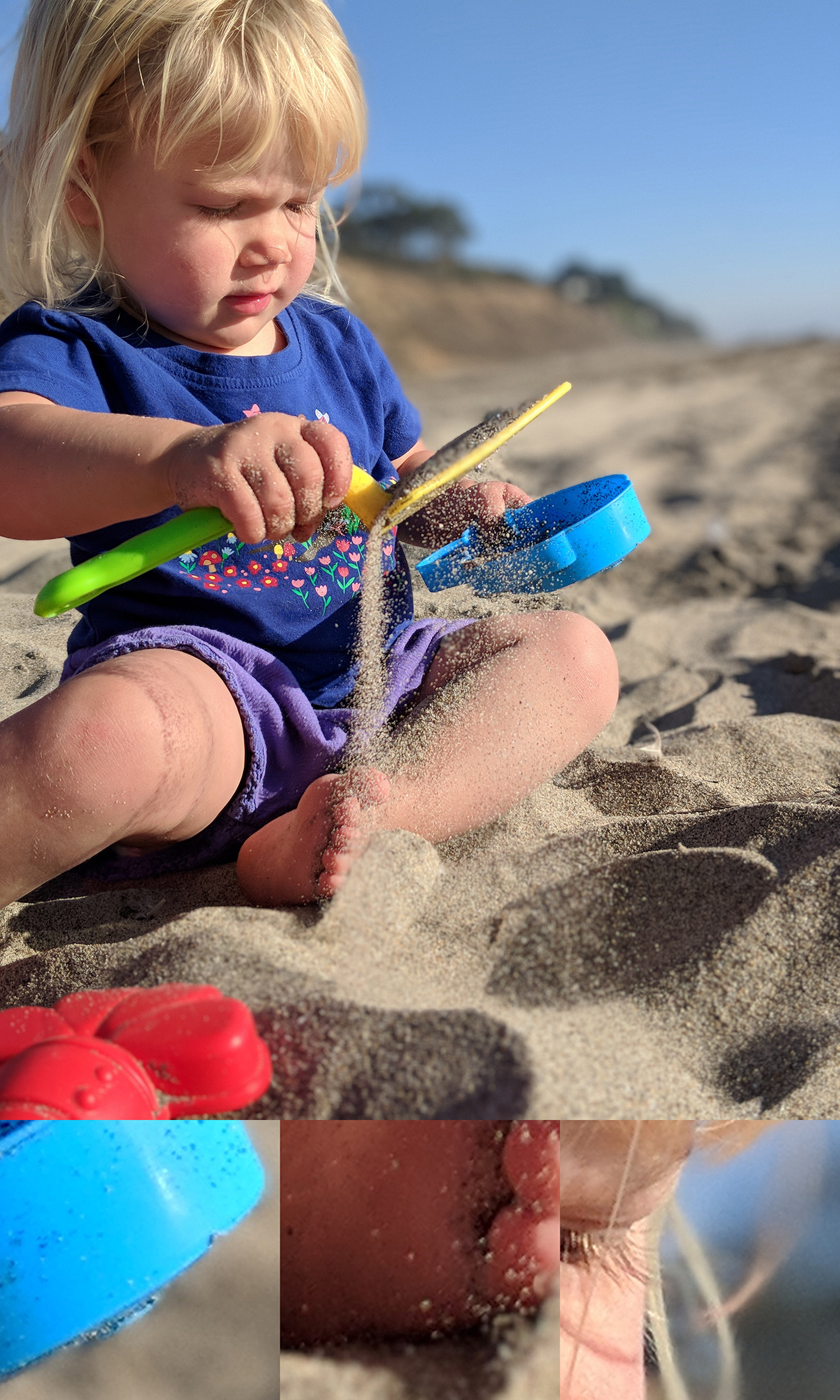} \\
    \includegraphics[width=\studyfigwidth]{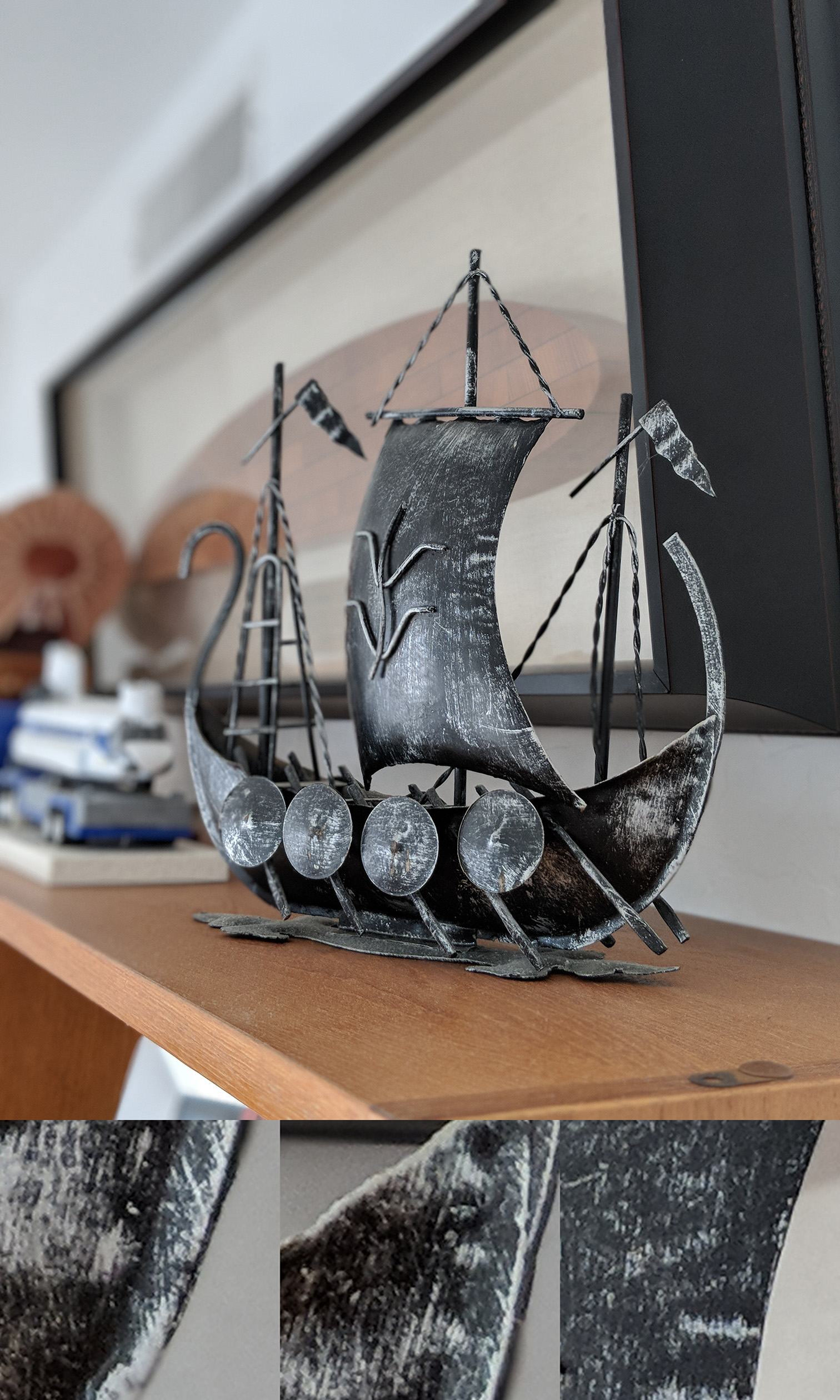} \\
    \includegraphics[width=\studyfigwidth]{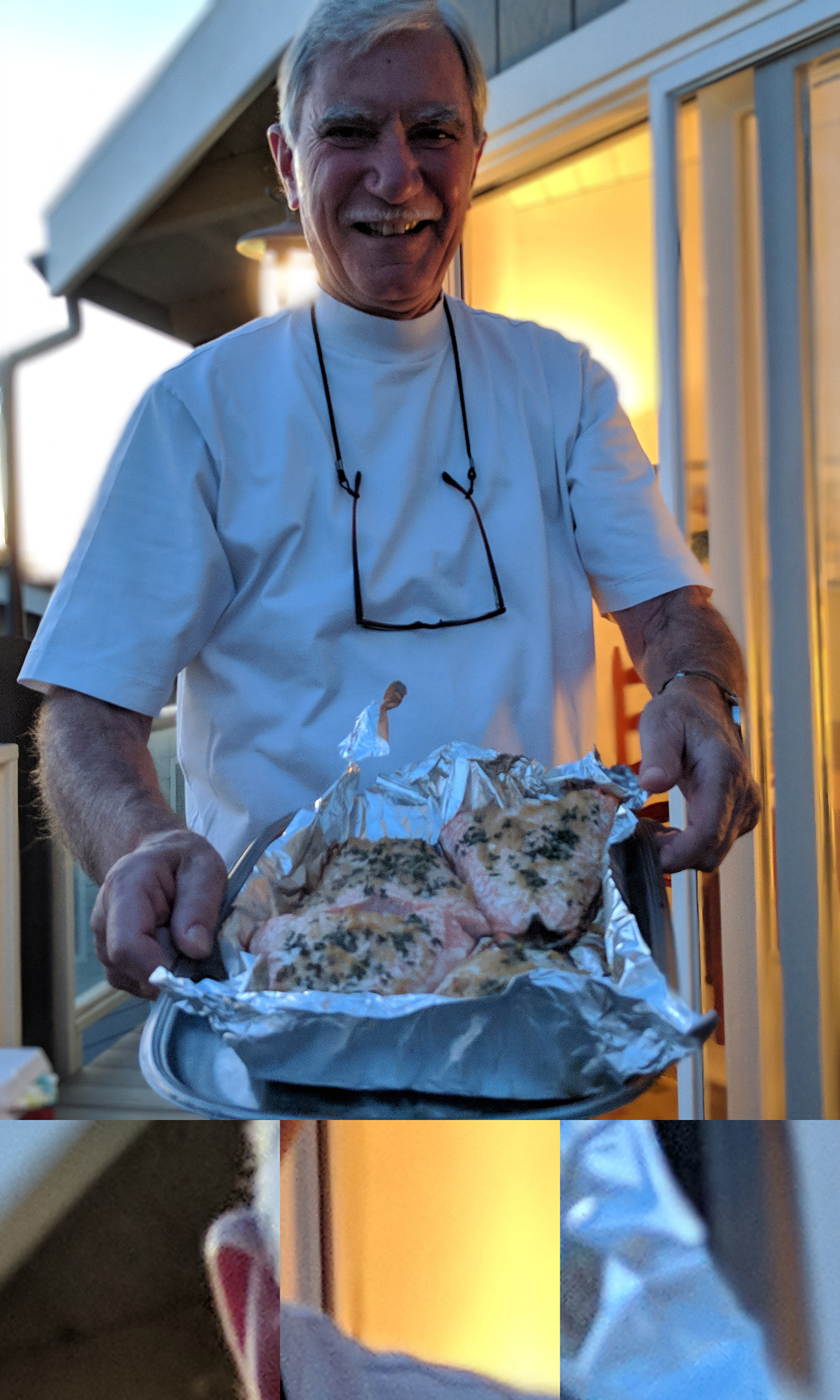} \\
    \includegraphics[width=\studyfigwidth]{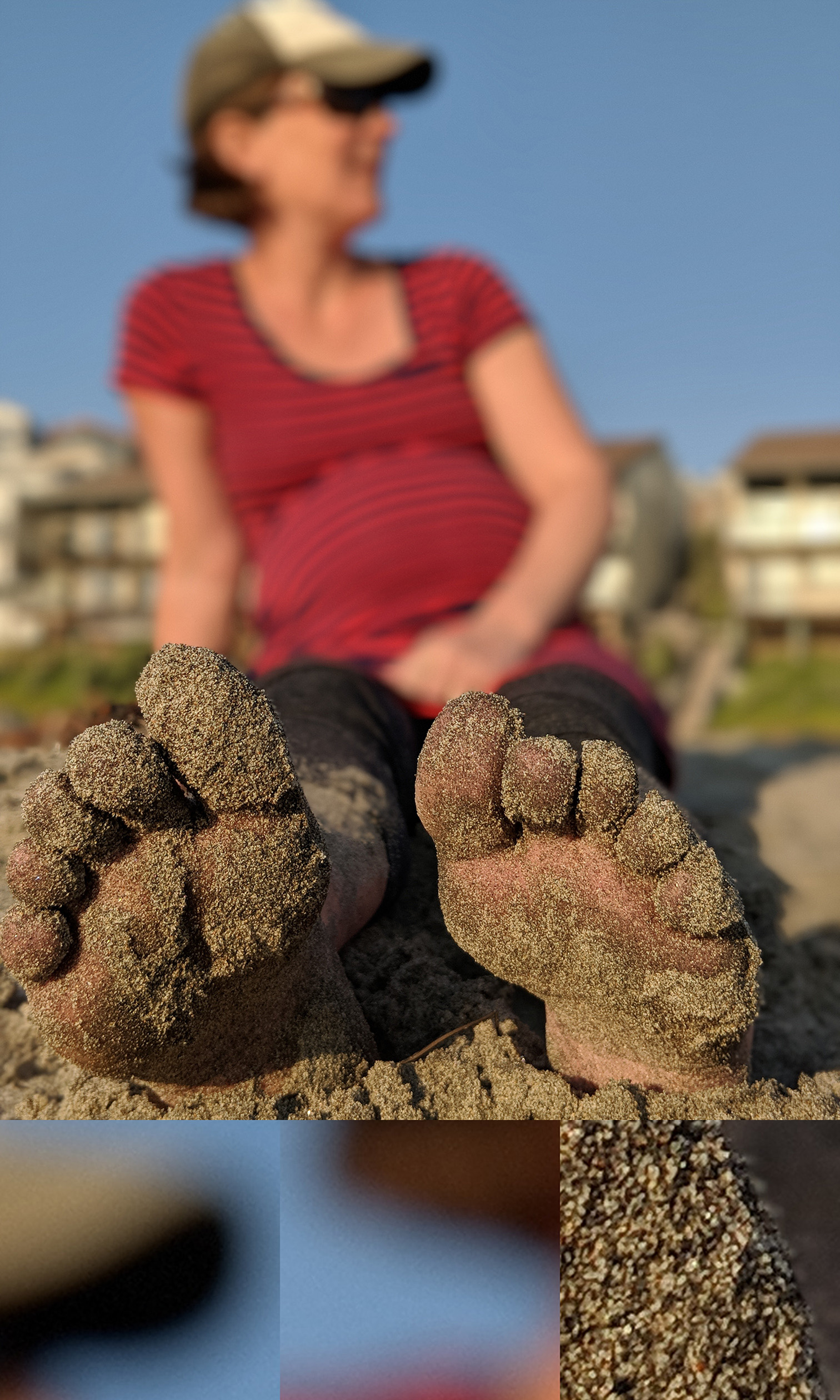}
    \end{tabular}
    }
    \caption{A sample of the images used during our user study, processed by our model (``DP + Segmentation''), two ablations of our model (``DP Only'' and ``Segmentation Only'') and two baseline algorithms. With each image we present the three $128 \times 128$ subimages with the largest total variance across all five output images that also lay on the boundary of the human subject (if present).}
    \label{fig:userstudy1}
\end{figure*}

\begin{figure*}
    \centering
    \subfigure[Barron \etal~\shortcite{barron2015fast}]{
    \begin{tabular}{@{}c@{}}
    \includegraphics[width=\studyfigwidth]{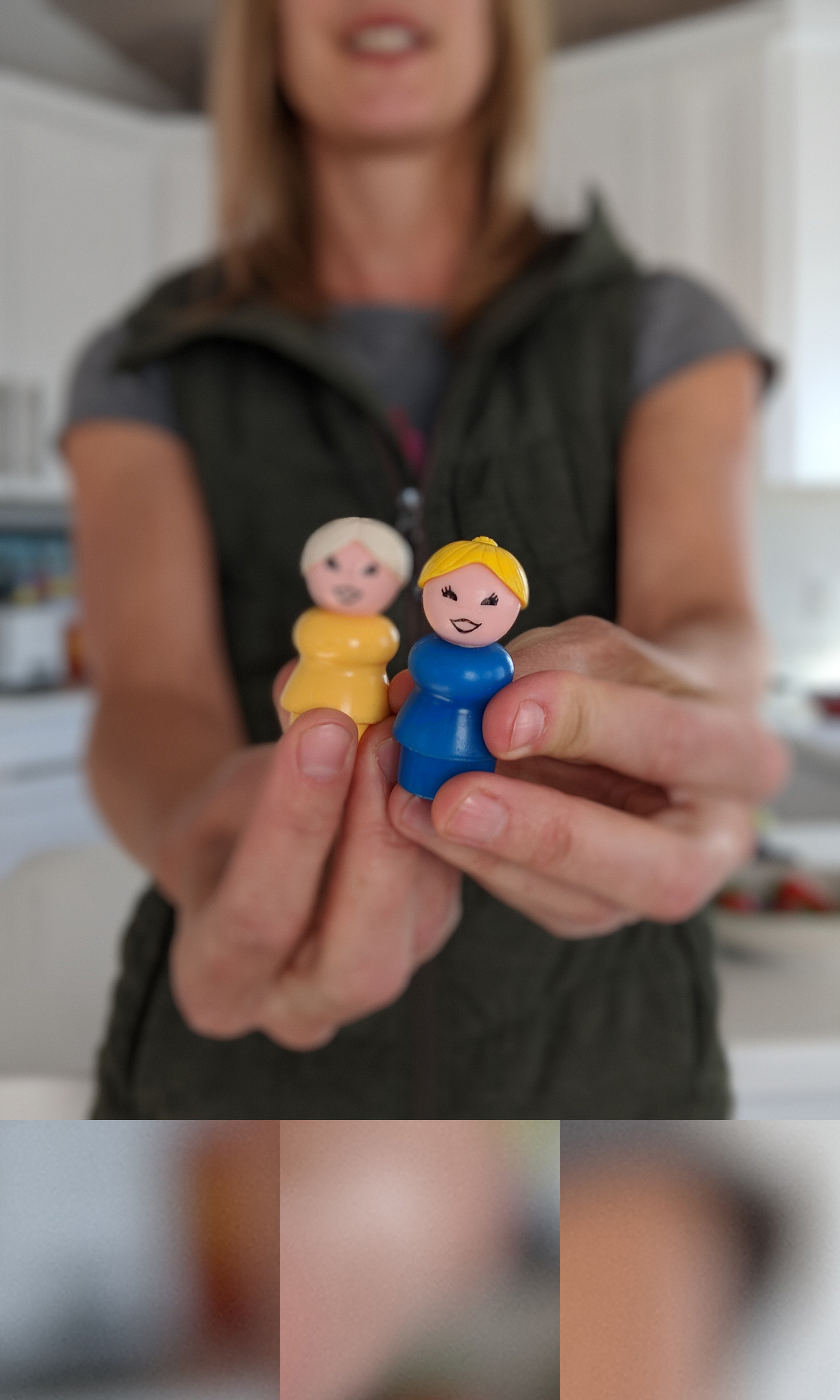} \\
    \includegraphics[width=\studyfigwidth]{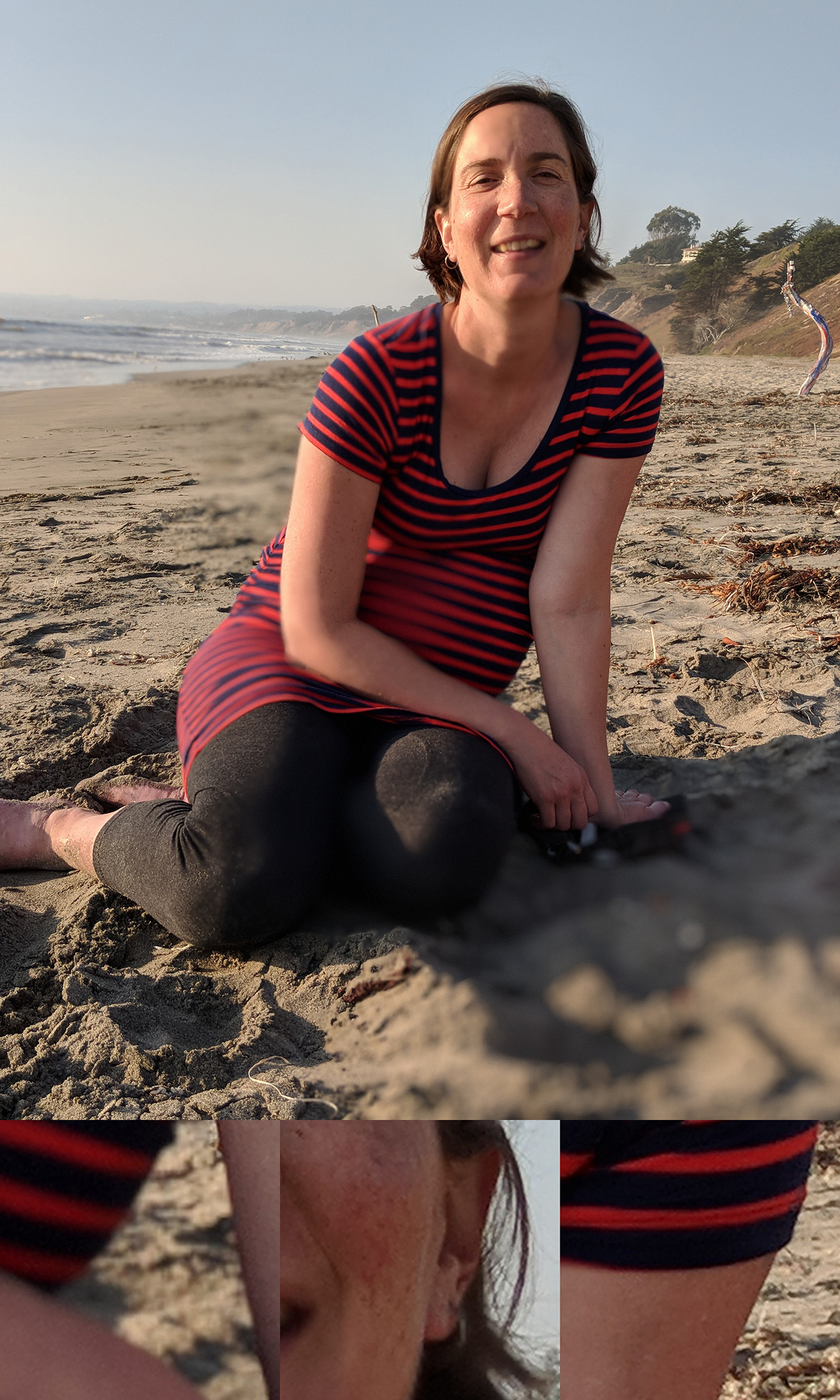} \\
    \includegraphics[width=\studyfigwidth]{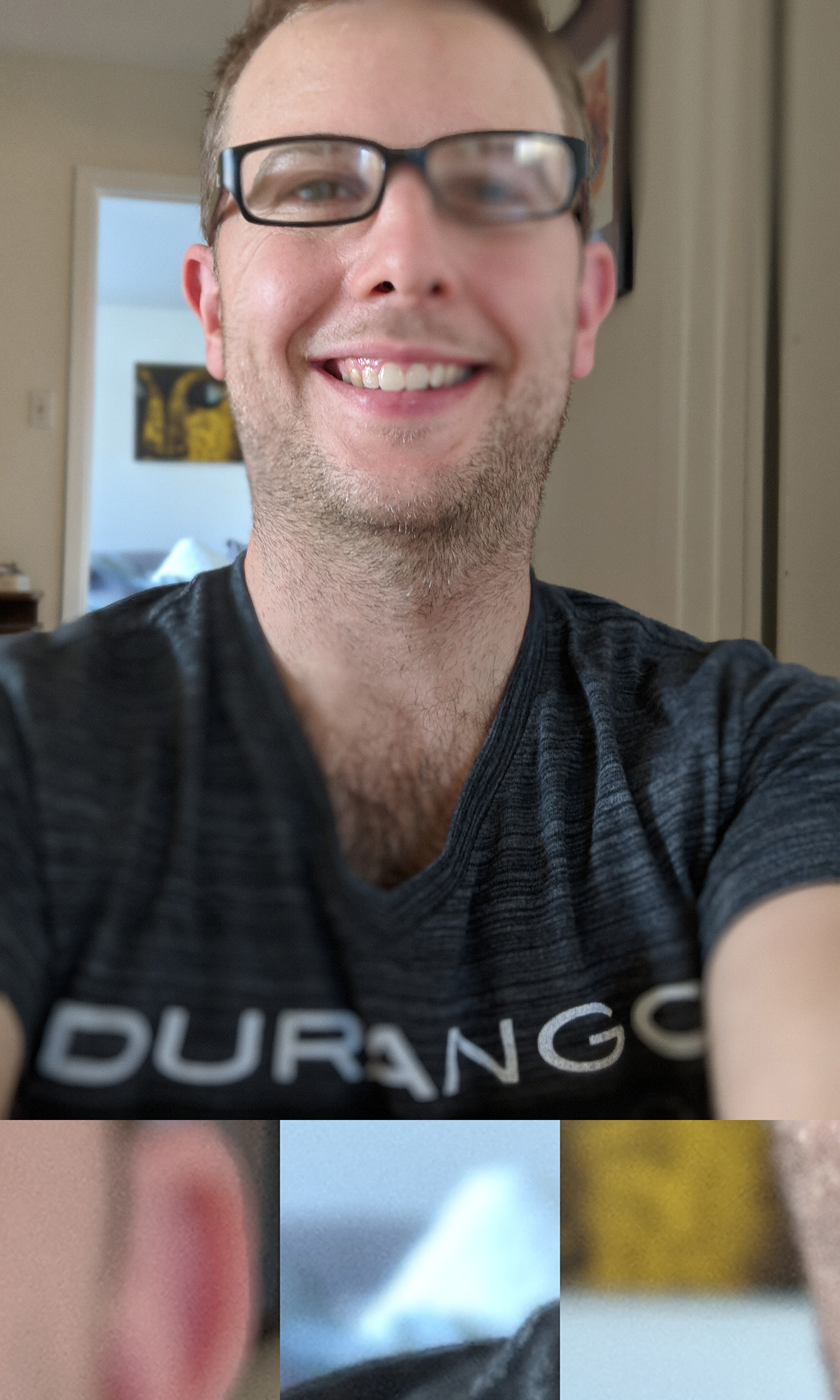} \\
    \includegraphics[width=\studyfigwidth]{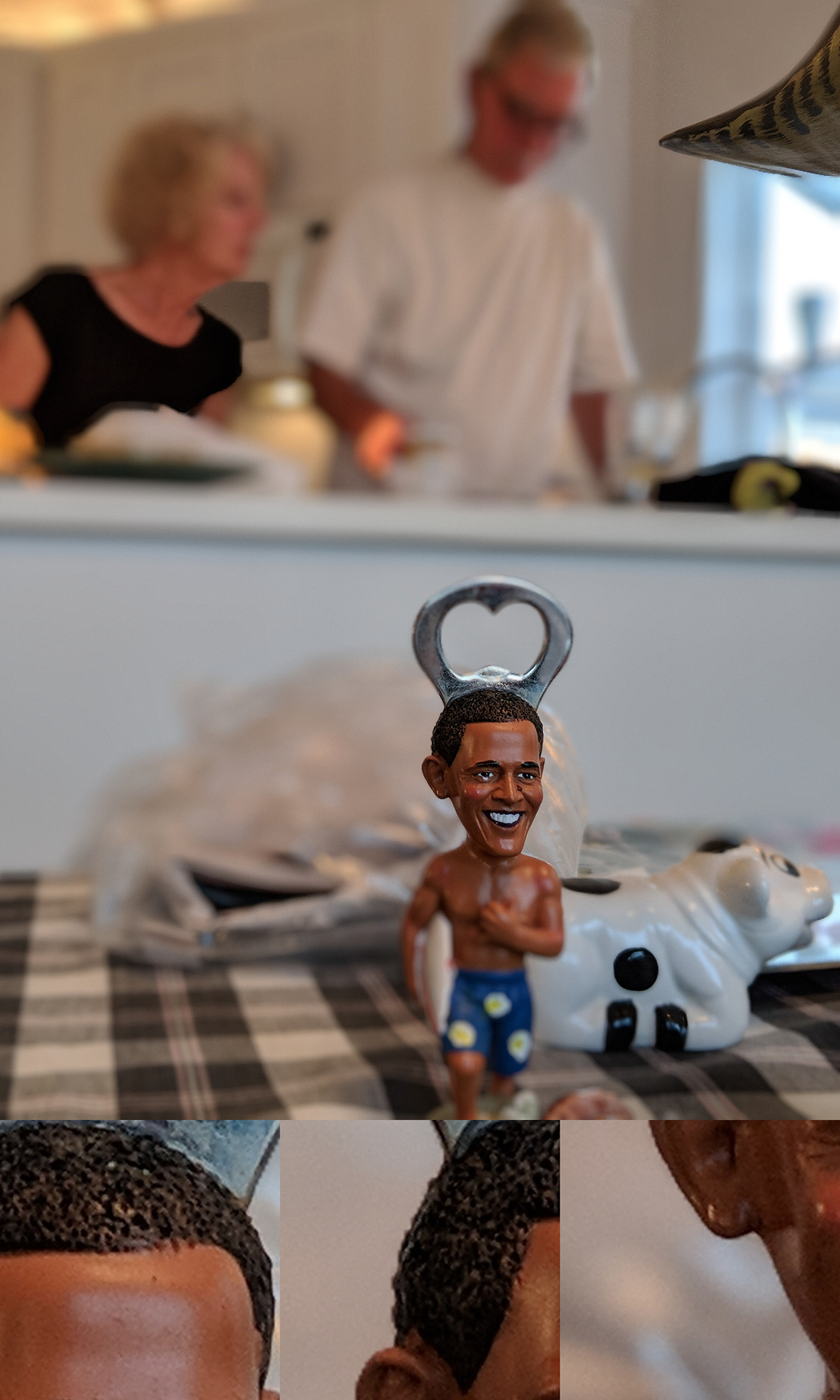}
    \end{tabular}
    }
    \subfigure[Shen \etal~\shortcite{shen2016automatic}]{
    \begin{tabular}{@{}c@{}}
    \includegraphics[width=\studyfigwidth]{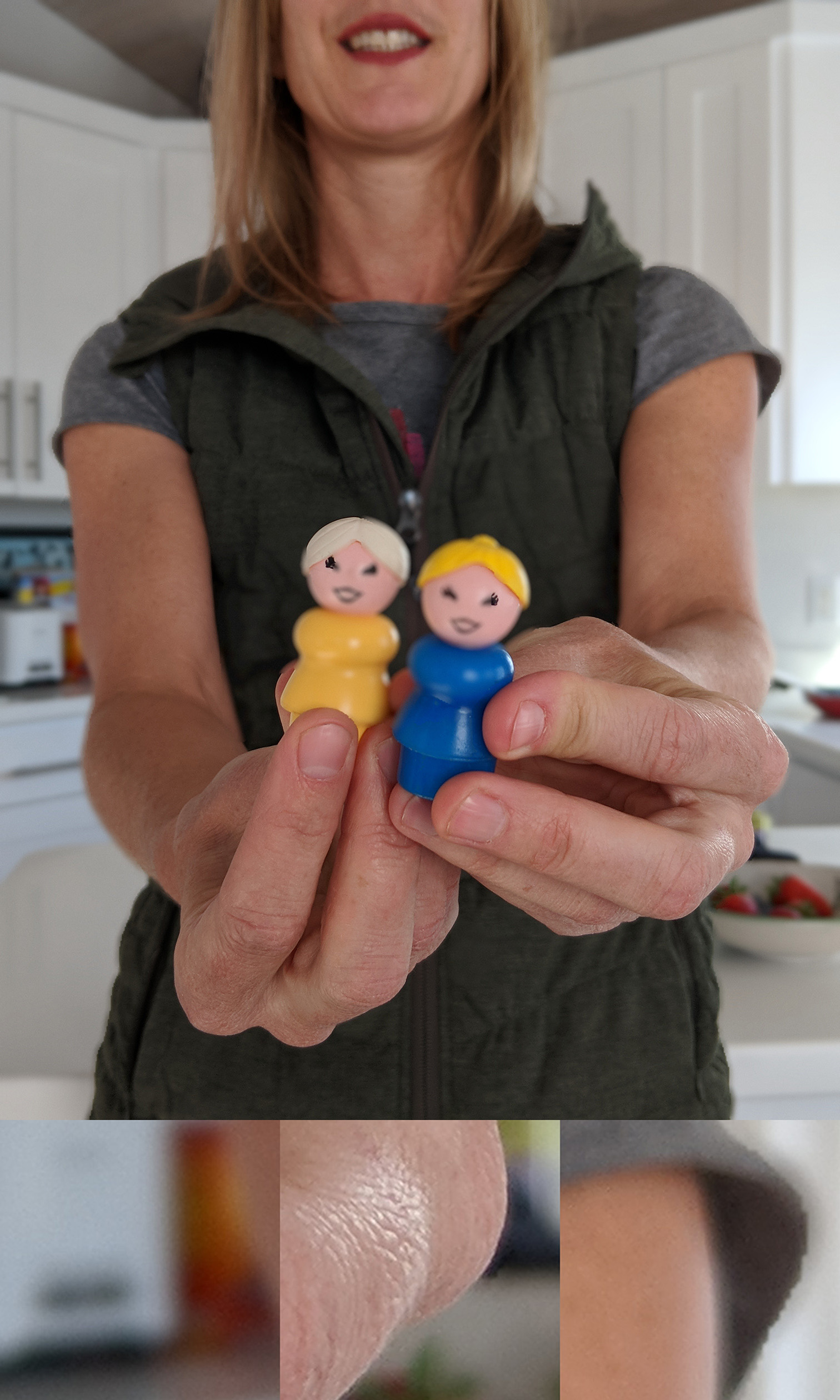} \\
    \includegraphics[width=\studyfigwidth]{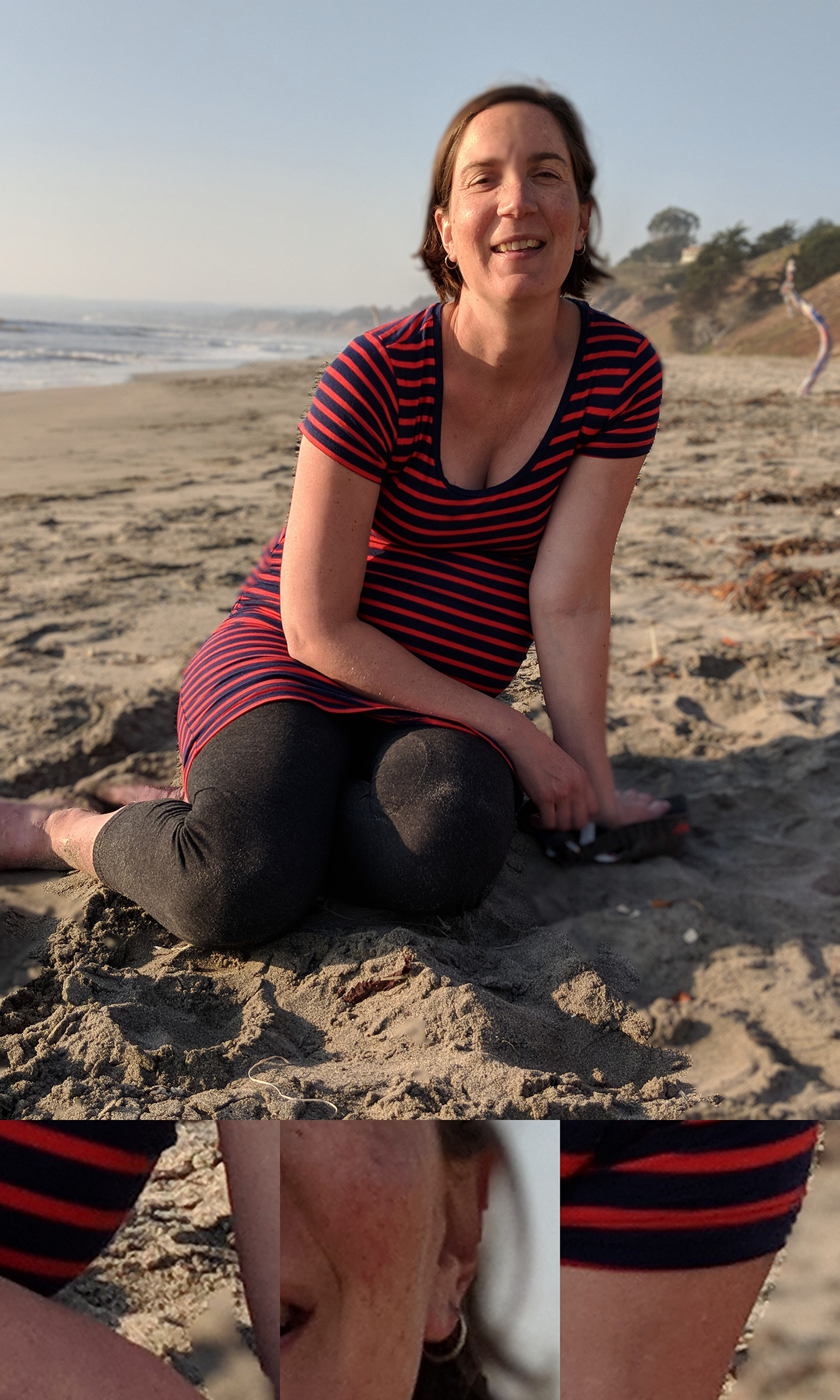} \\
    \includegraphics[width=\studyfigwidth]{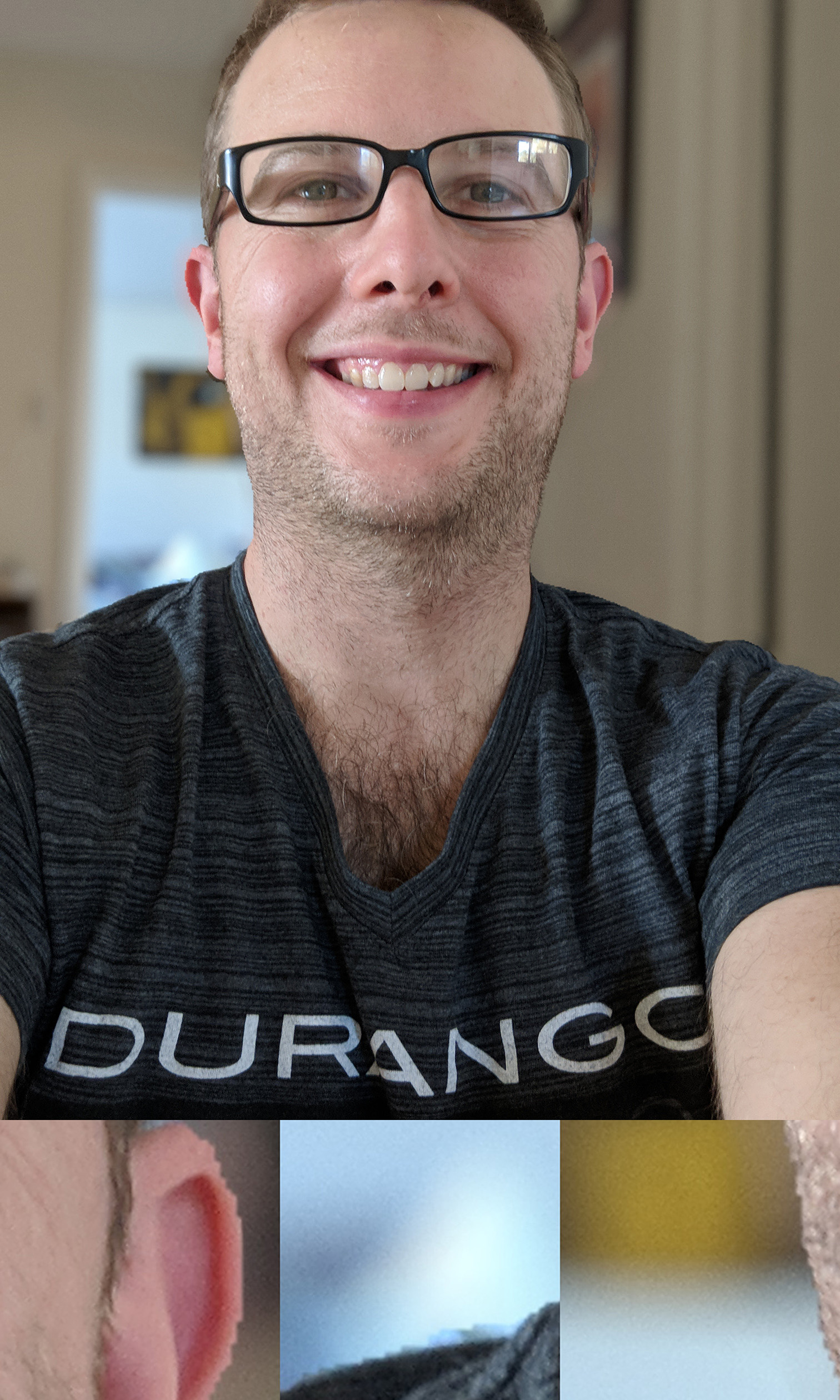} \\
    \includegraphics[width=\studyfigwidth]{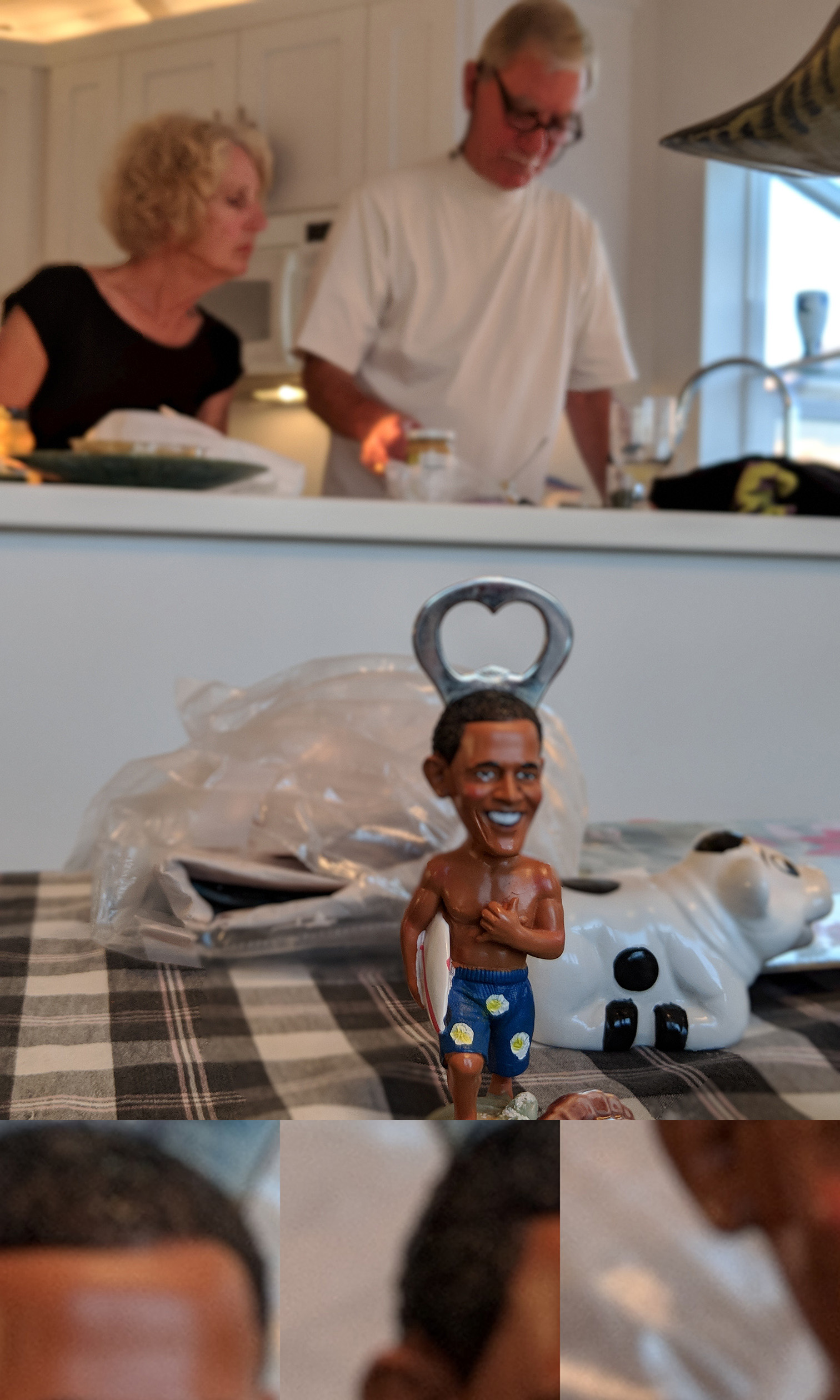}
    \end{tabular}
    }
    \subfigure[DP Only (ours)]{
    \begin{tabular}{@{}c@{}}
    \includegraphics[width=\studyfigwidth]{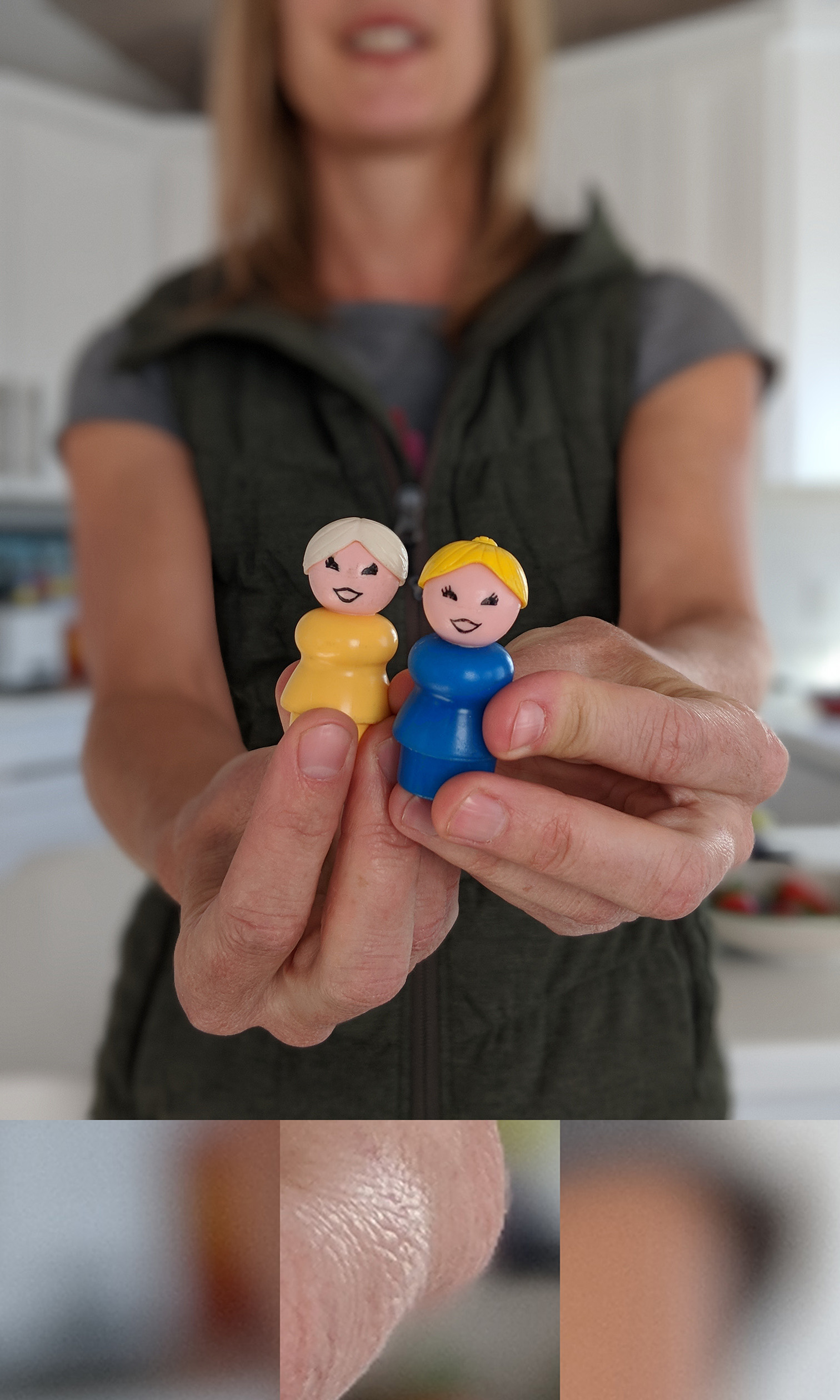} \\
    \includegraphics[width=\studyfigwidth]{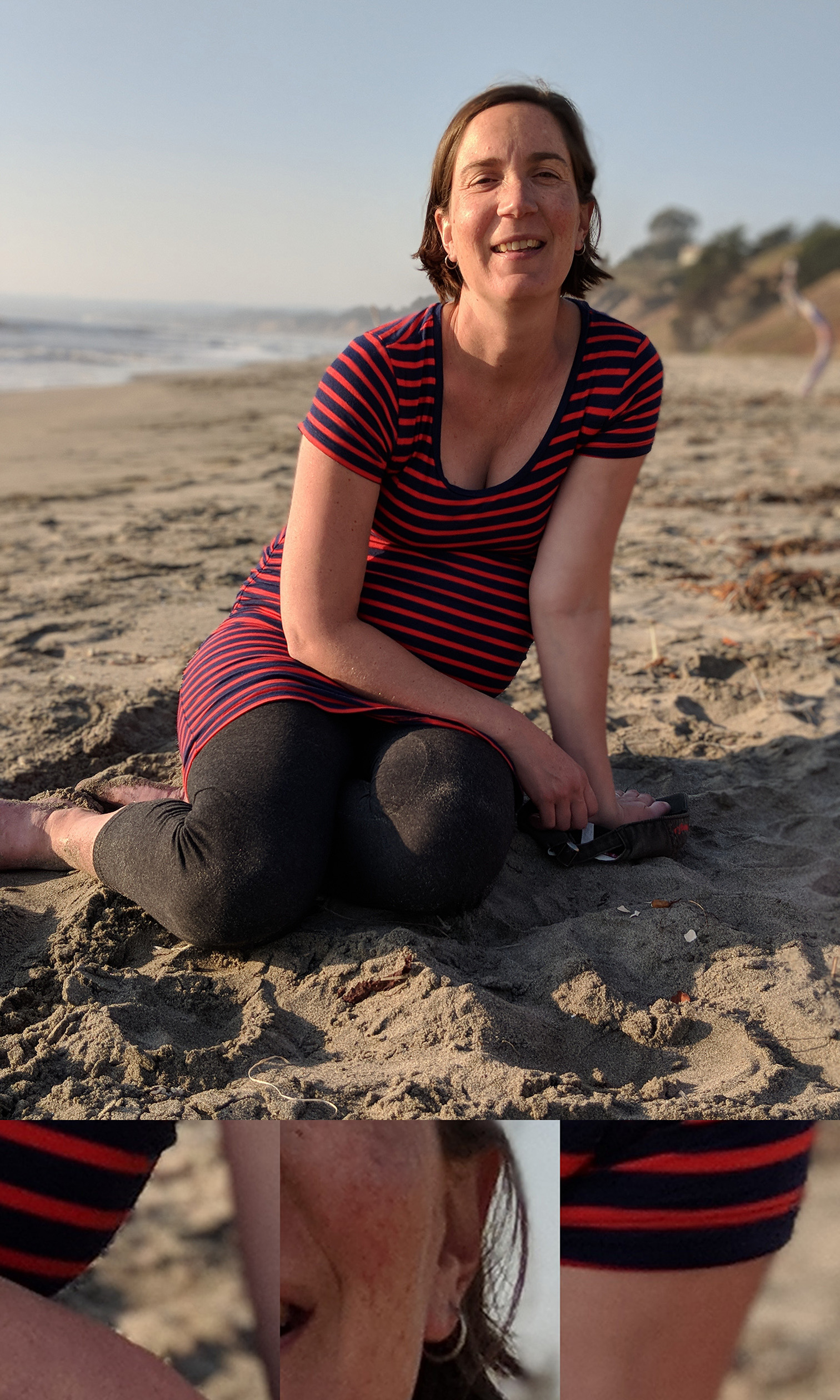} \\
    \includegraphics[width=\studyfigwidth]{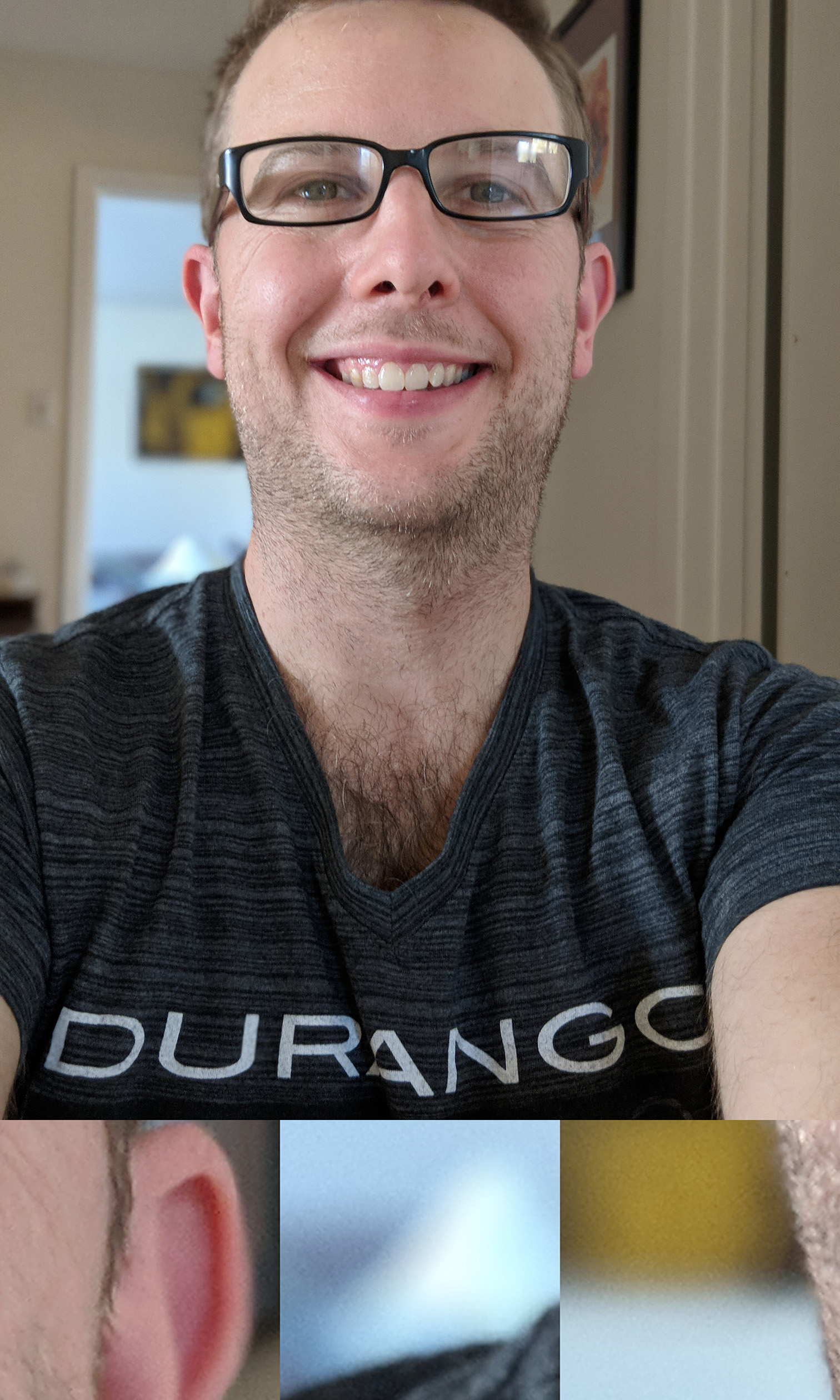} \\
    \includegraphics[width=\studyfigwidth]{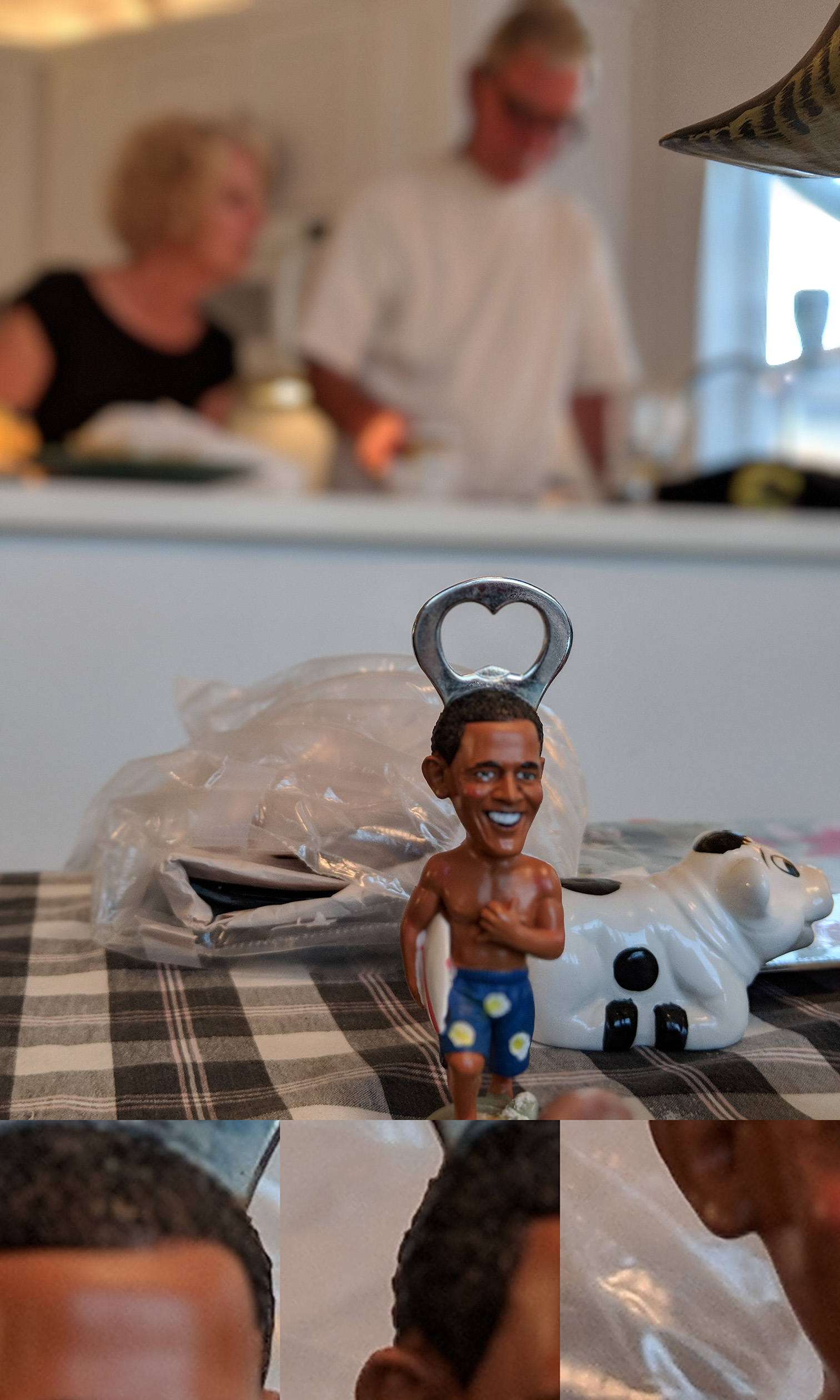}
    \end{tabular}
    }
    \subfigure[Segmentation Only (ours)]{
    \begin{tabular}{@{}c@{}}
    \includegraphics[width=\studyfigwidth]{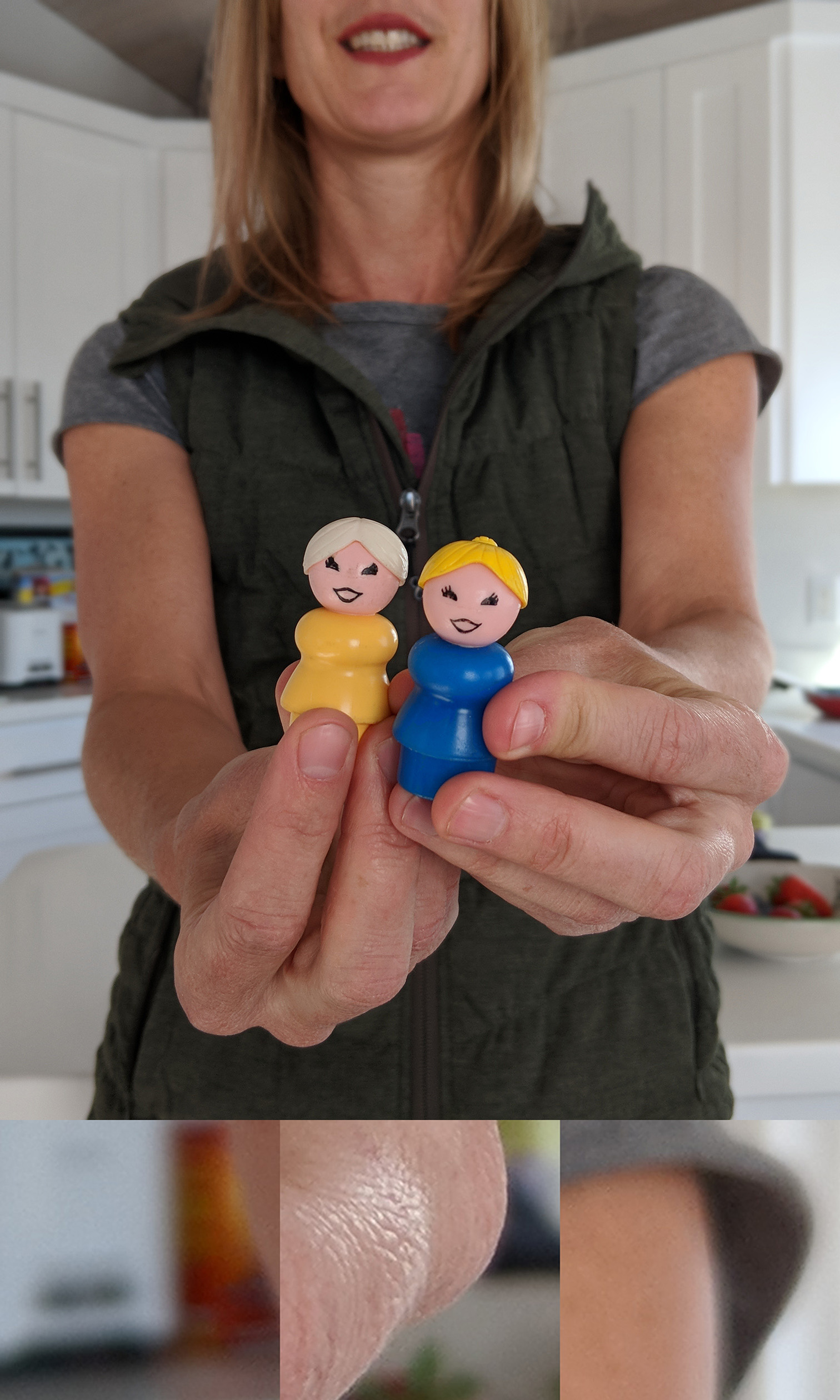} \\
    \includegraphics[width=\studyfigwidth]{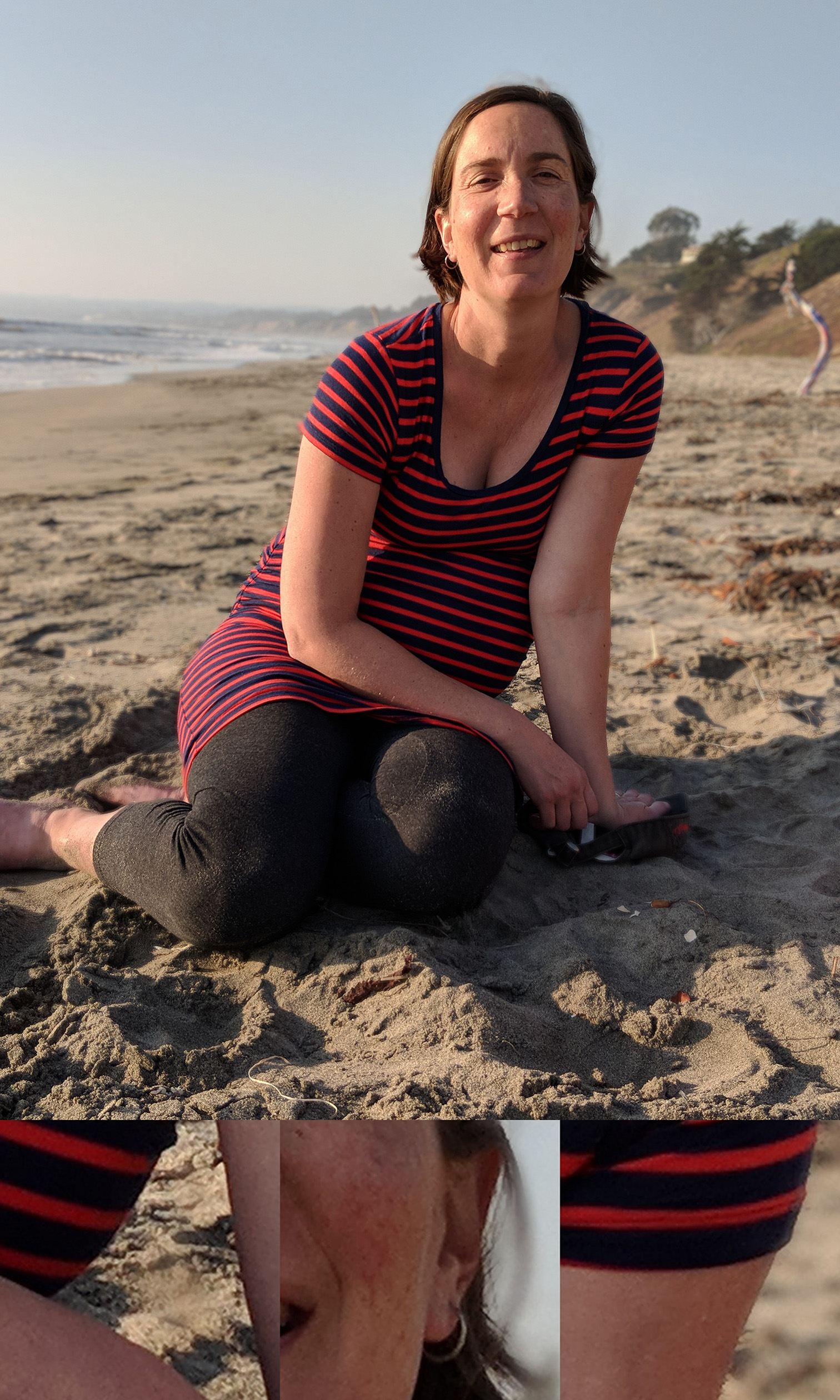} \\
    \includegraphics[width=\studyfigwidth]{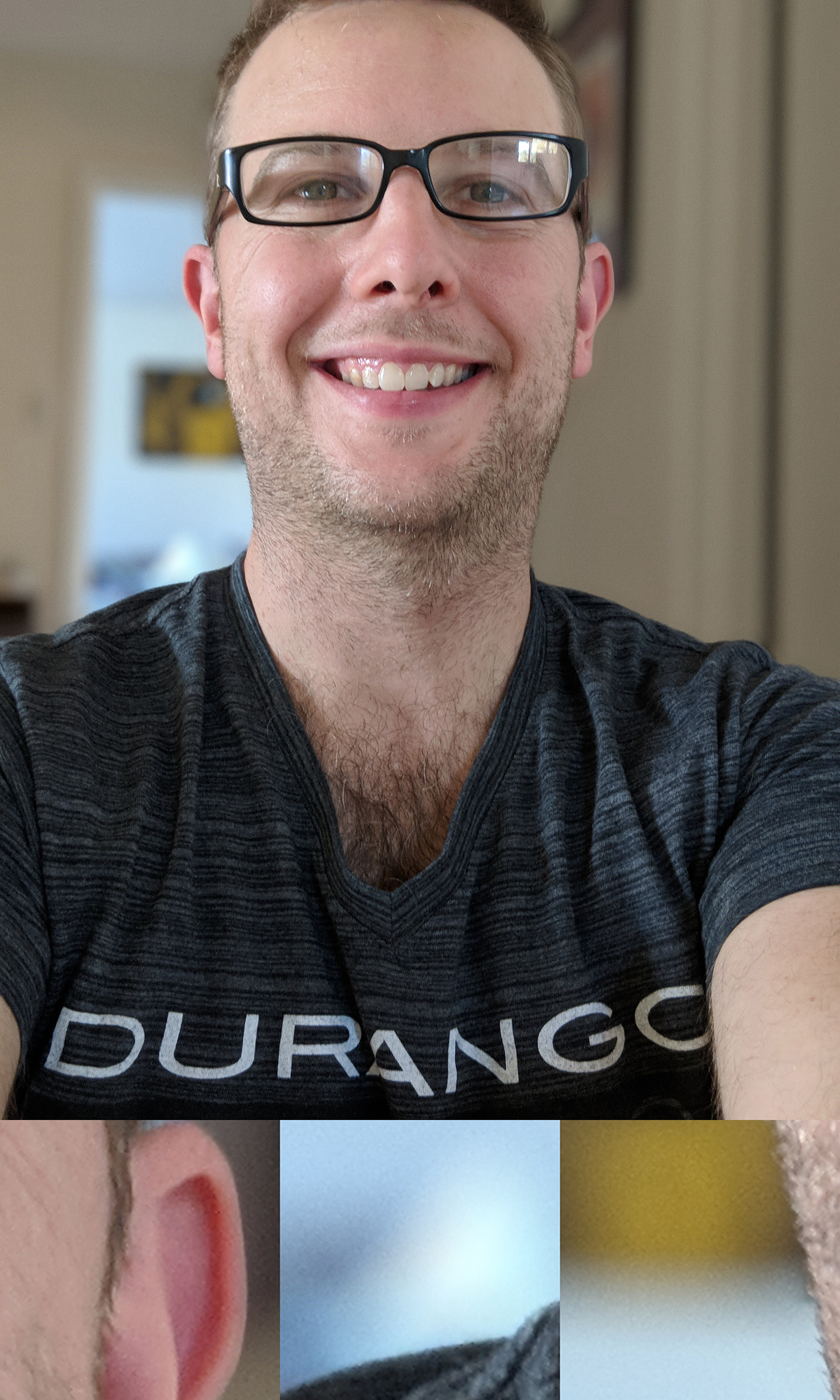} \\
    \includegraphics[width=\studyfigwidth]{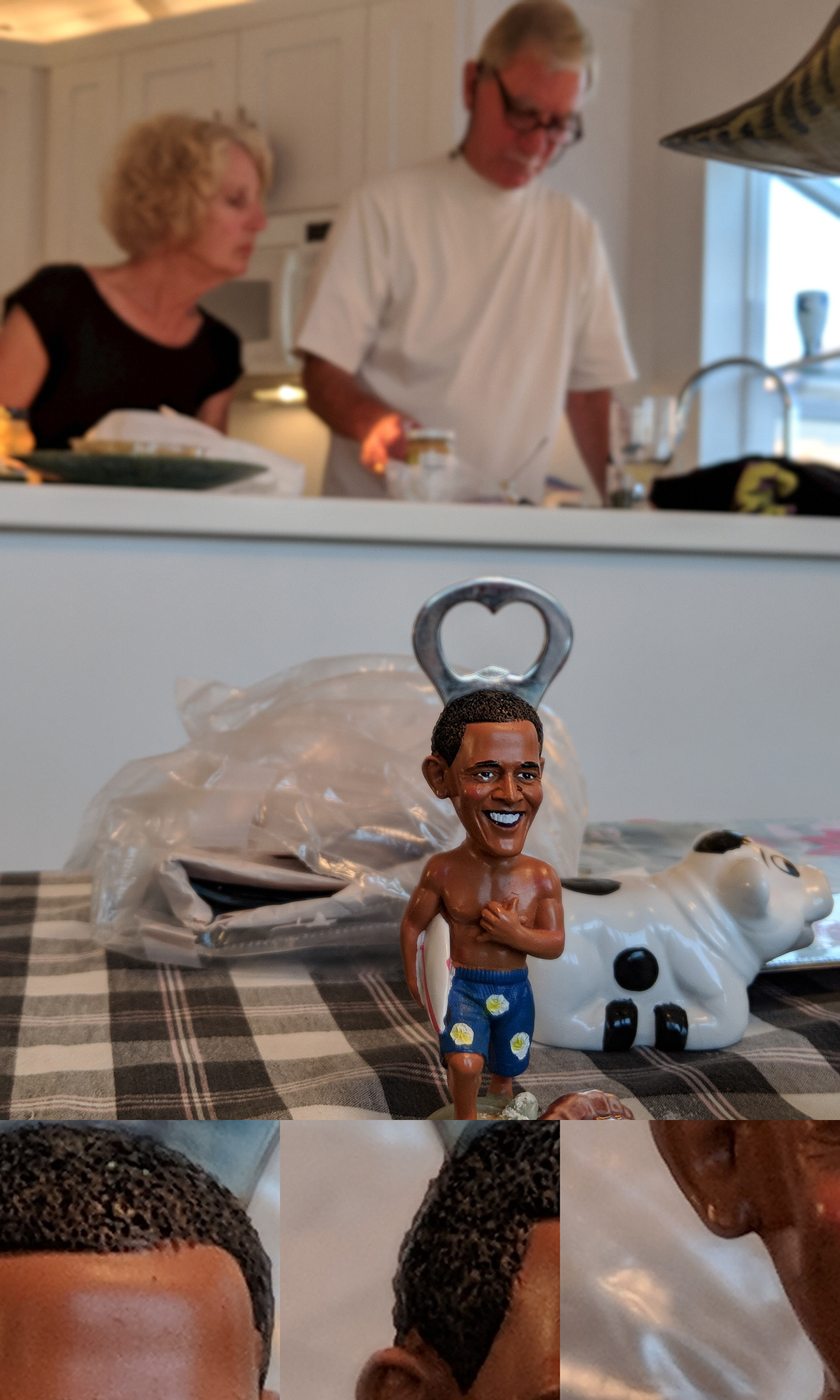}
    \end{tabular}
    }
    \subfigure[DP + Segmentation (ours)]{
    \begin{tabular}{@{}c@{}}
    \includegraphics[width=\studyfigwidth]{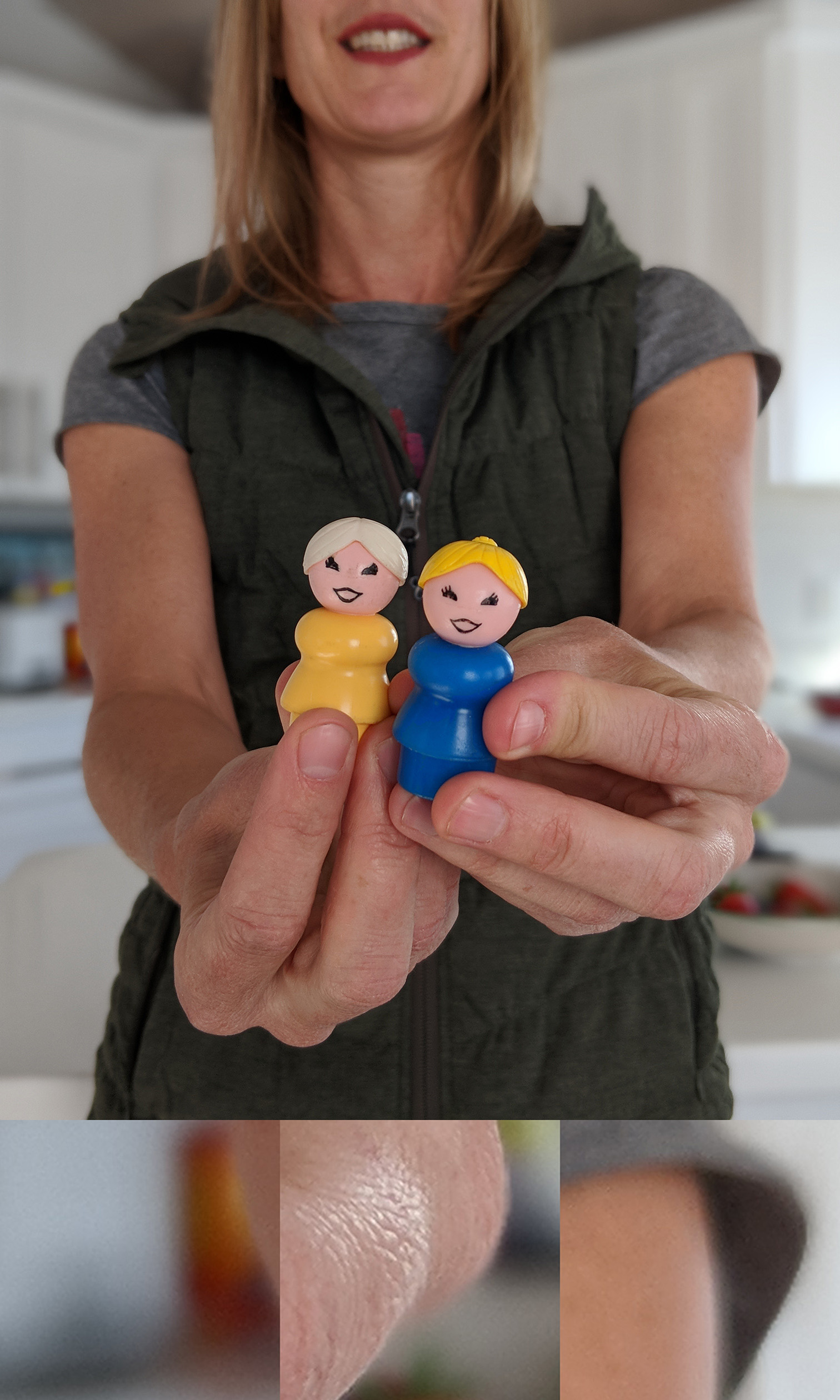} \\
    \includegraphics[width=\studyfigwidth]{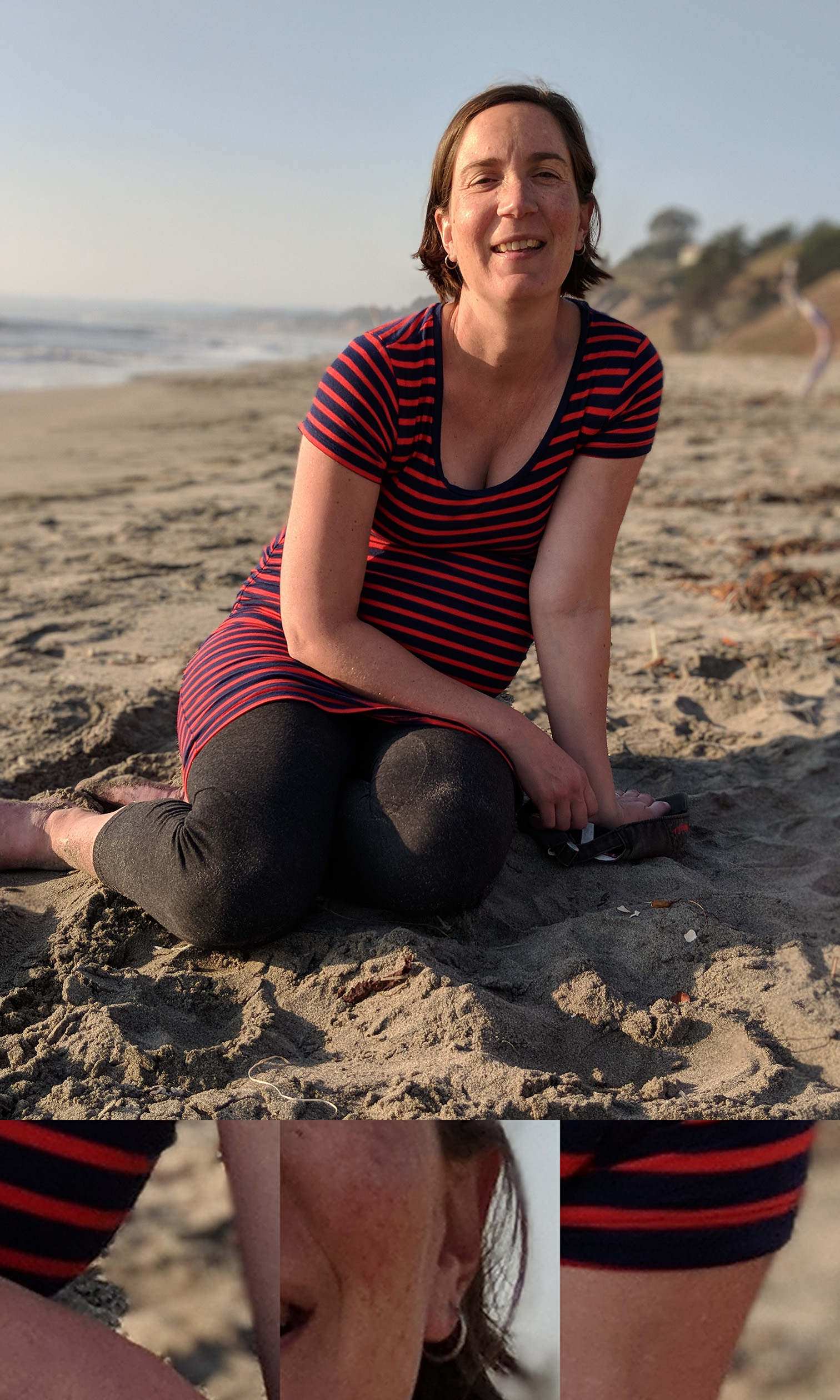} \\
    \includegraphics[width=\studyfigwidth]{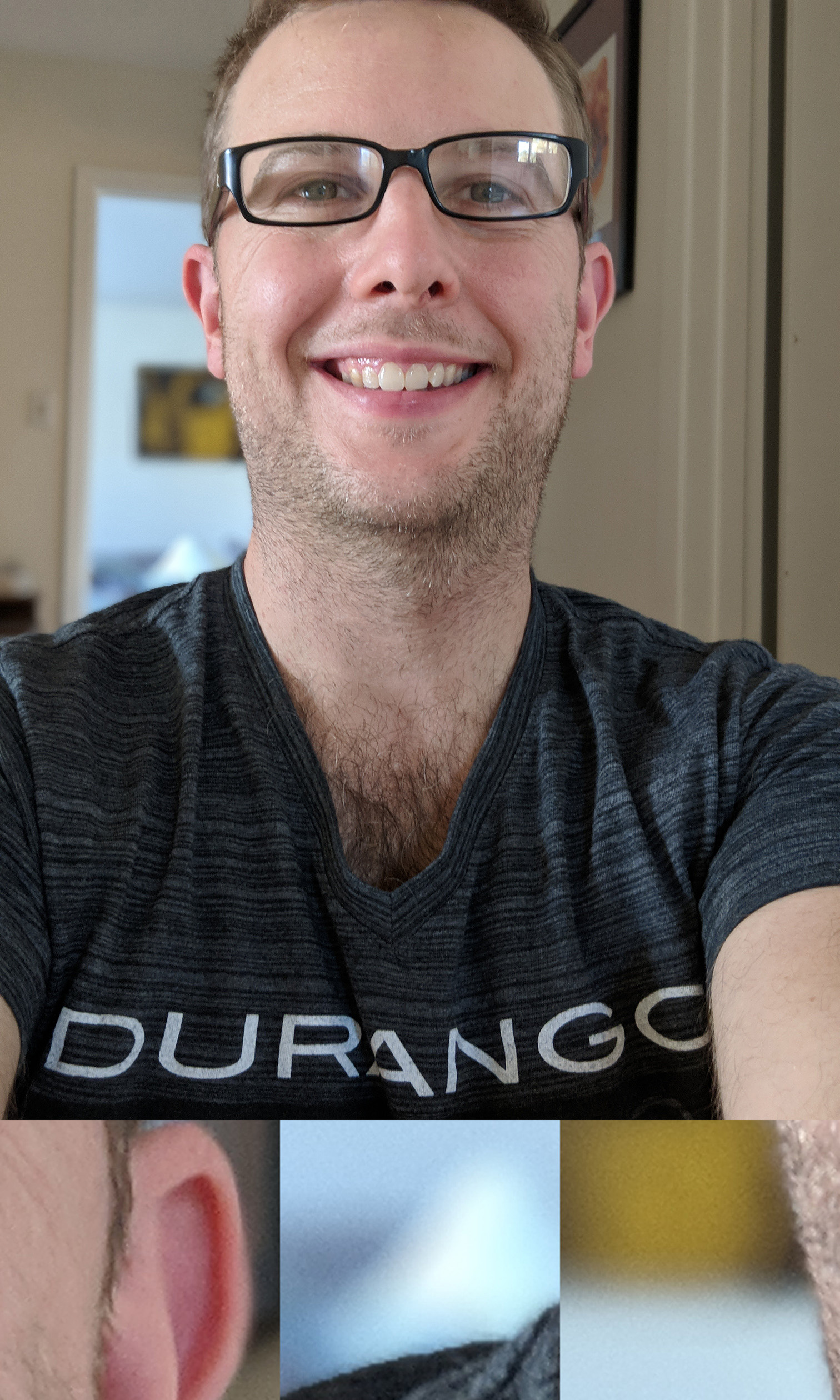} \\
    \includegraphics[width=\studyfigwidth]{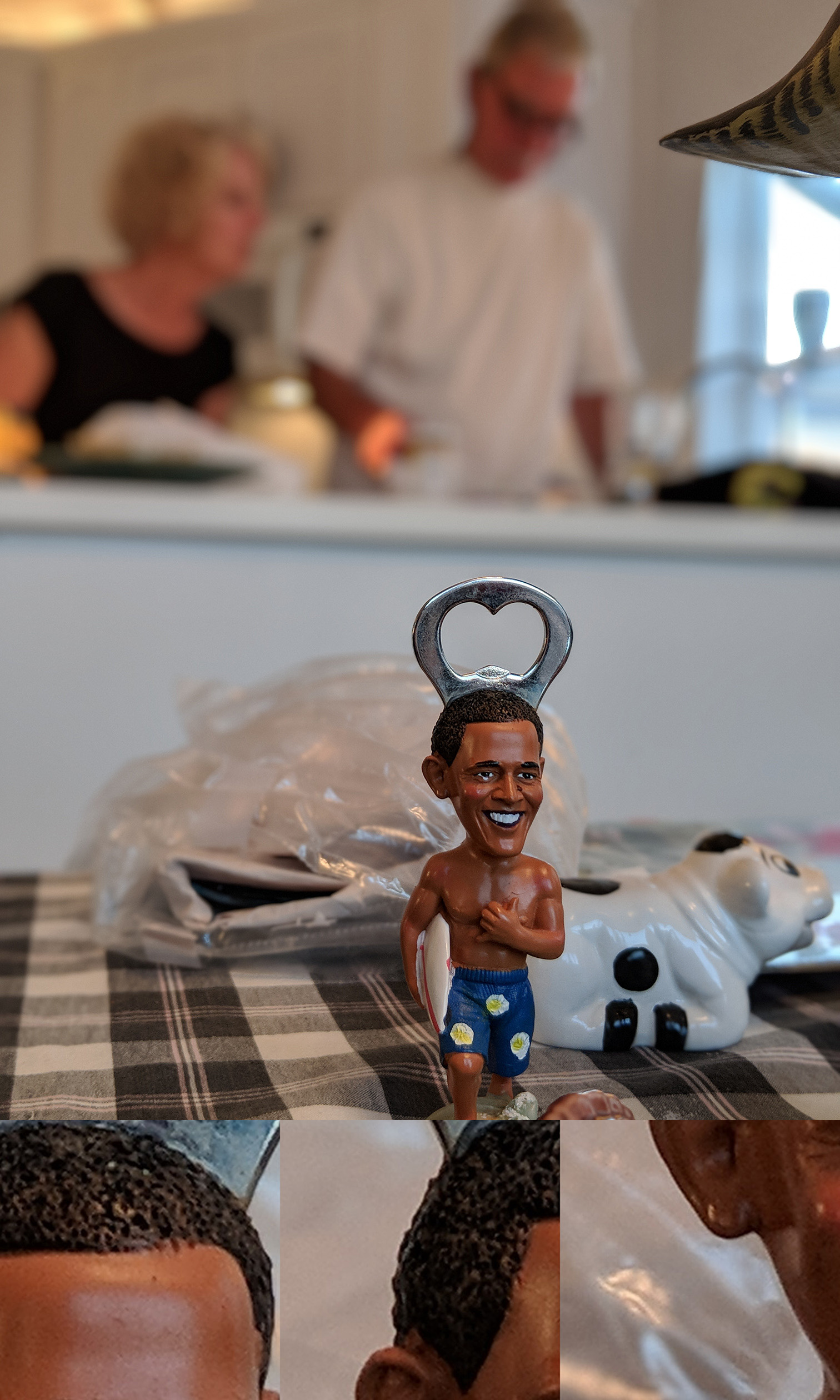}
    \end{tabular}
    }
    \caption{Another sample of the images used during our user study, shown in the same format as Fig.~\ref{fig:userstudy1}}
    \label{fig:userstudy2}
\end{figure*}

\bibliographystyle{ACM-Reference-Format}
\bibliography{references.bib}